\documentclass[Afour,sageh,times]{sagej}
\usepackage{moreverb,url}
\usepackage{algorithm}
\usepackage{algpseudocode}
\usepackage{booktabs}
\usepackage{multirow}
\usepackage[table]{xcolor}
\usepackage{pifont}
\usepackage{makecell}
\newcommand{\cmark}{\ding{51}} 
\newcommand{\xmark}{\ding{55}} 
\algrenewcommand\algorithmicrequire{\textbf{Input:}}
\algrenewcommand\algorithmicensure{\textbf{Output:}}

\usepackage[colorlinks,bookmarksopen,bookmarksnumbered,citecolor=red,urlcolor=red]{hyperref}

\newcommand\BibTeX{{\rmfamily B\kern-.05em \textsc{i\kern-.025em b}\kern-.08em
T\kern-.1667em\lower.7ex\hbox{E}\kern-.125emX}}

\makeatletter
\renewcommand\paragraph[1]{\textbf{#1}}
\makeatother

\def\volumeyear{}
\def\volumenumber{}
\def\issuenumber{}
\def\journalname{}
\renewcommand{\titlesize}{\fontsize{17}{19pt}\selectfont} 

\makeatletter
\def\ps@title{%
  \def\@oddhead{}%
  \let\@evenhead\@oddhead
  \def\@oddfoot{}%
  \let\@evenfoot\@oddfoot}

\makeatletter
\renewcommand{\@maketitle}{%
\if@Royal
\vspace*{-20pt}
\fi
\if@Crown
\vspace*{-20pt}
\fi
\vspace*{-34pt}%
\null%
\begin{center}
\if@PCfour
\begin{rm}
\else
\begin{sf}
\fi
{\raggedright\titlesize\textbf{\@title} \par}%
\vskip 1.5em%
{\par\large%
\if@Royal
      \vspace*{6mm}
      \fi
      \if@Crown
      \vspace*{6mm}
      \fi%
      \lineskip .5em%
      {\raggedright\textbf{\@author}
      \par}}
     \vskip 20pt%
    {\noindent\usebox\absbox\par}
    {\vspace{20pt}%
      {\noindent\normalsize\@keywords}\par}
      \if@PCfour
      \end{rm}
      \else
      \end{sf}
      \fi
      \end{center}
      \if@Royal
      \vspace*{-4.5mm}
      \fi
      \if@Crown
      \vspace*{-4.5mm}
      \fi
      \vspace{22pt}
        \par%
  }
\makeatother

\begin{document}

\title{FlyCo: Foundation Model-Empowered Drones for Autonomous 3D Structure Scanning in Open-World Environments}

\author{Chen Feng\affilnum{1}, Guiyong Zheng\affilnum{2}, Tengkai Zhuang\affilnum{3}, Yongqian Wu\affilnum{3}, Fangzhan He\affilnum{3}, Haojia Li\affilnum{1}, Juepeng Zheng\affilnum{2}, Shaojie Shen\affilnum{1}, and Boyu Zhou\affilnum{3,4}}

\affiliation{\affilnum{1}Department of Electronic and Computer Engineering, The Hong Kong University of Science and Technology, Hong Kong, China\\
\affilnum{2}School of Artificial Intelligence, Sun Yat-sen University, Zhuhai, China \\
\affilnum{3}Department of Mechanical and Energy Engineering, Southern University of Science and Technology, Shenzhen, China\\
\affilnum{4}Differential Robotics, Hangzhou, China\\
Chen Feng and Guiyong Zheng contributed equally to this work.}

\corrauth{Boyu Zhou, Southern University of Science and Technology,
Shenzhen, China.}

\email{zhouby@sustech.edu.cn}

\pagestyle{empty}

\begin{abstract}
Autonomous 3D scanning of open-world target structures via drones remains challenging despite broad applications. 
Existing paradigms rely on restrictive assumptions or high-effort human priors, limiting practicality, efficiency, and adaptability. 
Recent foundation models (FMs) offer great potential to bridge this gap. 
Thus, this paper investigates a critical research problem: \textit{What system architecture can effectively integrate FM knowledge for open-world target structure scanning?} 
We answer it with \textbf{F}ly\textbf{C}o, a principled FM-empowered perception-prediction-planning loop enabling fully autonomous, prompt-driven 3D target scanning in diverse unknown open-world environments. 
\textbf{F}ly\textbf{C}o directly translates low-effort human prompts (text, visual annotations) into precise adaptive scanning flights via three coordinated stages: (1) perception fuses streaming sensor data with vision-language FMs for robust target structure grounding and tracking; (2) prediction distills FM knowledge and combines multi-modal cues to infer the partially observed target's complete geometry; (3) planning leverages predictive foresight to generate efficient and safe paths with comprehensive target coverage. 
Building on this, we further design key components to boost open-world target grounding efficiency and robustness, enhance prediction quality in terms of shape accuracy, zero-shot generalization, and temporal stability, and balance long-horizon flight efficiency with real-time computability and online collision avoidance.
Extensive challenging real-world and simulation experiments show \textbf{F}ly\textbf{C}o delivers precise scene understanding, high efficiency, and real-time safety, outperforming existing paradigms with lower human effort and verifying the proposed architecture's practicality. 
Comprehensive ablations validate each component's contribution. 
\textbf{F}ly\textbf{C}o also serves as a flexible, extensible blueprint, readily leveraging future FMs and robotics advances. 
Code and development environments will be released.
\end{abstract}


\maketitle

\section{Introduction}
Autonomous, accurate 3D scanning of human-intended open-world structures with aerial robots is a key enabler for infrastructure inspection (\citealt{infrasturcture}), disaster response (\citealt{disaster_management}), and heritage documentation (\citealt{cultural_preservation}).
However, achieving efficient scanning in diverse, unknown open-world environments remains challenging: the drone must precisely localize the user-specified structure amid clutter and execute compact, collision-free coverage without wasting time on irrelevant regions.

\begin{figure*}[t]
\centering
\includegraphics[width=0.9\linewidth]{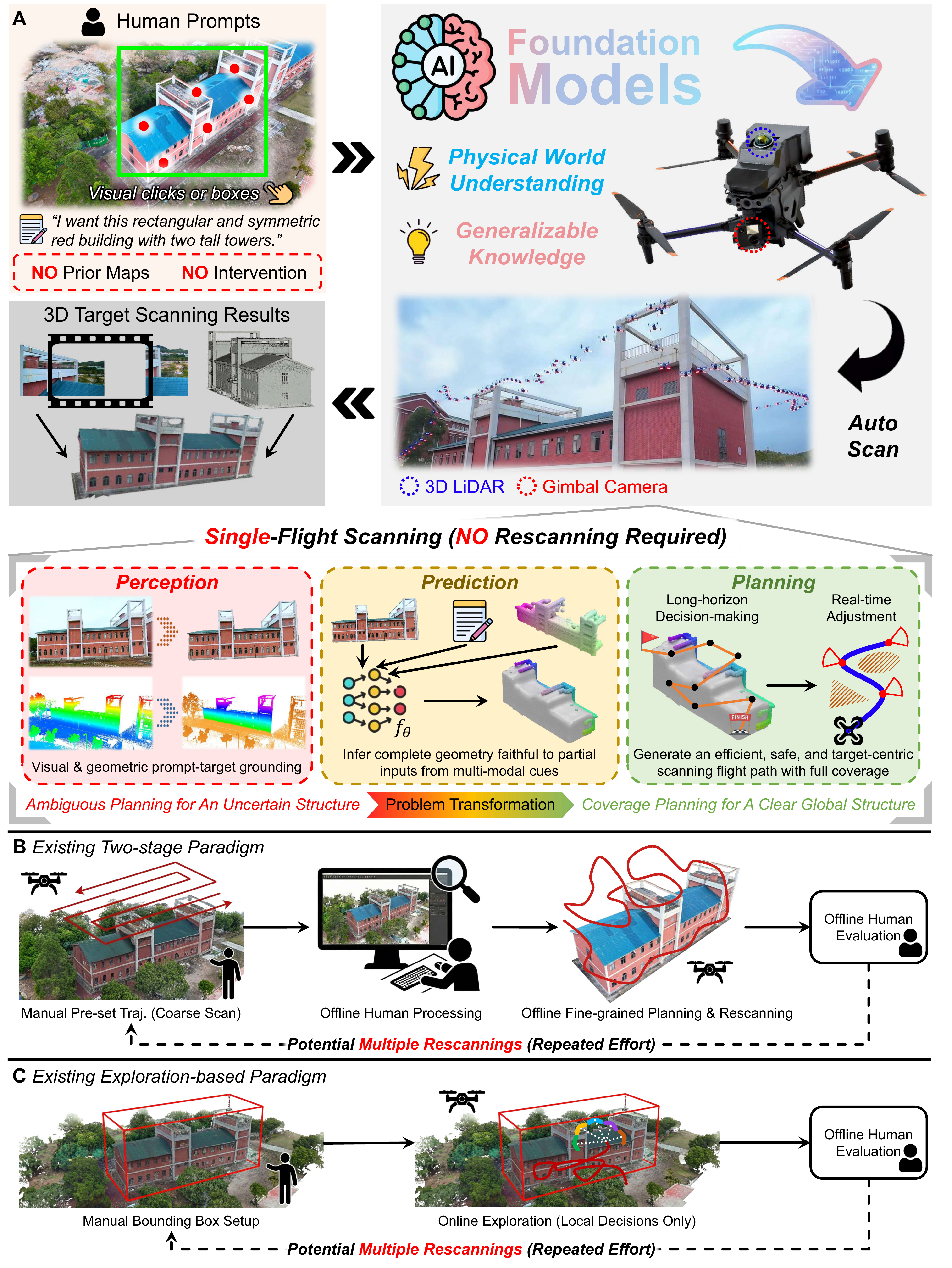}
\vspace{-0.3cm}
\caption{Teaser.
(A) Overview of the proposed FM-empowered aerial system to autonomously scan user-specified structures.
It seamlessly integrates physical world understanding and knowledge built into FMs with advanced autonomous flight capabilities, translating low-effort human prompts (text descriptions, visual annotations) into precise drone actions for efficient target scanning in unknown open-world environments.
This is achieved via a coordinated perception-prediction-planning loop, which operates entirely without human intervention or prior environmental knowledge.
(B-C) Existing paradigms typically rely on effortful human priors and involvement, and tend to trigger re-flights for incomplete scans, both of which limit practicality, efficiency, and adaptability in open-world scenarios.
}
\label{fig:Teaser}
\vspace{-0.5cm}
\end{figure*}

Human pilots handle this reliably.
Given an instruction such as ``\textit{scan the castle in the valley},'' they can quickly identify the intended structure, infer its full layout from partial views for guiding motion, and coordinate sensing and motion to efficiently acquire dense coverage while avoiding obstacles in real-time.
This stems from rich world knowledge about 3D spaces and the seamless integration with flight skills.
In contrast, existing drones lack such generalizable capabilities and still rely heavily on expert operation or high-effort human priors to balance efficiency, information completeness, and safety across new scanning tasks.

Early solutions (\citealt{djiflightplanner,skydio}) rely on pre-defined flight patterns (\textit{e.g.}, lawnmower), performing limited adaptability to complex geometries and requiring user-configured parameters.
Subsequent works (\citealt{roberts2017submodular,hepp2018plan3d,smith2018aerial}) improve coverage quality via two-stage pipelines, combining coarse scans with offline fine-grained planning over human-processed geometry (Fig. \ref{fig:Teaser}B).
Other approaches (\citealt{jing2016view,zhou2020offsite}) replace coarse scans with external priors like satellite imagery or pre-existing 3D models.
While better handling complex structures, they still depend on human setup or prior data and fail to handle unexpected obstacles.
Exploration-based methods (\citealt{song2017online,bircher2018receding,luo2024star}) further reduce human involvement by treating target scanning as online exploration (Fig. \ref{fig:Teaser}C), but typically operate within user-specified 3D bounding boxes, making performance highly sensitive to box accuracy (overly large boxes waste effort on irrelevant regions, whereas tight boxes may miss target parts).
Moreover, without global scene context, these systems react locally to current observations, generating inefficient paths.
In practice, both paradigms frequently necessitate post-flight human evaluations and repeated re-flights, undermining operational efficiency and flexibility.
Recently, several works (\citealt{feng2023predrecon,guedon2023macarons,chen2024gennbv}) incorporate reasoning about unobserved target geometry to provide global context.
Despite their promise, these methods are evaluated primarily in small-scale simulations or simplified environments with no target-irrelevant obstacles and restricted structure diversity, leaving their zero-shot generalization to real-world scenes and practical applicability unclear.
Collectively, existing paradigms fall short of the foundational understanding and world knowledge of human pilots, relying on manual setup, strong human priors, or idealized assumptions that hinder low-effort, efficient operations that are robust and support strong generalization in the open world.

The core bottleneck of existing aerial systems—\textit{lack of scene understanding and world knowledge}—is precisely where recent foundation models (FMs) offer the potential to reduce reliance on high-effort human priors.
Their success in vision-language alignment and visual generation suggests a compelling avenue towards human-like spatial understanding and imagination for drones.
Although FMs have already advanced ground-robot navigation and manipulation, naively mirroring them fails to address the unique challenges of open-world 3D aerial scanning. 

\textbf{First}, most prior works integrate FMs into robotic systems via end-to-end (e2e) learning, requiring massive expert demonstrations. 
For open-world aerial scanning, where the drone must simultaneously balance coverage, efficiency, and safety across highly diverse scenarios, collecting massive expert flight data is impractical. 
Additionally, FM's high inference latency conflicts with the real-time response needed for smooth, efficient flight. 
\textbf{Second}, FMs are primarily designed for language and 2D vision.
Distilling their knowledge into 3D spatial domains to enable drones to recognize complex targets amid clutter and predict their unseen parts remains underexplored.
Concretely, target recognition must handle streaming sensor data and stay robust to viewpoint shifts in long-sequence flights, while off-the-shelf FMs are memory-intensive and lack stability.
For reliable planning guidance, current prediction methods neither generalize well to novel structures in physical-scale nor infer plausible shapes from partial observations while strictly aligning with them; outputs unfaithful to observed parts risk geometric hallucinations, mismatches, and reduced scan efficiency (\citealt{yan2025symmcompletion}; Sec. \nameref{subsubsubsec:partial_loss}).
These gaps go far beyond trivially fine-tuning FMs using 3D data and call for careful designs. 
\textbf{Third}, FM-driven robot motion strategies typically rely on short-horizon reasoning, which misaligns with 3D scanning's demand for efficient global coverage planning. 
They also neglect temporal consistency with historical paths, risking trajectory inconsistencies and reduced efficiency. 
Moreover, global planning incurs heavy computational overhead, which leads to delayed planning updates that direct drones to target outdated surfaces while impairing real-time obstacle response. 
Together, these gaps raise an unanswered system-level question: \textit{What architecture can effectively integrate foundation-model knowledge with flight skills to truly achieve performant, practical aerial target structure scanning across diverse open-world environments?}

In this work, we address this question with \textbf{F}ly\textbf{C}o (Fig. \ref{fig:Teaser}A and Extension 1), a holistic FM-powered aerial system enabling precise, efficient, and safe scanning of diverse complex user-specified 3D structures in unknown open-world environments.
Driven solely by abstract human prompts (\textit{e.g.}, textual descriptions, visual annotations), it eliminates the need for effortful priors or interventions (\textit{e.g.}, 3D bounding boxes, human involvement) commonly required in existing paradigms.
To tackle the first challenge, we adopt a modular architecture inspired by expert pilots' cognitive processes for scanning tasks, comprising prompt-grounded scene understanding, spatial structural prediction, and efficient, safe flight planning.
Our design leverages FMs' strengths in perception and prediction while assigning long-horizon planning and real-time response, areas where FMs are deficient, to optimization-based methods.
This avoids the expert-data hunger of the e2e fashion while retaining the scene knowledge of FMs.
Asynchronous operation of the three modules eliminates mutual blocking in e2e frameworks.
Notably, the planning module harvests rich information from low-frequency FM inference outputs to conduct continual planning, reconciling the latency mismatch between slow FM reasoning and real-time flight needs.
The predictive capability enables global planning, overcoming the local myopia of existing methods.
\textbf{F}ly\textbf{C}o's flexible perception-prediction-planning loop and modular interfaces further enable future performance upgrades via direct module replacement.

For the second challenge, we carefully design two modules: prompt-grounded scene perception and multi-modal surface prediction.
Leveraging the generalizable understanding of FMs, perception aligns user prompts with streaming sensor data to segment target structure in 3D clutter, while our FM-tailored streaming-efficient adaptation and cross-modal refinement boost efficiency and ensure robustness across viewpoints and time.
For prediction, we introduce a multi-modal network that completes the target's 3D surface from partial views.
This design fuses complementary advantages from multiple sources, where partial geometric cues anchor outputs to physical measurements, while visual and textual cues that integrate distilled FM knowledge ensure generalization, enabling reliable inference of unseen parts.
Architecturally, we adhere to a minimal-inductive-bias principle with a post-fusion scheme, allowing the model to learn rich structural priors from data while effectively absorbing FM knowledge.
Beyond the network, we adopt complementary designs to address key challenges.
Specifically, a FM-based automatic data generation tool synthesizes aligned multi-modal data for large-scale annotation-free learning, boosting generalization across diverse structures and scales.
At inference, an adaptive densification strategy mitigates the inflexibility of fixed network outputs for structures with different scales, ensuring appropriate granularity for varying-sized targets without added model complexity.
Additionally, a lightweight partial-surface regularization during training explicitly encourages consistency with measured geometry, avoiding discrepancies between predicted and observed surfaces and enhancing temporal stability.
Together, these designs yield globally plausible and locally faithful predicted structures, converting uncertain unobserved regions into a clear global structural context to enable proactive goal-directed planning.

\begin{figure}[t]
\centering
\includegraphics[width=0.99\linewidth]{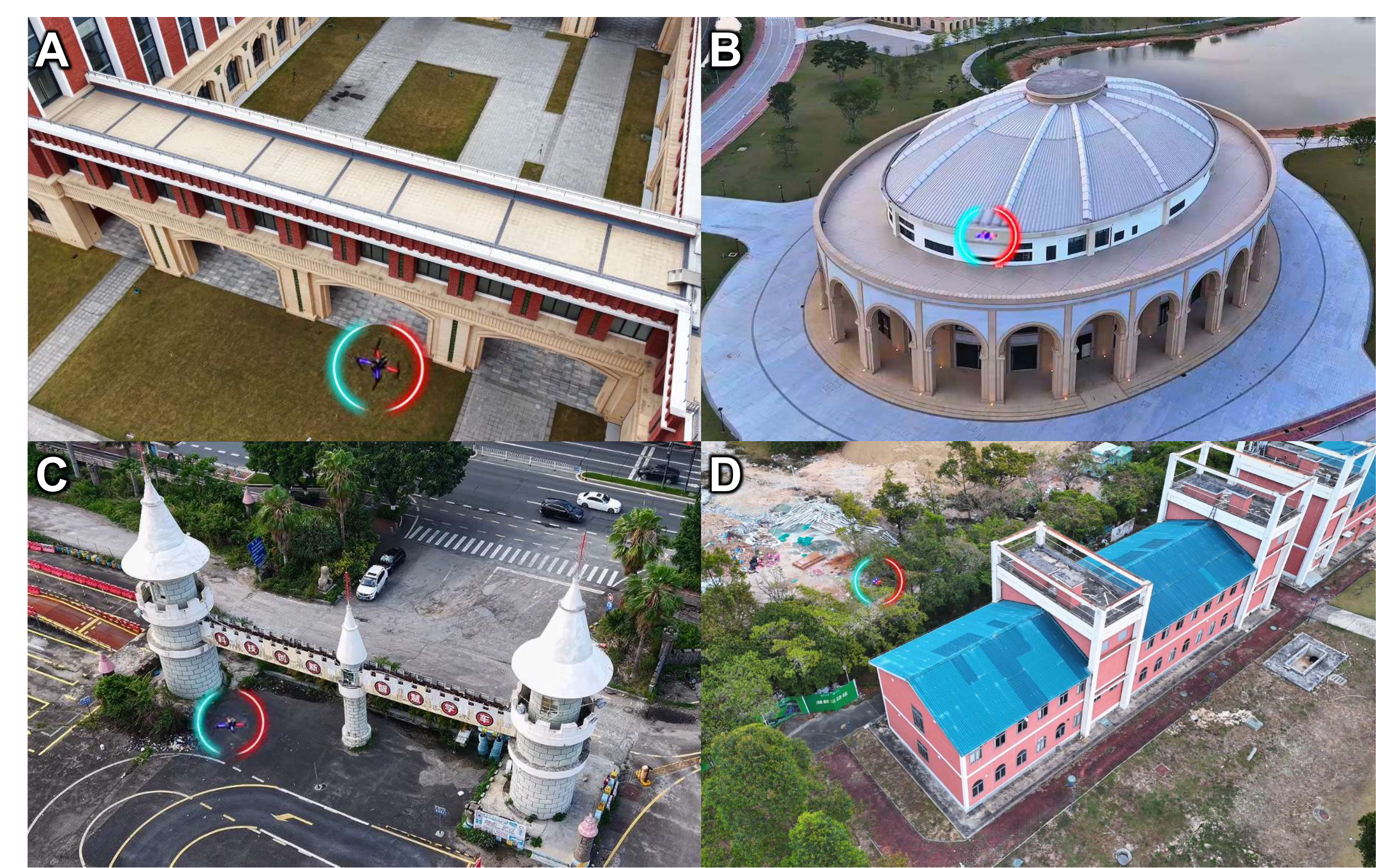}
\vspace{-0.3cm}
\caption{In-the-wild demonstrations across four scenarios.
The user-specified targets include: (A) an arch bridge linking buildings, (B) a large concert hall, (C) a castle gate, and (D) a red-brick building.
}
\label{fig:WildScenes}
\vspace{-0.4cm}
\end{figure}

Building on the prediction capability, we propose a prediction-aware hierarchical planner, reducing redundant flight caused by short-horizon myopia.
To ensure smooth path adjustments across historical and current paths, we present the injection of consistency awareness into global planning, improving long-term planning coherence and efficiency. 
Moreover, aiming at the high computational overhead of NP-hard coverage planning in physical-scale scenarios, we adopt two strategies: we employ inlier-driven intersection parity checking and gravitational-like viewpoint pruning to efficiently generate high-quality coverage viewpoints that keep pace with the latest prediction; we also adopt a two-level decomposition strategy to split the original problem into smaller solvable sub-problems, achieving the real-time solution of high-quality coverage paths. 
Finally, at the local level, guided by the global coverage path, we implement viewpoint-constrained trajectory optimization for high-frequency local planning, adapting to real-time safety requirements in the presence of emerging obstacles while preserving structural information completeness.

Overall, \textbf{F}ly\textbf{C}o leverages FMs to understand complex structures from simple text/visual prompts and seamlessly integrate scene knowledge with flight skills, enabling low-effort, efficient, reliable 3D scanning.
We evaluate the full \textbf{F}ly\textbf{C}o system on a customized aerial platform via diverse challenging field experiments at unmapped wild sites, covering key challenges: precise scene understanding, robust prediction in large-scale scenes, and efficient safe scanning in cluttered environments (Fig. \ref{fig:WildScenes}).
Field trials confirm \textbf{F}ly\textbf{C}o reliably scans user-specified structures from simple prompts with efficient, robust performance.
Benchmarked against state-of-the-art (SOTA) methods from existing paradigms in complex 3D simulated scenarios, \textbf{F}ly\textbf{C}o consistently outperforms competitors in terms of lighter human input, $1.25$-$3\times$ reduced flight time, $4.3$-$56.2\%$ higher target coverage, and $11.6$-$50.1\%$ higher success rate.
We complement these results with an in-depth analysis of \textbf{F}ly\textbf{C}o's advantages and practical viability relative to existing approaches. 
Ablation studies further validate the importance of each module, confirming our design choices.
In summary, our main contributions are as follows:
\begin{enumerate}
  \item[1)] A principled and flexible FM-empowered aerial system architecture for open-world 3D targeted structure scanning. 
  \textbf{F}ly\textbf{C}o seamlessly integrates generalizable FMs with advanced flight skills via a closed-loop perception-prediction-planning framework, enabling prompt-driven efficient scanning of complex structures in diverse, unknown open-world environments.
  \item[2)] A prompt-grounded perception module enabling practical FM-based 3D scene understanding for aerial scanning, in which FM-tailored streaming-efficient adaptation and cross-modal refinement boost efficiency and robustness across viewpoints and time.
  \item[3)] A novel multi-modal surface prediction pipeline harnessing FM world knowledge, fusing partial geometric and visual/textual cues for metrically grounded shape completion. 
  It features four key designs: a minimal-bias, post-fusion network powered by FMs, annotation-free data synthesis for large-scale learning, adaptive densification, and partial-surface regularization, achieving globally plausible, locally faithful results with strong zero-shot generalization and temporal stability.
  \item[4)] A prediction-aware hierarchical planner for efficient, safe long-horizon planning. 
  It incorporates consistency awareness injection, efficient viewpoint generation and two-level decomposition for real-time and consistent coverage planning; the local layer adopts viewpoint-constrained optimization to ensure safety amid unseen obstacles while preserving information completeness.
  \item[5)] Extensive validation of \textbf{F}ly\textbf{C}o via real-world in-the-wild flights and large-scale simulation benchmarks against SOTA methods, complemented by analyses and ablations, collectively demonstrating the system's superior performance and architectural validity. 
  Source code and development environments will be publicly available.
\end{enumerate}

\section{Related Work}
\label{sec:related_work}

\subsection{Classic 3D Aerial Scanning}

Early work on aerial 3D scanning in practice and in the research literature largely followed pre-defined flight patterns, typically implemented as commercial tools for mapping and inspection (\citealt{djiflightplanner,djiterra,flylitchi,pix4dcapture,skydio}).
These systems generate simple, parameterized trajectories such as lawnmower, orbit, and grid patterns once the user has drawn a polygon or box around the area to be scanned and chosen parameters like altitude, overlap, and speed.
Because the paths are planned without explicit consideration of the underlying 3D geometry or real-time perception feedback, they cannot adapt to cluttered surroundings and often require manual adjustment by a skilled operator.
In addition, the high-altitude, geometry-agnostic nature of these trajectories typically limits the level of structural detail and leads to incomplete coverage, especially for near-ground or recessed features that fall outside the effective viewing geometry.

To better account for structural complexity, subsequent methods further began to integrate target geometry into the planning process via two-stage pipelines (\citealt{roberts2017submodular,hepp2018plan3d,smith2018aerial,peng2019adaptive,feng2024fc}).
Representative two-stage strategies first execute a coarse data-collection flight, then manually reconstruct the 3D model of the target from the captured data, and perform offline optimization to generate a dense inspection path, for example via submodular viewpoint selection for close-range coverage (\citealt{roberts2017submodular}). 
Building on this idea, \citealt{hepp2018plan3d} and \citealt{smith2018aerial} decoupled viewpoint generation from trajectory optimization and explicitly optimized sensor orientations to increase visibility, while \citealt{peng2019adaptive} introduced adaptive plane-wise grid sampling to reduce the viewpoint search from 3D to 2D.
Methods like (\citealt{feng2024fc}) further improved trajectory efficiency through topological guidance, while others (\citealt{jing2016view,zhou2020offsite,zhang2021continuous}) leveraged prior information such as satellite imagery, CAD/BIM models, or pre-existing reconstructions to replace the initial data-collection stage for reducing flight time.
Although these techniques improve scanning efficiency and achieve high-quality coverage, they remain tightly coupled to heavy human effort and scene-specific prior knowledge, and are typically executed in an offline fashion.
As a result, they offer limited real-time adaptability to unexpected obstacles and do not satisfy the level of autonomy required for open-world 3D target structure scanning.

\subsection{Exploration-based 3D Aerial Scanning}

To reduce manual setup and improve adaptability in unknown environments, a class of online autonomous exploration strategies (\citealt{song2021view,bircher2018receding,song2021view,luo2024star}) emerged.
They typically required users to specify a task range delineated by a 3D bounding box, under the premise that this volume fully encloses the target structure.
Subsequently, drones then survey all unknown space within this volume using frontier-based (\citealt{yamauchi1997frontier}) or next-best-view (NBV) (\citealt{connolly1985determination}) algorithms.
\citealt{song2017online} alternated between NBV selection and local scanning path planning, incrementally refining 3D coverage within the given bounded workspace.
\citealt{bircher2018receding} used a receding-horizon NBV strategy that chooses viewpoints in a Rapidly-exploring Random Tree (RRT) over free space and selects paths maximizing surface information gain for 3D exploration and inspection.
On the other hand, \citealt{song2021view} extracted frontiers between current known and unknown space, then sampled viewpoints to observe these frontiers, and planned minimum-distance paths that visit the viewpoints sequentially.
More recently, Star-Searcher (\citealt{luo2024star}) further improved efficiency by introducing a LiDAR-camera cooperation workflow.
It adopted a frontier-based strategy to autonomously survey all occupied space within the given box using LiDAR and simultaneously generated the scanning trajectories for the camera in real-time to completely and detailly inspect all explored surfaces.
By replacing time-intensive coarse-to-fine workflows with single-pass missions, these methods reduce human effort on providing priors and enable real-time obstacle avoidance.

However, this paradigm suffered from fundamental flaws.
User-specified bounding boxes are often imprecise, leading to excessive scanning of non-target areas or missing parts of the intended structure.
Consequently, performance becomes highly sensitive to the quality of human inputs, which undermines practicality in open-world scenarios.
Besides, frontier and NBV strategies typically make locally optimal decisions with respect to current volumetric or surface information gain, rather than global coverage of a specific structure.
Such local decision-making yields globally inefficient and temporally inconsistent trajectories, increasing overall mission time or causing revisits to already well-observed regions.

\subsection{Prediction Involvement in Aerial Scanning}

Beyond pure exploration, a line of work (\citealt{feng2023predrecon,guedon2023macarons,chen2024gennbv,liu2024dg}) has recently started to bring predictive capabilities into aerial 3D scanning to provide global context guidance for pursuing higher efficiency, either by explicitly inferring unseen geometry or by learning to anticipate informative views.
On the geometric side, several methods (\textit{e.g.}, \citealt{feng2023predrecon}) trained specialized neural networks bridging mapping and planning.
They completed the target structure point cloud from partial volumetric maps and then ran coverage planning on the predicted model rather than on the raw, partially observed information.
This allows the planner to reason about occluded or yet-unseen parts to synthesize more structured paths, compared to purely local frontier or NBV strategies.
In parallel, the 3D vision community has made rapid progress on point cloud completion and generative 3D modeling, which similarly infer reasonable full shapes from partial observations (\citealt{yuan2018pcn,aiello2022cross,rong2024cra,tochilkin2024triposr,li2025step1x}). 
While these models have not yet been deployed on real robotic platforms, they suggest a promising source for future aerial scanning systems.
On the view-selection side, learning-based approaches (\citealt{guedon2023macarons,chen2024gennbv,liu2024dg}) replace hand-crafted information gain heuristics with policies or scoring networks trained to rank candidate viewpoints, for example, by reinforcement learning or self-supervision. 
These policies implicitly encode priors about what shapes and configurations are likely and which viewpoints tend to maximize structural coverage in the next step.

Nevertheless, these methods are studied under highly simplified conditions, where the scene typically contains only the target structure, with no target-irrelevant obstacles or cluttered surroundings.
Evaluation is conducted on limited synthetic benchmark datasets rather than through sustained deployment on real aerial platforms.
As a result, their practical viability in open-world settings where the target is embedded among many other objects and sensing conditions vary significantly remains unclear.
In particular, despite encouraging results in controlled environments, there is still no reliable, experimentally validated scheme for endowing drones with robust, generalizable predictive capabilities for 3D structure understanding in the wild.

\subsection{Foundation Models in Robotics}

Recent years have seen a rapid adoption of foundation models (FMs) in robotics, primarily using large language models (LLMs), vision-language models (VLMs) for ground robots and manipulators.
LLM-based systems (\citealt{ahn2022can, driess2023palm}) took natural-language prompts to specify high-level tasks and rely on the LLM to perform task decomposition, sub-goal generation, or chain-of-thought reasoning for mobile manipulators.
In these frameworks, the language model is grounded through affordance critics or embodied encoders, enabling robots to execute multi-step manipulation and navigation tasks directly from free-form textual instructions.

VLM-based approaches (\citealt{zitkovich2023rt,zhang2024uni,kim2024openvla,black2410pi0,zhou2025vision}) extended this idea by jointly leveraging vision and language.
Uni-NaVid (\citealt{zhang2024uni}) employed pretrained VLMs to interpret language instructions and video observations for a range of navigation tasks (\textit{e.g.}, instruction-following, object-goal navigation) within a unified model.
Vision-language-action models (\citealt{zitkovich2023rt,kim2024openvla,black2410pi0,zhou2025vision}) further coupled large-scale vision-language pretraining with robot datasets, treating actions as tokens and learning policies that map images and language inputs directly to low-level control.
These systems have demonstrated broad generalization across many table-top manipulation and indoor navigation tasks and highlight the value of FM backbones as rich semantic priors for robot decision making.

At the same time, FM-in-robotics research follows a data-hungry, end-to-end learning paradigm and is developed for relatively structured, short- to medium-horizon tasks under generous compute budgets. 
Their interfaces to FMs are typically formulated in terms of symbolic task steps, sub-goals, or pixel-space waypoints, rather than explicit 3D spatial representations.
Consequently, while existing FM-based robotic schemes highlight the potential of large pretrained models to endow robots with scene understanding and world knowledge, they stop short of providing a practical recipe for open-world aerial 3D target structure scanning.

\section{Problem Statement}
\label{sec:problem_statement}

The problem considered in this work is to scan all surfaces of a user-intended 3D structure $S$ (\textit{e.g.}, a castle) in an unknown open-world environment $W = \{S, O_1,\dots, O_N\}$ using an autonomous aerial robot, where $O_i$ denotes surrounding non-target obstacles (\textit{e.g.}, trees, buildings).
The robot is equipped with a gimbal-mounted RGB camera and a 3D LiDAR sensor, which capture visual and geometric information, respectively.
It executes a five-degree-of-freedom (DoF) trajectory $\mathbf{x}(t)$ over $t \in [0, T]$, specifying the 3D position of the drone and the orientation of the camera gimbal (\textit{i.e.}, pitch and yaw angle).
The camera is modeled by a viewing frustum with a limited field of view (FoV) and maximum sensing range; at each time $t$, only those surface points of $S$ that lie within this frustum and maintain direct line-of-sight visibility are considered visually observed.
Since the LiDAR's effective range typically exceeds that of the camera, geometry sensing is rarely the limiting factor, and in practice the bottleneck for usable scan quality lies in the camera. 
Accordingly, in our formulation, overall scanning completeness is effectively governed by the camera coverage.

The motion must satisfy hard safety and feasibility constraints: the drone should not collide with any element of $W$ at any time $t$ and must respect its dynamic limits.
Given a feasible trajectory $\mathbf{x}(t)$, we measure efficiency by the total flight time $T(\mathbf{x})$, and define an information completeness index $C(\mathbf{x}) \in [0,1]$ based on the fraction of the surface of $S$ that is ever seen by the camera along $\mathbf{x}(t)$.
Moreover, human involvement is represented through a prior input $h$ with an associated qualitative effort measure $E(h)$.
We do not assume a precise numerical model of $E(h)$.
Instead, it reflects the relative level of human effort required by different forms of prior input.
Intuitively, abstract, semantic, or easily specified inputs (\textit{e.g.}, text descriptions or sparse visual annotations) that are far from low-level control signals correspond to lower $E(h)$, whereas more concrete specifications or priors that require external data collection (such as 3D bounding box, pre-defined trajectory, satellite imagery, or 3D models) entail higher $E(h)$.
To summarize these considerations, we characterize the task as a multi-objective problem over
\begin{equation}
  \label{eq:formulation}
  \mathbf{J}(\mathbf{x},h) = \big(T(\mathbf{x}),\, 1-C(\mathbf{x}),\, E(h)\big),
\end{equation}
which captures the competing objectives of flight efficiency, coverage completeness, and human effort.
This formulation is purely illustrative and is not intended to be explicitly optimized.
Conceptually, it serves to formalize the design principles that motivate our system architecture, which seeks to achieve shorter flight time and more comprehensive visual coverage while relying solely on low-effort prior inputs from humans.

\section{System Overview}
\label{sec:system_overview}

\begin{figure}[t]
\centering
\includegraphics[width=0.99\linewidth]{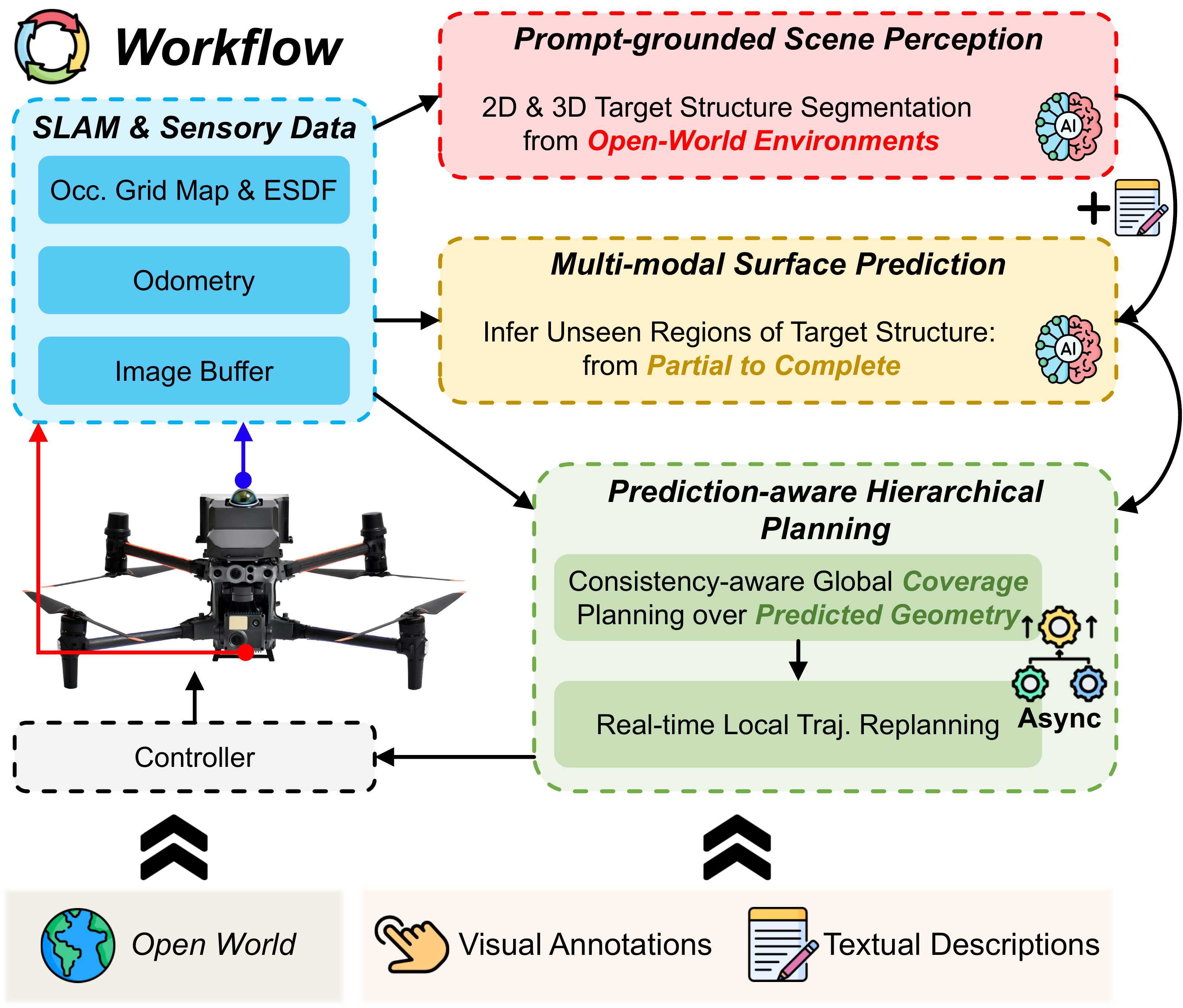}
\vspace{-0.3cm}
\caption{System architecture of \textbf{F}ly\textbf{C}o.
Given a human prompt, the system runs a closed-loop pipeline: leveraging FMs, perception grounds the target in 2D/3D from streaming RGB-LiDAR, and multi-modal prediction completes the target surface from partial observations; the planner asynchronously performs consistency-aware global coverage planning over the predicted geometry with real-time local trajectory replanning.
All modules are continuously updated until the mission terminates.
}
\label{fig:SystemOverview}
\vspace{-0.2cm}
\end{figure}

As illustrated in Fig. \ref{fig:SystemOverview}, \textbf{F}ly\textbf{C}o operates as a closed-loop system that starts from abstract human prompts and ends with a complete scan of the user-intended structure.
At the beginning of a mission, the user provides a high-level prompt $h$ (\textit{i.e.}, a textual description and sparse visual annotations on the onboard camera view) to specify the target structure $S$.
As the drone flies, \textbf{F}ly\textbf{C}o processes real-time streaming RGB images and LiDAR data with cross-modal refinement to continuously and precisely segment the target in cluttered 3D environments, thereby grounding the user prompt in the physical target structure (Sec. \nameref{subsec:perception}).
The resulting segmented visual and geometric observations of $S$ are then fused with the original textual prompt to infer a complete 3D surface representation of the target structure, including parts that have not yet been observed while remaining faithful to current measurements (Sec. \nameref{subsec:prediction}).
The predicted geometry serves as an intermediate bridge that distills the unknown and complicated open-world scanning task into a structured coverage planning problem with a clear global context.
Based on the latest prediction, \textbf{F}ly\textbf{C}o correspondingly plans a target-centric and efficient coverage path that balances short travel distances and flight consistency, surveying all unobserved structure areas without redundant scans.
In parallel, using the online map and following global paths, it dynamically generates the executable trajectory for the drone in real-time to ensure safety and information completeness. (Sec. \nameref{subsec:planning})
New sensor measurements together with the latest derived odometry and mapping results incrementally update perception and prediction, which in turn refine the planning objectives over time.
This loop repeats until the system autonomously detects that the target surface is sufficiently covered.
\textbf{F}ly\textbf{C}o then terminates the mission and delivers all scanned data to the user.
Throughout the above computation, all components run asynchronously in separate threads, reconciling slow FM-driven scene understanding with fast flight skills to execute the mission efficiently, accurately, and safely with low human effort.

\section{Methodology}
\label{sec:methodology}

\subsection{Prompt-grounded Scene Perception} 
\label{subsec:perception}

The primary role of the perception module is to interpret abstract user prompts and determine ``\textit{what to scan}'' at each moment.
It translates textual descriptions and visual annotations into precise pixel-level and point-level target segmentation in complex 3D environments from streaming RGB and LiDAR data.
Recent advances in visual segmentation foundation models, such as SAM2 (\citealt{ravi2024sam}), provide strong generic priors for object delineation and temporally consistent mask propagation in videos, making them a natural backbone for prompt-target grounding.
However, SAM2 natively supports only interactive visual inputs (points, boxes, and masks) and therefore cannot directly handle free-form textual prompts.
To enable text-conditioned segmentation, we integrate knowledge from VLMs such as BEiT3 (\citealt{wang2023image}) into SAM2 via an early-fusion scheme (\citealt{zhang2024evf}), which alleviates segmentation ambiguities that arise when using text or visual prompts alone.
Nevertheless, naively applying such off-the-shelf models is still insufficient for our task, due to two key challenges: (1) adapting them to real-time streaming data, and (2) transferring their segmentation capability to 3D LiDAR data while ensuring robust target grounding across viewpoints and over long-horizon flights.

\subsubsection{Streaming-efficient Adaptation for SAM2}
\label{subsubsec:streaming_adaptation}

The first practical challenge is that SAM2 is designed for offline video segmentation: it maintains a growing memory bank over all video frames to propagate masks, causing GPU memory usage and runtime to increase with sequence length.
In our long-horizon aerial scanning setting, where the drone may fly for thousands of frames, this behavior is impractical for onboard deployment.
Moreover, the original SAM2 model is computationally heavy for edge-computing aerial platforms, making real-time processing of streaming image data infeasible.
We address these issues with two key designs.
Instead of storing all frames, we maintain a sliding-window memory bank that retains only the most recent $W_m$ segmentation results, discarding older entries and thereby keeping the memory footprint bounded over long flights.
In parallel, we optimize the SAM2 backbone for onboard inference via lightweight quantization and graph-level compilation, which reduce per-frame latency without retraining.

\subsubsection{Cross-modal Refinement for Robust Target Grounding}
\label{subsubsec:cross_modal_refinement}

As for the second challenge, a straightforward idea is to project the segmented 2D masks from the camera images onto the 3D LiDAR points using known extrinsics and intrinsics to identify the target structure in 3D space.
However, this naive projection entirely relies on the accuracy and robustness of 2D segmentation, which may fluctuate or fail under strong viewpoint changes and occlusions during long-horizon flights (Sec. \nameref{subsubsec:ablation_perception}).
To overcome this limitation, we propose a cross-modal refinement mechanism that jointly leverages visual and geometric cues to stabilize target grounding, as summarized in Algo. \ref{alg:cross_modal_refinement}.

In parallel to 2D segmentation, an independent branch processes the raw LiDAR point cloud. 
At the beginning of a mission, the initial LiDAR data is partitioned into multiple groups using a density-based clustering algorithm (\citealt{ester1996density}) that accounts for both point density and spatial proximity. 
Each cluster is then projected onto the image plane, and its overlap with the 2D mask in the first frame is evaluated. 
Clusters whose projected points lie entirely within the mask are labeled as belonging to the target structure. 
During subsequent frames, geometric clustering is updated incrementally for computational efficiency: newly acquired LiDAR points are assigned to the nearest existing cluster if within a distance threshold, while points sufficiently far from all clusters are used to spawn new ones. 
This yields a continuously updated geometric estimate of the target in 3D space.

\begin{algorithm}[t]
\caption{Cross-modal Refinement Mechanism}
\label{alg:cross_modal_refinement}
\begin{algorithmic}[1]
\Require Current point cloud $\mathcal{P}_{t}$, current image $\mathcal{I}_t$, point clusters $\mathcal{C}_t$, memory bank $\mathcal{B}_t$, textual prompt $h_{\text{text}}$, reinitialization threshold $\kappa_{\text{reinit}}$
\Statex \hspace*{-\algorithmicindent}\textbf{Notation:} Current target 3D cluster $\mathcal{P}^{seg}_t$, memory keyframes $\mathcal{K}_t$, current 2D mask $M_t$, projected mask $M^{\text{proj}}_t$, virtual prompt points $p_{\text{v}}$
\State $\mathcal{P}^{seg}_t,\mathcal{C}_{t+1} \gets \text{GeometricClustering}(\mathcal{P}_{t},\mathcal{C}_t)$
\State $\mathcal{K}_t \gets \text{KeyframeSelection}(\mathcal{B}_t,\mathcal{P}^{seg}_t)$
\State $M_t,\mathcal{B}_{t+1} \gets \text{SAM2}(\mathcal{I}_t,\mathcal{K}_t,h_{\text{text}})$
\State $M^{\text{proj}}_t \gets \text{BackProjection}(\mathcal{P}^{seg}_t)$
\If{\texttt{BoxIoU}($M_t, M^{\text{proj}}_t$) $< \kappa_{\text{reinit}}$}
  \State $p_{\text{v}} \gets \text{FarthestPointSampling}(\mathcal{P}^{seg}_t)$
  \State $M_t,\mathcal{B}_{t+1} \gets \text{SAM2}(\mathcal{I}_t,p_{\text{v}},h_{\text{text}})$ \Comment{re-query with virtual prompts}
  \State $\mathcal{C}_{t+1} \gets \text{UpdateGeometricClustering}(\mathcal{P}_{t},M_t)$
\EndIf
\Ensure $M_t$, $\mathcal{P}^{seg}_t$
\end{algorithmic}
\end{algorithm}

Given the current geometric clustering, we first use it to choose informative and temporally stable keyframes from the memory bank. 
Vanilla SAM2 simply selects the most recent \(W_s\) frames (\(W_s < W_m\)) as memory keyframes for next-frame tracking, which ignores temporal consistency and geometric support, and can lead to drift. 
To encode the temporal coherence of 2D masks, we maintain a linear Kalman filter over their motion in image space. The state vector is defined as
\begin{equation}
  \mathbf{b} = [b_x, b_y, b_w, b_h, \dot{b}_x, \dot{b}_y, \dot{b}_w, \dot{b}_h]^\top,
\end{equation}
where \((b_x, b_y)\) denote the center of the minimum bounding box of the segmentation, \(b_w, b_h\) its width and height, and the remaining entries their velocities. 
Using a constant-velocity motion model \(\mathbf{F}\), the state prediction at time \(t\) is
\begin{equation}
  \hat{\mathbf{b}}_{t|t-1} = \mathbf{F}\hat{\mathbf{b}}_{t-1|t-1}.
\end{equation}
We then quantify the temporal consistency of the current 2D mask \(M_t\) by the box Intersection-over-Union (IoU) between \(\hat{\mathbf{b}}_{t|t-1}\) and the bounding box of \(M_t\),
\begin{equation}
  s_{\text{kf}} = \texttt{BoxIoU}(\hat{\mathbf{b}}_{t|t-1}, M_t).
\end{equation}
The Kalman update step,
\begin{equation}
  \hat{\mathbf{b}}_{t|t} = \hat{\mathbf{b}}_{t|t-1} + \mathbf{G}_t \big(M_t - \mathbf{H}\hat{\mathbf{b}}_{t|t-1}\big),
\end{equation}
with gain \(\mathbf{G}_t\) and observation model \(\mathbf{H}\), keeps the tracker locked to the current segmentation. 
Concurrently, we measure cross-modal geometric consistency by projecting the current 3D target cluster onto the image and computing the pixel-level IoU between the projected mask \(M^{\text{proj}}_t\) and the 2D mask \(M_t\),
\begin{equation}
  s_{\text{geo}} = \texttt{PixelIoU}(M^{\text{proj}}_t, M_t).
\end{equation}
For each candidate frame in the memory bank, we combine its temporal consistency and geometric alignment into a single keyframe score
\begin{equation}
  s_i = s_{\text{kf},i} + s_{\text{geo},i},
\end{equation}
where \(s_{\text{kf},i}\) and \(s_{\text{geo},i}\) are computed as above for frame \(i\). 
Instead of naively keeping the latest \(W_s\) frames, we select the top-\(W_s\) frames with the highest \(s_i\) as keyframes for SAM2's memory attention, favoring frames that are both temporally stable and geometrically representative for achieving robust 2D mask propagation.

Finally, the 3D cluster also serves as an anchor to detect and correct misidentification or drift in 2D segmentation over time. 
At each frame, we compare the current 2D mask \(M_t\) with the projection \(M^{\text{proj}}_t\) of the geometric target cluster via box IoU. 
If this score falls below a threshold \(\kappa_{\text{reinit}}\), we regard the two modalities as inconsistent, indicating either segmentation drift or corrupted clustering. 
In this case, we perform a cross-modal re-initialization: a subset of target points is selected from current target 3D cluster by farthest-point sampling (FPS) and projected to the image as virtual prompt points; these prompts are fed back to SAM2 with the original textual prompt to obtain a corrected mask, while the Kalman filter is re-initialized and geometric clustering is updated on the new result. 
Conversely, when the IoU stays above \(\kappa_{\text{reinit}}\), both visual and geometric estimates are considered reliable, and the geometric information is used solely for keyframe selection as described above. 
This bidirectional interaction—using 3D geometry to both repair and curate 2D memory—substantially improves the robustness of target grounding across viewpoints and over long-horizon flights.
The cross-refined 2D and 3D target regions with raw data are populated into a shared buffer, serving as the input for the prediction and planning stages.

\subsubsection{Simultaneous Localization and Mapping (SLAM)}
\label{subsubsec:slam}

Our aerial system achieves robust localization by fusing raw LiDAR point clouds with IMU data via a LiDAR-inertial odometry algorithm (\citealt{xu2022fast}).
To efficiently manage 3D information for facilitating downstream processes, we maintain two distinct representations of the environment: an Occupancy Grid Map (OGM) and a Euclidean Signed Distance Field (ESDF).
The OGM is updated in real-time by integrating newly localized LiDAR scans.
Based on preceding segmentation results, each occupied voxel is classified as either ``target'' or ``non-target''.
Concurrently, the ESDF is incrementally computed from the OGM to support efficient collision checking during trajectory optimization.

\subsection{Multi-modal Surface Prediction}
\label{subsec:prediction}

\begin{figure*}[t]
\centering
\includegraphics[width=0.99\linewidth]{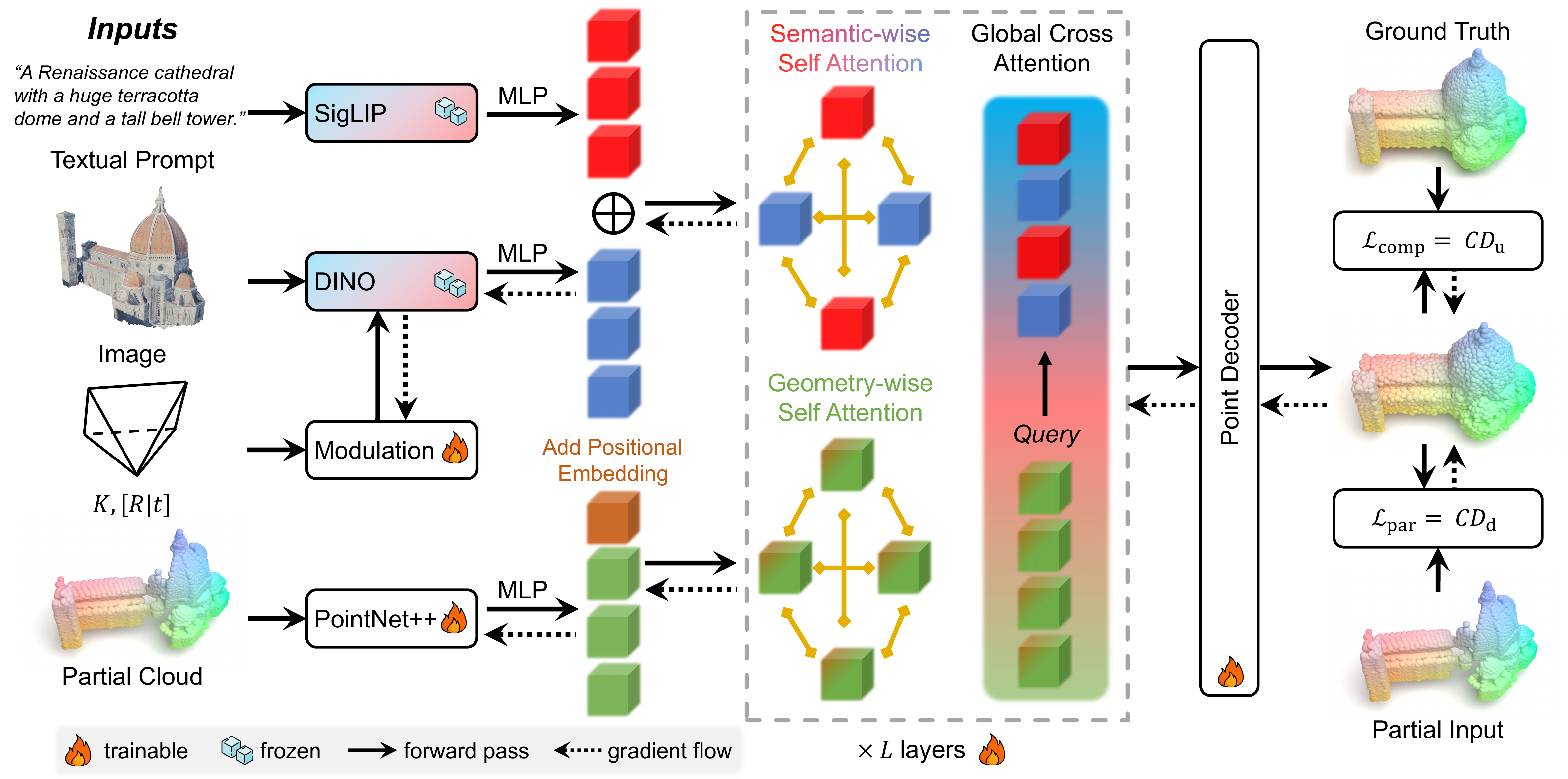}
\vspace{-0.3cm}
\caption{Overview of our multi-modal surface predictor.
It first independently encodes each modality with foundation models, fuses heterogeneous features via alternating attention, and decodes the completed shape supervised by joint partial and completion Chamfer losses.
}
\label{fig:PredOverview}
\vspace{-0.4cm}
\end{figure*}

This module answers ``\textit{what the full structure looks like}'' by predicting a complete 3D surface of the target from partial observations, empowering planning with comprehensive structural knowledge of the target.
Concretely, it consumes three observation sources: (1) segmented 3D points from the perception module, (2) the corresponding segmented RGB images, and (3) the original textual prompt describing the target's global semantics.
Relying on a single modality is fundamentally insufficient in our setting.
Purely geometric completion from LiDAR is severely constrained: as unordered 3D points without texture or rich semantic cues, 3D coordinates provide weak evidence about object parts, especially across large unobserved regions.
As a result, they tend to lean heavily on learned shape priors and often fail on out-of-distribution structures.
Conversely, vision and language FMs offer strong zero-shot generalization in isolation, but predictions obtained directly from a single visual or textual modality are geometrically ill-posed: without accurate metric scale or camera poses, they struggle to produce surfaces that match physical measurements.
Multi-modal fusion that exploits the complementarity among these signals is therefore essential to realize the ``\textit{bridging}'' role of prediction in our system (Sec. \nameref{subsubsubsec:modality}).
To this end, we design a multi-modal surface prediction network together with an efficient training-inference pipeline that jointly encodes and fuses geometric, visual, and textual features, effectively incorporating foundation-model knowledge to infer a complete 3D surface of the target with correct metric scale, fine-grained granularity, globally plausible shape, and local fidelity to the inputs, as depicted in Fig. \ref{fig:PredOverview}.

\subsubsection{Multi-modal Inputs and Preprocessing}
\label{subsubsec:network_input}

The geometric input $\mathcal{P}_{\text{tar}}$ is obtained by collecting all voxels labeled as ``target'' in the latest OGM, converting them into 3D points, and downsampling to a fixed set of $N_i$ points using FPS. 
For the visual input $\mathcal{I}_{\text{tar}}$, we project $\mathcal{P}_{\text{tar}}$ onto the image plane of each segmented RGB frame from the perception module and compute the fraction of projected points that fall inside the corresponding 2D mask.
The frame with the highest fraction is selected as $\mathcal{I}_{\text{tar}}$, ensuring that the most informative view of the target is used.
We also record the associated camera parameters $\mathbf{g}_{\text{tar}} = \mathbf{K}\big[\mathbf{R}_{\text{tar}} \mid \mathbf{t}_{\text{tar}}\big]$ for later use, where $\mathbf{K}$, $\mathbf{R}_{\text{tar}}$, and $\mathbf{t}_{\text{tar}}$ denote the intrinsic matrix, rotation matrix, and translation vector, respectively.
The textual input $h_{\text{text}}$ is directly taken from the user's original prompt.
To stabilize training and handle targets of varying physical scales, we apply a local normalization to both $\mathcal{P}_{\text{tar}}$ and $\mathbf{g}_{\text{tar}}$ using the centroid $\mathbf{c}_{\text{tar}}$ and scale $s_{\text{tar}}$ of $\mathcal{P}_{\text{tar}}$:
\begin{equation}
  \label{eq:scale}
  {\mathcal{P}}_{\text{tar}} = \frac{\mathcal{P}_{\text{tar}} - \mathbf{c}_{\text{tar}}}{s_{\text{tar}}}, \quad {\mathbf{t}}_{\text{tar}} = \frac{\mathbf{t}_{\text{tar}} - \mathbf{c}_{\text{tar}}}{s_{\text{tar}}}.
\end{equation}
For notational simplicity, we reuse $\mathcal{P}_{\text{tar}}$ and $\mathbf{g}_{\text{tar}}$ to denote the normalized point cloud and camera pose in the following.
Based on these inputs, our multi-modal predictor $f_{\theta}$ learns a mapping from $\{\mathcal{P}_{\text{tar}}, \mathcal{I}_{\text{tar}}, \mathbf{g}_{\text{tar}}, h_{\text{text}}\}$ to the complete surface $\mathcal{P}_{\text{comp}}$ of the target structure with $N_o$ points,
\begin{equation}
  f_{\theta}:(\mathcal{P}_{\text{tar}}, \mathcal{I}_{\text{tar}}, \mathbf{g}_{\text{tar}}, h_{\text{text}}) \mapsto \mathcal{P}_{\text{comp}} \in \mathbb{R}^{N_o \times 3}.
\end{equation}

\subsubsection{Network Architecture}
\label{subsubsec:network_architecture}

At a high level, our predictor $f_\theta$ adopts a feed-forward, post-fusion architecture with three stages: independent encoding of each modality, geometry-semantics fusion, and geometric decoding. 
The design is guided by two principles: (1) retain only minimal 3D inductive bias in the network, so that most structural priors come from ample 3D-annotated data and FMs rather than hand-crafted features; and (2) maximally reuse rich world knowledge from existing vision-language FMs, while explicitly steering this knowledge into the geometric stream that ultimately defines the 3D surface. 
Specifically, we let language and vision FMs first produce rich, task-agnostic embeddings, and defer multi-modal interaction to a late fusion stage where the geometric branch queries these embeddings through alternating self-cross attention layers. 
In this way, the fusion layer primarily aligns and retrieves semantic context for geometry, keeping the overall architecture simple while allowing the 3D predictions to benefit directly from the expressive latent spaces of FMs and enabling inference compatible with closed-loop flight.

\paragraph{Modality-specific encoders with FM reuse.}
Instead of training heavy encoders from scratch, we reuse pre-trained FMs to obtain compact latent representations for language and imagery.
In particular, we use the text encoder in SigLIP (\citealt{zhai2023sigmoid}) to derive a textual tokens $\mathcal{F}_\text{text} \in \mathbb{R}^{1 \times D_{\text{t}}}$ from the prompt $h_{\text{text}}$, and apply DINO (\citealt{oquab2023dinov2}) to the input image $\mathcal{I}_{\text{tar}}$, yielding a set of $K$ patchified tokens $\mathcal{F}_{\mathcal{I}} \in \mathbb{R}^{K \times D_{\mathcal{I}}}$.
This strategy avoids costly encoder training, accelerates convergence, and injects broad semantics about spatial structures into our predictor.

Given the current lack of powerful 3D FMs comparable to vision-language models, we deliberately adopt a lightweight PointNet++ backbone (\citealt{qi2017pointnet++}) to encode the partial point cloud $\mathcal{P}_{\text{tar}}$ into geometric features $\mathcal{F}_{\mathcal{P}} \in \mathbb{R}^{N_i \times D_{\mathcal{P}}}$. 
This design allows the geometric branch focus on capturing local shape patterns while high-level semantics and world priors are injected later through the fusion stage. 
All three feature sets are then projected to a common feature dimension $D_{\text{f}}$ using individual linear layers.

While vision FMs provide strong generalizable knowledge, their representations are largely spatially agnostic with respect to the metric 3D frame: the same image feature may correspond to very different physical scales or viewpoints. 
To mitigate this issue, we incorporate camera parameters $\mathbf{g}_{\text{tar}}$ into each layer of DINO via parameter-efficient feature modulation.
A small MLP predicts per-channel scale and shift terms
\begin{equation}
[\boldsymbol{\gamma}, \boldsymbol{\beta}] = \mathrm{MLP}(\mathbf{g}_{\text{tar}}),
\end{equation}
which modulate the visual features as
\begin{equation}
\mathcal{F}_{\mathcal{I}} \leftarrow (\mathbf{1} + \boldsymbol{\gamma}) \odot \mathcal{F}_{\mathcal{I}} + \boldsymbol{\beta},
\end{equation}
where $\odot$ denotes element-wise multiplication. 
This conditioning makes $\mathcal{F}_{\mathcal{I}}$ aware of 3D spaces with negligible additional parameters, easing subsequent alignment with geometric features.

\paragraph{Geometry-semantics fusion via alternating attention.}
To simplify multi-modal fusion among these heterogeneous features, we group them into two types: a geometric one $\mathcal{F}_{\mathcal{P}}$ and a semantic one obtained by concatenating visual and textual features,
\begin{equation}
\mathcal{F}_{\text{S}} = \mathcal{F}_{\mathcal{I}} \oplus \mathcal{F}_{\text{text}},
\label{eq:concat_vt}
\end{equation}
where $\oplus$ denotes concatenation along the feature dimension. 
The core of our fusion stage is the proposed alternating-attention layers, aligning two information sources into a shared latent space effectively.
Each layer consists of geometry-wise and semantic-wise self-attention followed by global cross-attention:
\begin{equation}
\begin{aligned}
\widehat{\mathcal{F}_\mathcal{P}} &= \mathrm{SelfAttention}(\mathcal{F}_\mathcal{P}), \\
\widehat{\mathcal{F}_{\text{S}}} &= \mathrm{SelfAttention}(\mathcal{F}_{\text{S}}), \\
\widetilde{\mathcal{F}_\mathcal{P}} &= \mathrm{CrossAttention}(\widehat{\mathcal{F}_\mathcal{P}}, \widehat{\mathcal{F}_\text{S}}).
\end{aligned}
\label{eq:attention_block}
\end{equation}
Geometry-wise self-attention captures intra-structural relations and details within the partial 3D shape, while semantic-wise self-attention aligns and consolidates prior knowledge coming from separate vision and language FMs.

Crucially, in global cross-attention, we treat the geometric features as query and the fused semantic features as key and value. 
This directional design reflects our modeling goal: the geometric features actively inherit world knowledge embedded in these FMs and draw upon the rich contextual information from semantic modalities.
This fashion allows the network to effectively ``grow'' a complete 3D shape from partially observed parts by absorbing complementary guidance from the image and text.
By stacking $L$ such layers, we obtain a deeply fused representation $\widetilde{\mathcal{F}_\mathcal{P}}$ that encodes both the local measured geometry and global semantic priors in a unified latent space.
Notably, Sec. \nameref{subsubsubsec:fusion} confirms that this fusion scheme effectively harvests FM knowledge into geometric predictions. 

\paragraph{Geometric decoding.} The final decoding stage maps the fused geometric feature back into 3D space. 
A MLP first predicts a set of candidate 3D points from $\widetilde{\mathcal{F}_\mathcal{P}}$. 
To preserve the characteristics of the input surface while extrapolating into unobserved regions, we concatenate these predicted points with the original partial input $\mathcal{P}_{\text{tar}}$ and apply farthest-point sampling to obtain a fixed-size output point cloud,
\begin{equation}
\mathcal{P}_\text{comp} = \mathrm{FPS}\big(\mathrm{MLP}(\widetilde{\mathcal{F}_\mathcal{P}}) \oplus \mathcal{P}_{\text{tar}}\big)
\in \mathbb{R}^{N_o \times 3},
\label{eq:decode_points}
\end{equation}
In this way, the resulting $\mathcal{P}_\text{comp}$ augments the physically measured surface with semantically guided completions and remains directly usable as a metric 3D representation.

\subsubsection{Training Objectives}
\label{subsubsec:training}

As a source of foresight for planning, the predictor is required to satisfy two complementary objectives. 
First, it must recover the complete surface of the target structure so that the planner can reason about occluded and yet-unseen regions and optimize global coverage. 
Second, it must remain strictly faithful to the partial observations provided by the LiDAR, preserving fine details on already observed surfaces and preventing the network from ``\textit{correcting}'' valid measurements to avoid guiding redundant scans.
In practice, many completion methods that optimize only for global completeness (for example, by minimizing a symmetric distance to the full ground truth (\citealt{yuan2018pcn,yu2021pointr,rong2024cra})) tend to hallucinate shapes that deviate from the input, a pathology that becomes particularly severe on complex or out-of-distribution structures.

Prior works, such as SDS-Complete (\citealt{kasten2023point}), enforce consistency with the input by maintaining an auxiliary implicit representation (\textit{e.g.}, a neural signed distance field) and constraining the partial observations to lie on its zero-level set. 
While effective, such schemes introduce expensive additional memory and computation overhead due to the extra representation that must be stored and updated during training. 
Instead, we adopt a simple yet effective dual-objective supervision strategy that encodes both completeness and input consistency directly at the point-set level, without any auxiliary fields (Sec. \nameref{subsubsubsec:partial_loss}).

Our formulation is built on the permutation-invariant Chamfer distance (CD), used in both undirected and directed forms (denoted by \(\mathrm{CD}_{\mathrm{u}}\) and \(\mathrm{CD}_{\mathrm{d}}\)), defined for point sets \(\mathcal{X}, \mathcal{Y} \subset \mathbb{R}^3\) as
\begin{equation}
\begin{aligned}
  \mathrm{CD}_{\mathrm{u}}(\mathcal{X}, \mathcal{Y}) 
  & = \frac{1}{|\mathcal{X}|}\sum_{x \in \mathcal{X}} \min_{y \in \mathcal{Y}} \|x-y\|_2^2 + \\
  & \frac{1}{|\mathcal{Y}|}\sum_{y \in \mathcal{Y}} \min_{x \in \mathcal{X}} \|x-y\|_2^2,
  \label{eq:ucd}
\end{aligned}
\end{equation}
\begin{equation}
    \mathrm{CD}_{\mathrm{d}}(\mathcal{X}, \mathcal{Y})
    = \frac{1}{|\mathcal{Y}|}\sum_{y \in \mathcal{Y}} \min_{x \in \mathcal{X}} \|x-y\|_2^2.
\end{equation}
To promote global completeness, we minimize the undirected CD between the prediction \(\mathcal{P}_\text{comp}\) and the ground-truth surface \(\mathcal{P}_\text{gt}\), termed as $\mathcal{L}_\text{comp} = \mathrm{CD}_{\mathrm{u}}(\mathcal{P}_\text{comp}, \mathcal{P}_\text{gt})$. 
To enforce faithfulness to measurements with negligible extra training burden, we add a partial-surface regularization term given by the directed CD from the prediction \(\mathcal{P}_\text{comp}\) to the input partial point cloud \(\mathcal{P}_\text{tar}\), termed as $\mathcal{L}_\text{par} = \mathrm{CD}_{\mathrm{d}}(\mathcal{P}_\text{comp}, \mathcal{P}_\text{tar})$. 
Our overall training loss function thus reads
\begin{equation}
    \mathcal{L} = \mathcal{L}_\text{comp} + \mathcal{L}_\text{par}.
    \label{eq:loss}
\end{equation}

\begin{figure}[h]
\centering
\includegraphics[width=0.99\linewidth]{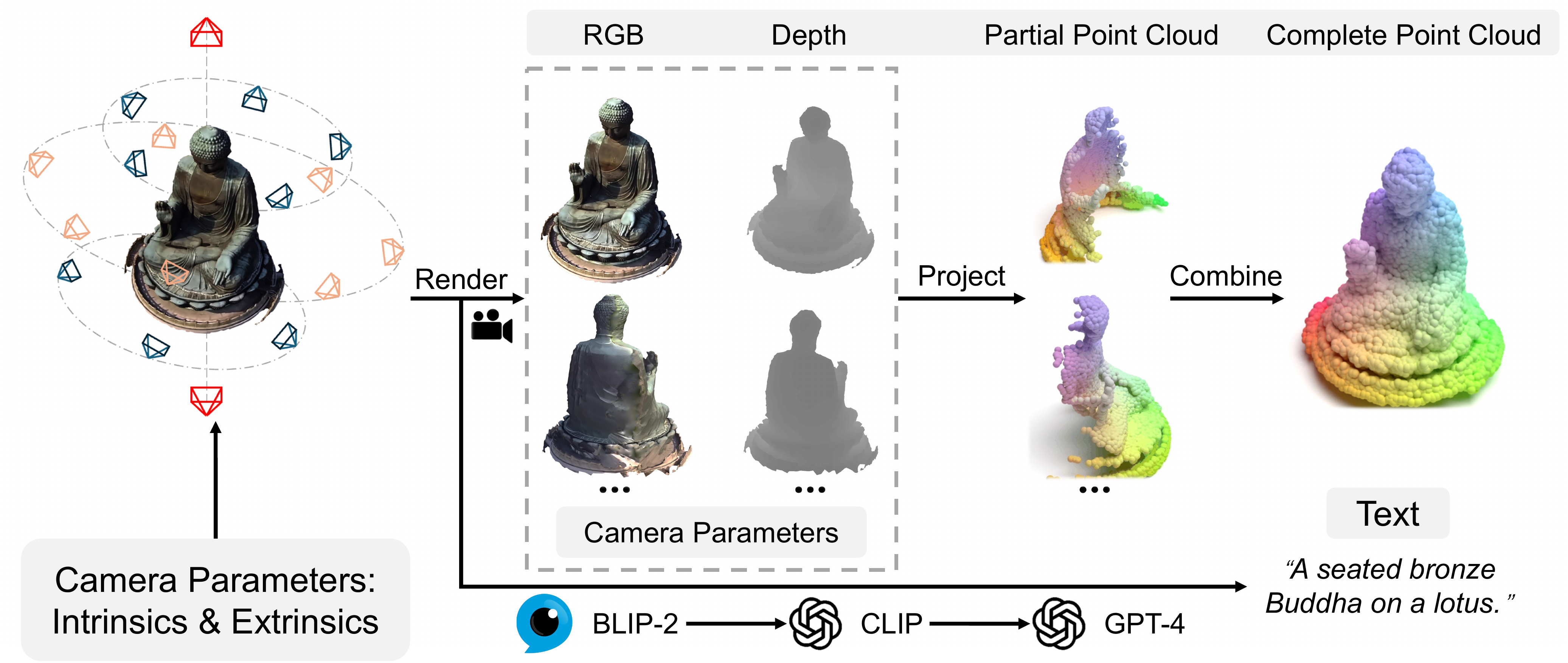}
\vspace{-0.3cm}
\caption{Automatic multi-modal data generation pipeline, which synthesizes partial point clouds, RGB images, and textual descriptions from the original 3D models.
}
\label{fig:DataGen}
\vspace{-0.4cm}
\end{figure}

\subsubsection{Automatic Multi-modal Data Generation}
\label{subsubsec:data_gen}

Training the proposed multi-modal predictor requires large quantities of aligned data across 3D geometry, images, camera poses, and text. 
However, existing datasets rarely provide all these modalities in a consistent form. 
We therefore build an automatic data generation pipeline supported by FMs (Fig. \ref{fig:DataGen}) that converts raw 3D assets into synchronized point clouds, RGB views, camera parameters, and textual descriptions at scale.

\paragraph{Rendering and point cloud synthesis.}
Given a 3D model, we use Blender\footnote{\url{https://www.blender.org/}} to render $32$ RGB-depth pairs from cameras placed uniformly on a viewing sphere around it, with distance equal to twice the radius of the model's bounding sphere and fixed intrinsics. 
Each depth map is back-projected into 3D to obtain a per-view point cloud. 
The complete ground-truth cloud $\mathcal{P}_\text{gt}$ is constructed by merging the points from all views.
To generate diverse partial observations, we repeatedly sample a random subset of $1$-$8$ viewpoints and fuse their points into a single input cloud $\mathcal{P}_\text{tar}$ in $32$ times.
This stochastic strategy yields a wide distribution of inputs, from very sparse to densely populated partial views, which is crucial for the model's robustness.

\paragraph{Text generation for global semantics.}
To attach a global textual prompt to each 3D asset, we build an automatic captioning tool on top of the rendered multi-view images.
For each rendered RGB view, a pre-trained image captioning model (\citealt{li2023blip}) produces five candidate sentences, and CLIP (\citealt{radford2021learning}) scores them against the corresponding image; the top-scoring one is kept as the view-level caption.
All view-level captions for a given asset are then passed to GPT-4 (\citealt{achiam2023gpt}), which fuses them into a single, globally consistent description by removing view-dependent details, merging redundancies, and normalizing style.
The resulting captions are short, single-sentence, model-centric descriptions that emphasize overall shape, appearance, and salient geometric structure, such as ``\textit{a stone Gothic church featuring a slender spire.}''
These consolidated sentences serve as the textual input $h_{\text{text}}$ used during training, and are designed to match the abstraction level of human prompts at test time.

\paragraph{Data augmentation.}
Each training sample is formed as an aligned tuple
$(\mathcal{P}_\text{tar}, \mathcal{I}_\text{tar}, \mathbf{g}_\text{tar}, h_{\text{text}}, \mathcal{P}_\text{gt})$. 
Here, $\mathcal{P}_\text{tar}$ and $\mathcal{I}_\text{tar}$ with its camera parameters $\mathbf{g}_\text{tar}$ are randomly selected. 
To enhance generalization, we apply several augmentations during training.
Variations in camera intrinsics are simulated by cropping the rendered RGB images—an efficient method that enhances robustness to different sensors without computationally expensive re-rendering. 
Robustness to camera pose noise is improved by applying random $\text{SE}(3)$ perturbations to the camera extrinsics, \textit{i.e.}, injecting small translations and rotations into the camera pose.
We further reduce sensitivity to the absolute world frame by randomly shifting the coordinate origin via a global translation of the input coordinates.
The diversity of the input is further enriched by applying a random masking operation to the partial point clouds, which varies their point density and spatial distribution.

Overall, this fully automatic pipeline generates large-scale, consistently aligned multi-modal data without manual labeling and can be readily extended to more 3D assets, forming a practical basis for future scaling of the predictor.

\begin{figure}[h]
\centering
\includegraphics[width=0.99\linewidth]{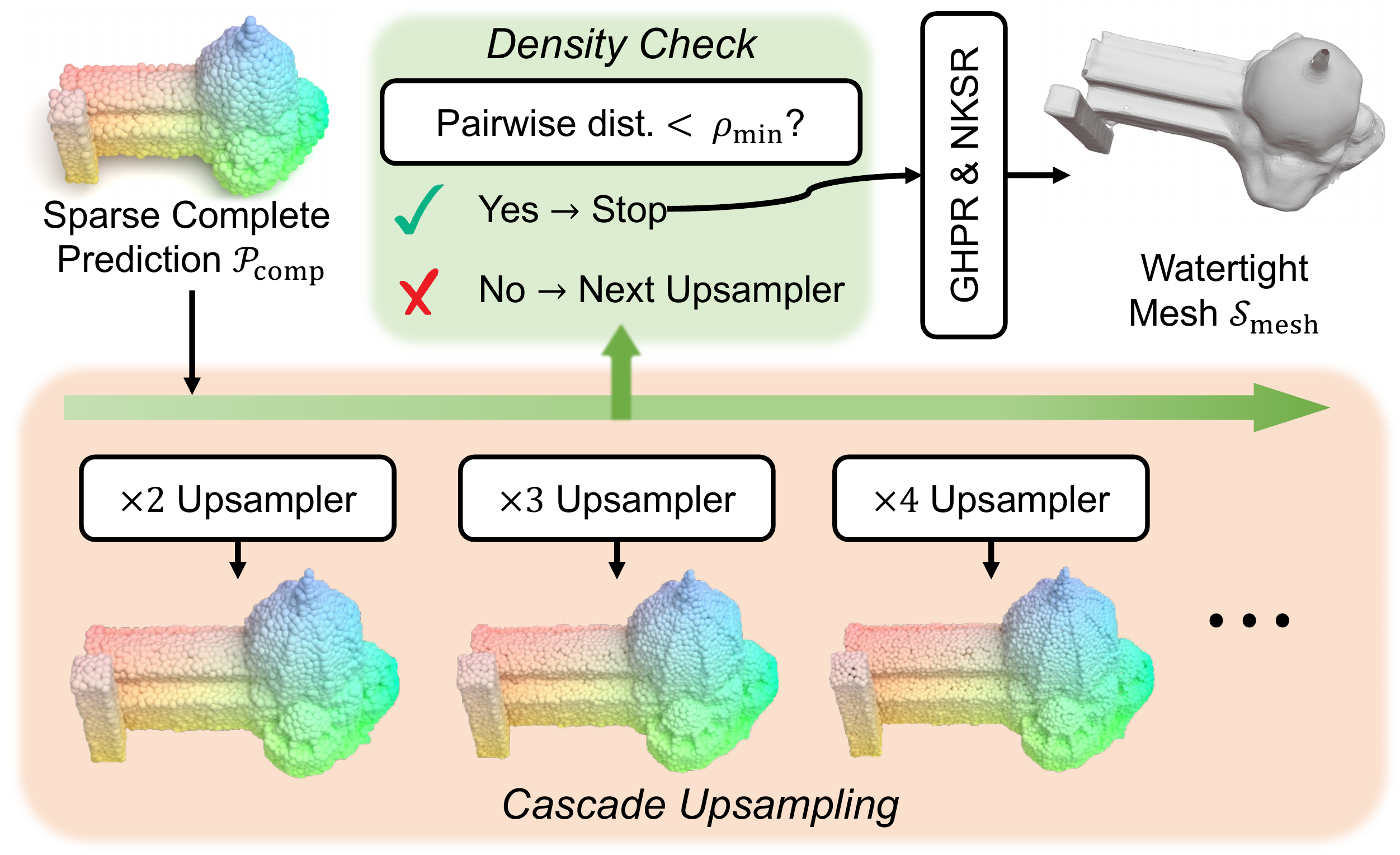}
\vspace{-0.3cm}
\caption{Inference-time Geometry Densification Scheme.
Starting from sparse completion, it applies cascade upsampling adaptive to physical granularity, then extracts a watertight mesh.
}
\label{fig:InferDense}
\vspace{-0.1cm}
\end{figure}

\subsubsection{Inference-time Geometry Densification}
\label{subsubsec:inference_time}

While the predictor already outputs a complete surface in the form of a fixed-size point set $\mathcal{P}_\text{comp}$, a single resolution cannot accommodate the wide range of target scales encountered in practice. 
For instance, a point budget that is sufficient to describe a small gate becomes too sparse for a large building facade, causing loss of fine details and, thereby hindering the provision of accurate guidance for planning.
Blindly increasing the network output size would inflate training and inference cost, and is unnecessary for smaller targets. 
Instead, we decouple prediction quality from network output resolution by introducing a lightweight, inference-time densification scheme, as shown in Fig. \ref{fig:InferDense}.

Concretely, we train a small family of independent upsampling networks that map $\mathcal{P}_{\text{comp}}$ to progressively denser point sets with $2N_o, 3N_o, \dots, kN_o$ points, using the same dual-objective supervision as the main predictor. 
At inference time, given the normalized prediction $\mathcal{P}_{\text{comp}}$, we apply these upsamplers in ascending order of density. 
For each upsampled result $\mathcal{P}^{(r)}_{\text{dense}}$, we rescale it back to the original metric frame by inversely applying the normalization in Eq. \eqref{eq:scale}, and compute the minimum pairwise distance between points. 
Once this spacing falls below a prescribed resolution threshold $\rho_\text{min}$, we regard the current density as sufficient for the target and stop the cascade, taking the corresponding point cloud as the final densified prediction $\mathcal{P}_{\text{dense}}^{*}$. 
This procedure provides scale-adaptive geometric detail while adding only modest computation and no extra supervision to the core predictor.

\begin{figure*}[t]
\centering
\includegraphics[width=0.99\linewidth]{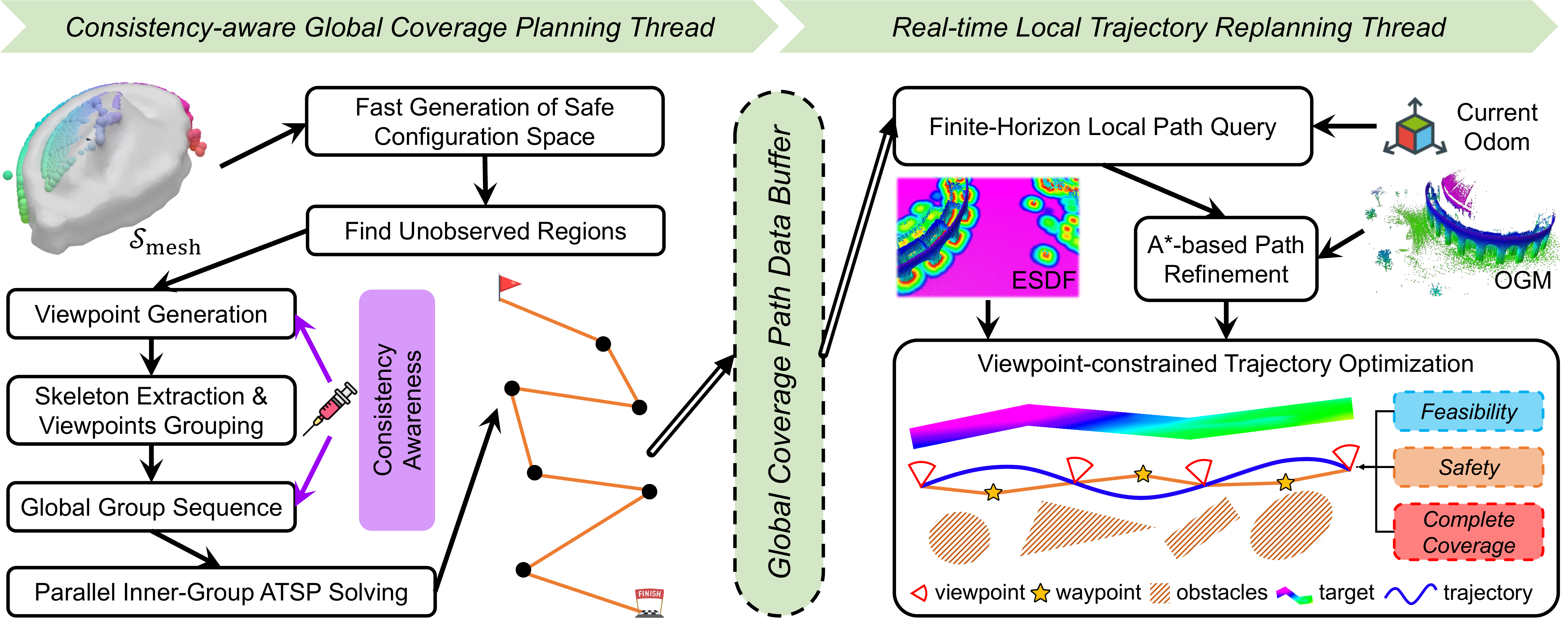}
\vspace{-0.2cm}
\caption{Overview of our prediction-aware hierarchical planner.
A consistency-aware global thread builds viewpoints over the predicted mesh and solves ATSP in parallel to populate a global coverage path buffer, meanwhile a real-time local thread concurrently generates minimum-time, feasible, and safe trajectories with complete target coverage.
}
\label{fig:PlanOverview}
\vspace{-0.4cm}
\end{figure*}

We then refine $\mathcal{P}_{\text{dense}}^{*}$ into a watertight mesh $\mathcal{S}_{\text{mesh}} = (V_{\text{mesh}}, E_{\text{mesh}})$ using NKSR (\citealt{huang2023neural}). 
Before meshing, we perform a standard hidden-point-removal operation (GHPR) (\citealt{katz2015visibility}) to suppress outliers and small floaters induced by noise, which improves the geometric reliability of the final surface. 
The resulting $\mathcal{S}_{\text{mesh}}$ is directly consumed by the planner, giving it a scale-appropriate, metrically consistent geometric representation without enlarging the main prediction network.

\subsection{Prediction-aware Hierarchical Planning}
\label{subsec:planning}

As the decision-making component of our system, the planner addresses ``\textit{how to scan}'' under predictive foresight, both efficiently and safely.
Although the perception-prediction stack transforms the original open-world problem into a clear 3D coverage task over a predicted target structure, three key difficulties prevent the direct application of off-the-shelf coverage planners (\citealt{cao2020hierarchical,wu2024uav,du2025efficient}).
First, the predicted geometry is inherently time-evolving: as new observations arrive, the completed surface is progressively refined.
Existing planners are typically designed as offline, one-shot solvers on static models and treat each updated prediction as an independent coverage problem, disregarding past plans.
Such snapshot-wise planning often yields paths that are locally optimal but temporally inconsistent, leading to detours or back-and-forth motions that inflate total flight time.
This calls for temporally consistent coverage planning, where updates to the predicted geometry induce smooth adjustments of the planned path rather than drastic reversals of flight intent.
Second, these methods frequently struggle to produce high-quality paths promptly due to their reliance on combinatorial optimization formulations.
It makes computation increasingly expensive for large targets, hindering timely adaptation to evolving predictions and potentially wasting flight time on regions derived from outdated geometry.
Therefore, rapid computation is essential for improving mission efficiency.
Third, even with prediction filling in unseen parts of the target, the surrounding environment remains previously unseen obstacles that may emerge at any time, requiring the planner to react with low latency to maintain safety. 
Yet existing solutions tightly couple coverage path generation and trajectory optimization, causing the update rate scale with the size of the target and undermining safe, agile operation in large, cluttered environments.

These challenges motivate our prediction-aware hierarchical planning framework, which decouples long-horizon objectives—information completeness and mission efficiency—from the immediate requirement of real-time safety while staying consistent and computationally efficient under the evolving predicted geometry.
The following subsections detail this framework and its two asynchronously executed planning layers.

\subsubsection{Planning Framework Overview}
\label{subsubsec:planning_framework}

As illustrated in Fig. \ref{fig:PlanOverview}, our planner is organized as a two-layer hierarchy that runs asynchronously at different rates.
A global layer computes coverage paths over the predicted target geometry at a deliberative rate $r_\text{G}$, and a high-rate local layer ($r_\text{L} \gg r_\text{G}$) continuously adjusts the executable trajectory using the online map while following the intent of the global path.

At the higher level, the global planner targets long-horizon efficiency and structural coverage.
At each planning cycle, it ingests the latest predicted target mesh $\mathcal{S}_\text{mesh}$ and rapidly constructs a globally efficient, target-centric coverage path $\mathbf{P}_\text{G}$.
This path is designed to (1) cover all yet-unobserved regions of the target with short travel distance, and (2) remain consistent with the historical flight to avoid unnecessary revisits and detours.
Specifically, the planner first generates a minimal set of safe viewpoints $\mathcal{V}_\text{G}$ that jointly observe all uncovered surface patches, and then efficiently determines a traversal order that approximately minimizes path length, inserting safe intermediate waypoints as needed to form a continuous $\mathbf{P}_\text{G}$.
Crucially, temporal-consistency awareness is incorporated throughout both stages, ensuring that successive refinements of $\mathcal{S}_\text{mesh}$ induce smooth evolutions of the coverage path.
Furthermore, we introduce two acceleration techniques to enhance computational tractability: fast safety checking for candidate viewpoints and a skeleton-based decomposition that enables parallelized path optimization.

At the lower level, the local planner handles real-time safety and dynamic trajectory adaptation.
At each update, it queries the most recent $\mathbf{P}_\text{G}$ and extracts the portion lying within a finite planning horizon $H$ ahead of the current drone position, denoted $\mathbf{P}_\text{L}$.
Using the current ESDF and OGM, $\mathbf{P}_\text{L}$ is first refined by the A* algorithm (\citealt{hart1968formal}) to ensure it lies entirely within the free space.
Afterwards, the local planner aims to convert $\mathbf{P}_\text{L}$ into a minimum-time, dynamically feasible, collision-free, and smooth trajectory in real-time. 
Meanwhile, this trajectory must also preserve the intended target coverage along $\mathbf{P}_\text{L}$ to reduce detours induced by backtracking motions (Sec. \nameref{subsubsubsec:vcto}).
To this end, a tailored viewpoint-constrained trajectory optimization is formulated, rigorously guaranteeing the above requirements while safeguarding real-time computability.
Finally, the resulting trajectory is transformed into low-level commands to a PID controller augmented with an online Extended State Observer (\citealt{han2009pid}) for robust execution.

This purpose-built planning framework directly addresses the challenges outlined above: it plans a temporally consistent global coverage path timely as the predicted structure is refined, and sustains low-latency responses to newly revealed obstacles regardless of target scale (Sec. \nameref{subsubsubsec:async}).

\begin{figure}[h]
\centering
\includegraphics[width=0.99\linewidth]{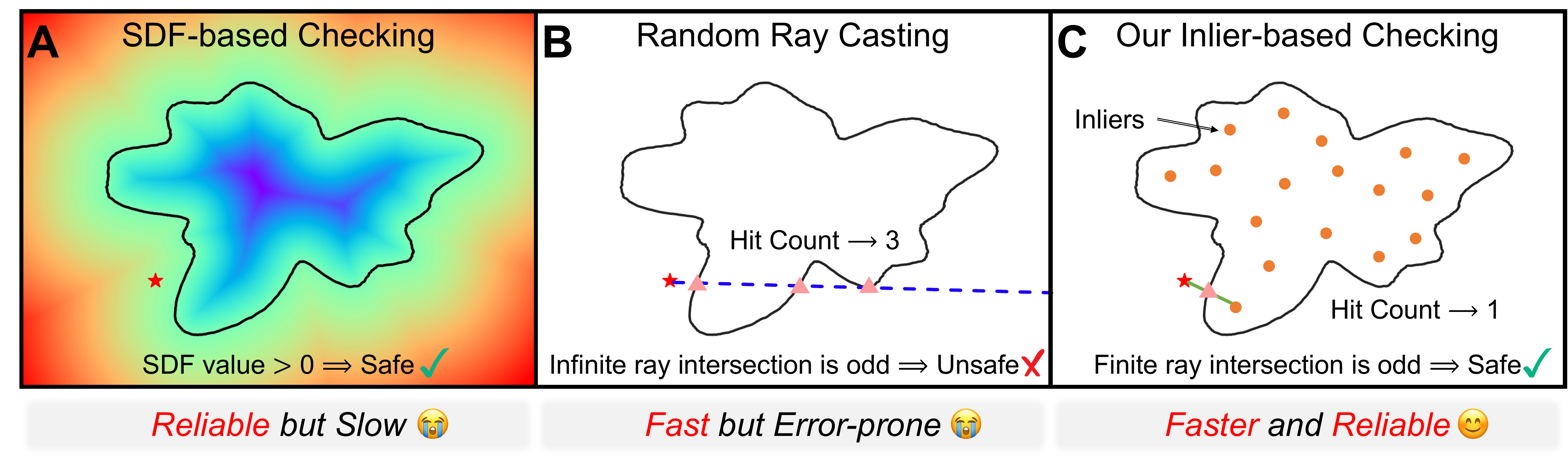}
\vspace{-0.3cm}
\caption{Illustration and comparison of safe-configuration checks in a 2D example: (A) SDF-based evaluation, (B) Ray casting, and (C) the proposed inlier-based method. The red star denotes the viewpoint candidate to be checked.
}
\label{fig:CfgGen}
\vspace{-0.3cm}
\end{figure}

\subsubsection{Consistency-aware Global Coverage Planning on Predicted Geometry}
\label{subsubsec:global_planning}
\par\vspace{0.6\baselineskip}\noindent

\paragraph{Fast generation of safe configuration space.}
\phantomsection
\makeatletter
\def\@currentlabelname{Fast generation of safe configuration space.}
\makeatother
\label{subsubsubsec:fast_gen}
Viewpoint generation first requires a safe configuration space to exclude viewpoints inside $\mathcal{S}_\text{mesh}$ (essential to avoid collision risks from surface penetration).
Conventional methods face a trade-off: SDFs are reliable but computationally costly for large structures, while faster ray casting algorithm suffers errors in complex non-convex geometries.
To reconcile safety and efficiency, our solution takes both strengths in two steps: 
(1) identifying inliers within $\mathcal{S}_\text{mesh}$ and (2) verifying viewpoint safety via ray intersection tests (Fig. \ref{fig:CfgGen}).

We use a cutting plane method to compute inliers of $\mathcal{S}_\text{mesh}$.
For efficiency, we downsample $\mathcal{S}_\text{mesh}$ vertices to a manageable number (\textit{e.g.}, 100) using FPS as representatives, then determine inliers for each.
For each representative, we first find an optimal cutting plane (its orientation $\mathbf{n}_{\text{opt}}$) minimizing angular variance relative to normals ($\mathcal{N}_{\text{v}}$) of neighboring vertices.
This is solved iteratively via quadratic programming:
\begin{equation}
	\mathbf{n}_{\text{opt}}^{i+1} = \mathop{\arg\min}_{\mathbf{n}_{\text{opt}} \in \mathbb{R}^3, ||\mathbf{n}_{\text{opt}}||_2=1} {\mathbf{n}_{\text{opt}}^{i}}^{\mathsf{T}} \text{Cov}^{i}(\mathcal{N}_{\text{v}}) \mathbf{n}_{\text{opt}}^{i},
	\label{eq:cutting_plane_orientation}
\end{equation}
where $i$ is the iteration index, and $\text{Cov}^{i}(\mathcal{N}_{\text{v}})$ is the covariance matrix of neighboring vertex normals for the current plane.
This quadratic problem is solved analytically via singular value decomposition, starting from a random $\mathbf{n}_{\text{opt}}^0$.
Upon $\mathbf{n}_{\text{opt}}$ convergence, the corresponding inlier $\mathbf{p}_{\text{inlier}}$ is calculated as:
\begin{equation}
	\mathop{\arg\min}_{\mathbf{p}_{\text{inlier}} \in \mathbb{R}^3} \sum\limits_{\mathbf{n}_{\text{v}} \in \mathcal{N}_{\text{v}}} ||(\mathbf{p}_{\text{inlier}} - \mathbf{p}_{\text{v}}) \times \mathbf{n}_{\text{v}}||_2^2,
	\label{eq:cutting_plane_inlier}
\end{equation}
where $\mathbf{p}_{\text{v}}$ is the position of neighboring vertices on the optimal plane.
Intuitively, $\mathbf{p}_{\text{inlier}}$ represents the closest confluence of lines through each normal in $\mathcal{N}_{\text{v}}$; this point is robustly positioned within $\mathcal{S}_\text{mesh}$ due to the mesh's closed-surface property.

Inliers are organized into a KD-tree for efficient querying.
For each viewpoint candidate, we retrieve its nearest inlier, cast a finite ray from itself to this inlier, and test intersections with $\mathcal{S}_\text{mesh}$.
Safety is determined by intersection parity: an odd count indicates the viewpoint is outside the mesh, while an even count indicates it is inside.
Sec. \nameref{subsubsubsec:cfg_gen} shows this method achieves reliable safety checks with high efficiency.

\paragraph{Viewpoint generation.}
We next generate a minimal viewpoint set $\mathcal{V}_{\text{G}}$ that ensures full coverage of all currently unobserved regions on $\mathcal{S}_\text{mesh}$.
Each viewpoint is parameterized by the same five DoF as in Sec.~\nameref{sec:problem_statement}, 
$\mathbf{vp} = \{p_x,p_y,p_z,\theta,\phi\}$.
We first construct a dense pool of viewpoint candidates.
For each unobserved vertex on $\mathcal{S}_\text{mesh}$, we compute its normal and sample two candidates by translating the vertex along the normal in both directions by a user-defined distance.
Using the fast safety evaluation described above, we discard any candidate that lies inside $\mathcal{S}_\text{mesh}$ or violates safety, and retain the remaining ones as feasible viewpoints.
This dense set is then pruned using a gravitation-like model that merges viewpoints that cover fewer regions into those covering more.
We sort all candidates by their visible vertex counts $vc_i$ in descending order.
Processing them in this order, for each viewpoint $i$ we collect its neighbors $\mathcal{V}_{\text{N}}$ within a given radius, and then update its position according to a coverage-weighted ``gravitational'' pull:
\begin{equation}
  \mathbf{p}_i \gets \mathbf{p}_i + \sum\limits_{\mathbf{p}_n \in \mathcal{V}_{\text{N}}} \frac{vc_n}{vc_i} (\mathbf{p}_n-\mathbf{p}_i), \, \text{s.t.} \, vc_n < vc_i.
\end{equation}
Pitch and yaw are updated in the same way.
All neighbors in $\mathcal{V}_{\text{N}}$ are then removed from the candidate set.
We repeat the above procedure until all unobserved surface regions on $\mathcal{S}_\text{mesh}$ are covered, and the remaining viewpoints constitute the minimal set $\mathcal{V}_{\text{G}}$.

\begin{figure}[h]
\centering
\includegraphics[width=0.99\linewidth]{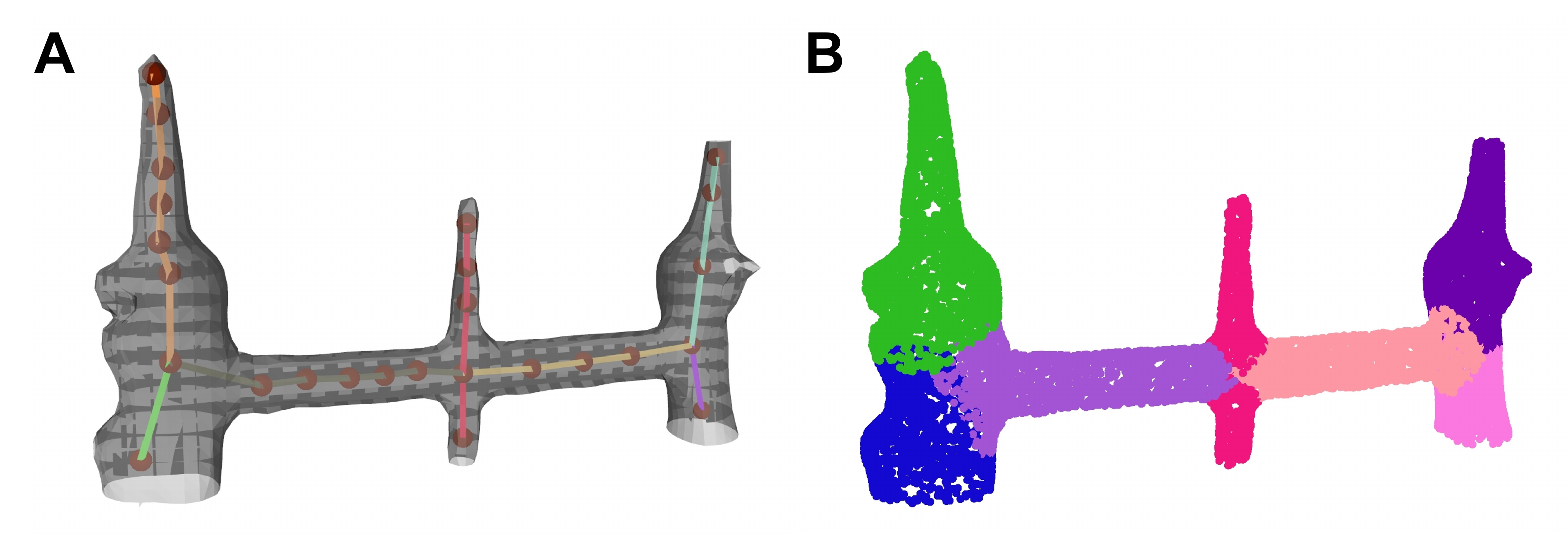}
\vspace{-0.3cm}
\caption{Illustration of skeleton-based space decomposition in castle gate. (A) Input mesh and extracted skeleton, (B) Space decomposition using the skeleton.
}
\label{fig:Skeleton}
\vspace{-0.4cm}
\end{figure}

\paragraph{Route determination in parallel.}
As discussed above, directly determining the optimal visiting order of $\mathcal{V}_{\text{G}}$ is computationally expensive due to its NP-hard nature, especially for large-scale target structures.
To make the problem tractable, we adopt a two-level decomposition strategy that splits the original problem into smaller, efficiently solvable sub-problems.

First, we extract a 3D skeleton from the previously computed inliers to capture the topology of $\mathcal{S}_\text{mesh}$ (Fig. \ref{fig:Skeleton}A).
Based on this skeleton, the target mesh is decomposed into multiple independent subspaces characterized by simple geometry (Fig. \ref{fig:Skeleton}B), following our prior work (\citealt{feng2024fc}).
Each viewpoint in $\mathcal{V}_{\text{G}}$ is then assigned to its nearest subspace according to Euclidean distance, yielding a set of disjoint viewpoint groups.
Unlike our previous approach, which plans directly within each subspace, we further refine this grouping to balance computational load.
Specifically, for any subspace whose viewpoint group spans a radius larger than a predefined threshold $R_\text{g}$, we construct a visibility graph where edges connect pairs of viewpoints with unobstructed line-of-sight (\textit{i.e.}, no occlusion by $\mathcal{S}_\text{mesh}$).
The viewpoints are then partitioned into approximately uniform, inter-visible convex groups, each with radius below $R_\text{g}$.
This decomposition reduces the original route planning problem into two Asymmetric Traveling Salesperson Problems (ATSPs): (1) optimizing the group sequence; (2) finding the optimal traversal order of viewpoints within each group.
Both problems are solved using a mature ATSP solver (\citealt{helsgaun2000effective}) and can be processed in parallel across groups.

This two-level strategy offers two key advantages: (1) it significantly reduces problem dimensionality and enables inherent parallelization (\textit{e.g.}, decomposing a single $100\times100$ problem to four parallel $25\times25$ problems). (2) Group convexity obviates the need for collision checks among intra-group viewpoints.
While not guaranteeing strict global optimality, it empirically achieves a favorable balance between computational efficiency and path quality.

\begin{figure}[h]
\centering
\includegraphics[width=0.99\linewidth]{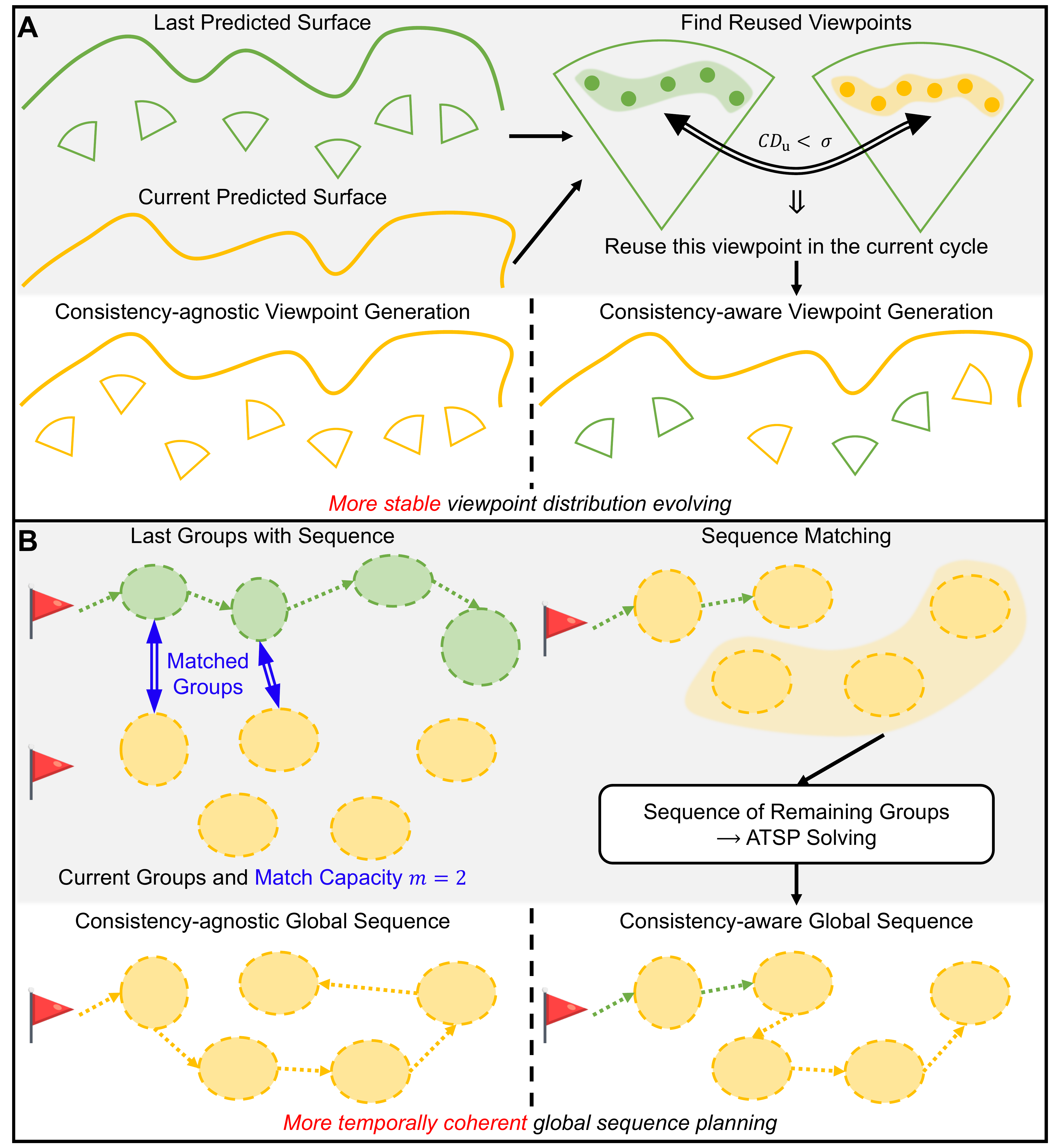}
\vspace{-0.3cm}
\caption{Illustration of consistency-promoting designs in global planning: (A) Reuse prior viewpoints for stable coverage across cycles; (B) Match prior groups to yield a temporally coherent global sequence.
Shown in 2D for clarity.
}
\label{fig:PlanConsistency}
\vspace{-0.4cm}
\end{figure}

\paragraph{Injecting consistency awareness.}
As previously emphasized, the global path $\mathbf{P}_{\text{G}}$ should also align with historical flight.
Thus, we integrate simple yet effective consistency-promoting designs into each step of the global planning process, as shown in Fig. \ref{fig:PlanConsistency}.
First, for unobserved regions whose geometry remains unchanged since the last planning cycle, we preferentially reuse viewpoints from the previous set to minimize temporal variations in the viewpoint distribution (Fig. \ref{fig:PlanConsistency}A).
Before current viewpoint generation, we evaluate viewpoint reuse by comparing past and current observations:
\begin{equation}
    \text{ReuseVP}(\mathbf{vp}) = \left\{
    \begin{array}{rcl}
        \text{True},  &  & {\text{CD}_{\text{u}}(\mathcal{O}^{\text{last}}_\mathbf{vp}, \mathcal{O}^{\text{cur}}_\mathbf{vp}) < \sigma}\\
        \text{False}, &  & {\text{else}}
    \end{array} \right.
\end{equation}
where $\sigma$ is the maximum of the average edge lengths of the previous and current meshes; $\mathcal{O}^{\text{last}}_\mathbf{vp}$ and $\mathcal{O}^{\text{cur}}_\mathbf{vp}$ are the last and current observations for viewpoint $\mathbf{vp}$. 

\begin{algorithm}[t]
\caption{Consistency-aware Global Coverage Planning}
\label{alg:global_plan}
\begin{algorithmic}[1]
\Require Current predicted mesh $\mathcal{S}_\text{mesh}^{\text{cur}}$
\Statex \hspace*{-\algorithmicindent}\textbf{Notation:} Inliers $\mathbf{I}$, skeleton $\mathcal{Q}$, viewpoint set $\mathcal{V}_{\text{G}}$, viewpoint groups $\mathcal{G}_\mathcal{V}$, group centroids $\mathcal{E}$, group sequence $Seq_{\text{G}}$, global path $\mathbf{P}_{\text{G}}$, reused viewpoints $\mathcal{V}_{\text{R}}$, matched sequence $Seq_{\text{M}}$
\State $\mathbf{I} \gets \text{ConfigurationSpaceGeneration}(\mathcal{S}_\text{mesh}^{\text{cur}})$
\If{$\mathcal{V}_{\text{G}}^{\text{last}} \neq \emptyset$}
  \State $\mathcal{V}_{\text{R}} \gets \text{FindReusedViewpoints}(\mathcal{V}_{\text{G}}^{\text{last}}, \mathcal{S}_\text{mesh}^{\text{cur}}, \mathcal{S}_\text{mesh}^{\text{last}}, \mathbf{I})$
\Else $\mathcal{V}_{\text{R}} = \emptyset$
\EndIf
\State $\mathcal{V}_{\text{G}}^{\text{cur}} \gets \text{ViewpointGeneration}(\mathcal{V}_{\text{R}},\mathcal{S}_\text{mesh}^{\text{cur}},\mathbf{I})$
\State $\mathcal{Q} \gets \text{SkeletonExtraction}(\mathbf{I})$
\State $\mathcal{G}_\mathcal{V}^{\text{cur}}, \mathcal{E}^{\text{cur}} \gets \text{ViewpointGrouping}(\mathcal{V}_{\text{G}}^{\text{cur}},\mathcal{E}^{\text{last}},\mathcal{Q})$
\If{$Seq_{\text{G}}^{\text{last}} \neq \emptyset$}
  \State $Seq_{\text{M}}, \mathcal{G}_\mathcal{V}^{\text{cur}} \gets \text{SequenceMatching}(Seq_{\text{G}}^{\text{last}}, \mathcal{G}_\mathcal{V}^{\text{cur}})$
\Else $Seq_{\text{M}} = \emptyset$
\EndIf
\State $Seq_{\text{G}}^{\text{cur}} \gets \text{FindGroupSequence}(Seq_{\text{M}}, \mathcal{G}_\mathcal{V}^{\text{cur}})$
\State $\mathbf{P}_{\text{G}} \gets \text{ParallelATSPSolving}(Seq_{\text{G}}^{\text{cur}},\mathcal{G}_\mathcal{V}^{\text{cur}})$
\Ensure $\mathbf{P}_{\text{G}}$
\end{algorithmic}
\end{algorithm}

Second, to bolster temporal consistency in viewpoint clustering, group centroids from the preceding iteration serve as initial seeds for the current clustering.
Third, to avoid erratic flight from unstable group visiting sequences, a matching step (Fig. \ref{fig:PlanConsistency}B) is added before invoking the ATSP solver: we align the first $m$ groups from the historical sequence with current clustering results, via:
\begin{equation}
	\mathop{\arg\min}_{g^{\text{cur}} \in \mathcal{G}_{\mathcal{V}}^{\text{cur}}} \text{CD}_{\text{d}}(g^{\text{last}}, g^{\text{cur}}),
	\label{eq:consistency_seq}
\end{equation}
where $\mathcal{G}_{\mathcal{V}}^{\text{cur}}$ is the current cycle's groups, and $g^{\text{last}}$ is a group from the previous sequence.
This matching is done sequentially for each of the first $m$ historical groups; once matched, the current group is removed from $\mathcal{G}_{\mathcal{V}}^{\text{cur}}$ to prevent duplicate assignments.
Collectively, as summarized in Algo. \ref{alg:global_plan}, the global planner integrates these careful designs to yield a temporally consistent global path $\mathbf{P}_{\text{G}}$ with complete coverage, ready for the local planner to access. 
Sec. \nameref{subsubsubsec:consistency} demonstrates the efficiency improvement contributed by consistency-aware designs.

\subsubsection{Viewpoint-constrained Trajectory Optimization}
\label{subsubsec:local_traj}

For simultaneous optimization of spatial configuration and temporal allocation, the local trajectory $\mathbf{x}_{\text{L}}(t)$ is parameterized via a differentiable piecewise polynomial class, MINCO (\citealt{wang2022geometrically}):
\begin{equation}
	\mathbf{x}_{\text{L}}(t) = \left\{z_i(t) = \mathbf{c}_i^{\mathsf{T}}\psi(t) \in \mathbb{R}^5, t \in [0, T_i] | i = 1, \dots, K \right\},
	\label{eq:traj_param}
\end{equation}
\begin{equation}
	\mathbf{c}_i \in \mathbb{R}^{(2s-1) \times 5}, \psi(t) = \left[1, t, t^2, \dots, t^{2s-1}\right]^{\mathsf{T}},
	\label{eq:traj_basis}
\end{equation}
where the trajectory comprises $K$ polynomial pieces.
The coefficients matrix $\mathbf{c}_i$ and temporal basis function $\psi(t)$ depend on the polynomial degree $d$.
MINCO reformulates $\mathbf{c}_i$ via a differentiable mapping $\mathcal{M}(\cdot)$ using intermediate points $\mathbf{Q} = \left\{\mathbf{q}_i \in \mathbb{R}^5 | i = 0,1,\dots,K \right\}$ (on the refined $\mathbf{P}_{\text{L}}$) and time profiles $\mathbf{T} = \left\{T_i | i = 1,\dots,K \right\}$:
\begin{equation}
	\mathbf{c}_i = \mathcal{M}(\mathbf{q}_i, T_i).
	\label{eq:traj_coeff}
\end{equation}
This enables efficient spatio-temporal trajectory deformation by manipulating $\mathbf{Q}$ and $\mathbf{T}$, highly advantageous for gradient-based optimization.
For brevity, we split $\mathcal{M}(\cdot)$ into positional ($\mathcal{M}^{\text{pos}} \in \mathbb{R}^{(2s-1) \times 3}$) and orientational ($\mathcal{M}^{\text{ori}} \in \mathbb{R}^{(2s-1) \times 2}$) components:
\begin{equation}
	\mathbf{c}_i = \mathcal{M}^{\text{pos}}(\mathbf{q}_i, T_i) \oplus \mathcal{M}^{\text{ori}}(\mathbf{q}_i, T_i),
	\label{eq:traj_coeff_split}
\end{equation}
\begin{equation}
	z_i^{\text{pos}}(t) = \mathcal{M}^{\text{pos}}(\mathbf{q}_i, T_i)\psi(t), z_i^{\text{ori}}(t) = \mathcal{M}^{\text{ori}}(\mathbf{q}_i, T_i)\psi(t),
	\label{eq:traj_split}
\end{equation}
generating coefficients for 3D position and orientation (\textit{i.e.}, pitch and yaw angles), respectively.

We then formulate the trajectory optimization problem as:
\begin{subequations}\label{eq:traj_opt}
\begin{align}
  \mathop{\min}_{\mathbf{Q}, \mathbf{T}} \;\;
  & \sum_{i=1}^{K} \int_0^{T_i} \|(z_i^{\text{pos}}(t))^{(d)}\|_2^2 \, dt
    + w_{\text{t}} T_i, \label{eq:traj_opt_obj}\\
  \text{s.t. } \;\;
  & -T_i < 0,\; -\Phi(z_i^{\text{pos}}(t)) < 0, \label{eq:traj_opt_cons1}\\
  & g_{\text{dyn}}(z_i^{\text{pos}}(t), z_i^{\text{ori}}(t)) < 0, \label{eq:traj_opt_cons2}\\
  & C(\mathbf{x}_{\text{L}}(t)) - C(\mathbf{P}_{\text{L}}) = 0. \label{eq:traj_opt_cons3}
\end{align}
\end{subequations}
The integral term promotes smoothness via the squared $d$-th derivative of position, and $w_{\text{t}} T_i$ penalizes total duration with weight $w_{\text{t}}$.
The inequality constraints ensure positive time allocation, collision avoidance (positive ESDF values $\Phi(\cdot)$), and dynamic feasibility (bounded velocity, acceleration, jerk, and angular velocity).
Eq. \eqref{eq:traj_opt_cons3} encodes coverage: ideally, the executed trajectory $\mathbf{x}_{\text{L}}(t)$ should achieve the same target coverage as the input path $\mathbf{P}_{\text{L}}$.
However, the coverage functional $C(\cdot)$ is a global, non-differentiable visibility measure, and enforcing \eqref{eq:traj_opt_cons3} directly would lead to a highly non-convex mixed-integer problem, unsuitable for real-time optimization.
Fortunately, by construction of the global coverage path, full coverage over the current horizon is achieved precisely by aggregating observations from all viewpoints embedded in $\mathbf{P}_{\text{L}}$.
This allows us to replace the abstract coverage constraint \eqref{eq:traj_opt_cons3} with a concrete geometric requirement: the optimized trajectory must traverse every viewpoint in $\mathbf{P}_{\text{L}}$.
Since $\mathbf{P}_{\text{L}}$ includes both viewpoints and waypoints, we decompose $\mathbf{Q}$ into $W$ optimizable waypoints ($\mathbf{Q}_{\text{{w}}} \in \mathbb{R}^{W \times 5}$) and $G$ fixed viewpoints ($\mathbf{Q}_{\text{{v}}} \in \mathbb{R}^{G \times 5}$).
$\mathbf{Q}$ is constructed using binary selection matrices $\mathbf{S}_{\text{{w}}}$ and $\mathbf{S}_{\text{{v}}}$:
\begin{equation}
	\mathbf{Q} = \mathbf{S}_{\text{{w}}}\mathbf{Q}_{\text{{w}}} + \mathbf{S}_{\text{{v}}}\mathbf{Q}_{\text{{v}}} \in \mathbb{R}^{(W+G) \times 5},
	\label{eq:traj_opt_var}
\end{equation}
where $\mathbf{S}_{\text{{w}}}$ and $\mathbf{S}_{\text{{v}}}$ satisfy Kronecker Delta Function $\delta_{i, \text{idx}(i)}$ ($\text{idx}(\cdot)$ denotes the index of a point (waypoint or viewpoint) on the local path).
This ensures $\mathbf{x}_{\text{L}}(t)$ passes through fixed viewpoints while optimizing via waypoints.
To ensure strictly positive $T_i$ without explicit inequality constraints, we introduce unconstrained virtual variables $\mathcal{T} = \{\tau_i\}$ and map them to $T_i$ through a smooth diffeomorphism $\Theta(\cdot)$:
\begin{equation}
  T_i = \Theta(\tau_i) =
  \begin{cases}
    1 + \tau_i + \dfrac{\tau_i^2}{2}, & \tau_i \geq 0,\\[4pt]
    \dfrac{2}{2 - 2\tau_i + \tau_i^2}, & \tau_i < 0,
  \end{cases}
  \label{eq:traj_opt_time}
\end{equation}
which yields $T_i > 0$ for all $\tau_i \in \mathbb{R}$.

Finally, we convert the inequality constraints to soft penalties by Smooth L1 loss terms and obtain an unconstrained objective:
\begin{equation}
  \mathop{\min}_{\mathbf{Q}_{\text{w}}, \mathcal{T}}
    \mathcal{J}_{\text{s}}
    + w_{\text{t}}\mathcal{J}_{\text{t}}
    + w_{\text{d}}\mathcal{J}_{\text{d}}
    + w_{\text{c}}\mathcal{J}_{\text{c}}
    + w_{\text{sc}}\mathrm{Cov}(\mathbf{Q}_{\text{w}})
    + w_{\text{tc}}\mathrm{Cov}(\mathcal{T}).
  \label{eq:traj_opt_obj_final}
\end{equation}
Here, $\mathcal{J}_{\text{s}}$ corresponds to the smoothness objective and $\mathcal{J}_{\text{t}}$ penalizes total flight time, while $\mathcal{J}_{\text{d}}$ and $\mathcal{J}_{\text{c}}$ are Smooth L1 penalties that softly enforce dynamic feasibility and collision avoidance.
The covariance regularizers $\mathrm{Cov}(\mathbf{Q}_{\text{w}})$ and $\mathrm{Cov}(\mathcal{T})$ discourage spatially oscillatory control points and highly uneven time allocation across segments, improving numerical stability.
The positive scalars $w$ are weighting coefficients that balance these terms.
We subsequently solve this unconstrained problem efficiently to produce the desired $\mathbf{x}_{\text{L}}(t)$ by employing a gradient-based approach, notably the L-BFGS algorithm (\citealt{liu1989limited}).

\section{Implementation Details}
\label{sec:implementation}

\subsubsection{Perception} We use the official small version of SAM2 as the segmentation FM to meet real-time requirements ($\sim2$ Hz), and set the sliding-window memory bank capacity to $W_m = 30$.
In cross-modal refinement, $W_s = 7$ keyframes are kept for memory attention and the reinitialization threshold is set to $\kappa_{\text{reinit}} = 0.7$.
For 3D mapping, we refer to an efficient volumetric method (\citealt{han2019fiesta}) with a voxel resolution of $0.1$ m and $10$-Hz update.

\begin{figure}[h]
\centering
\includegraphics[width=0.99\linewidth]{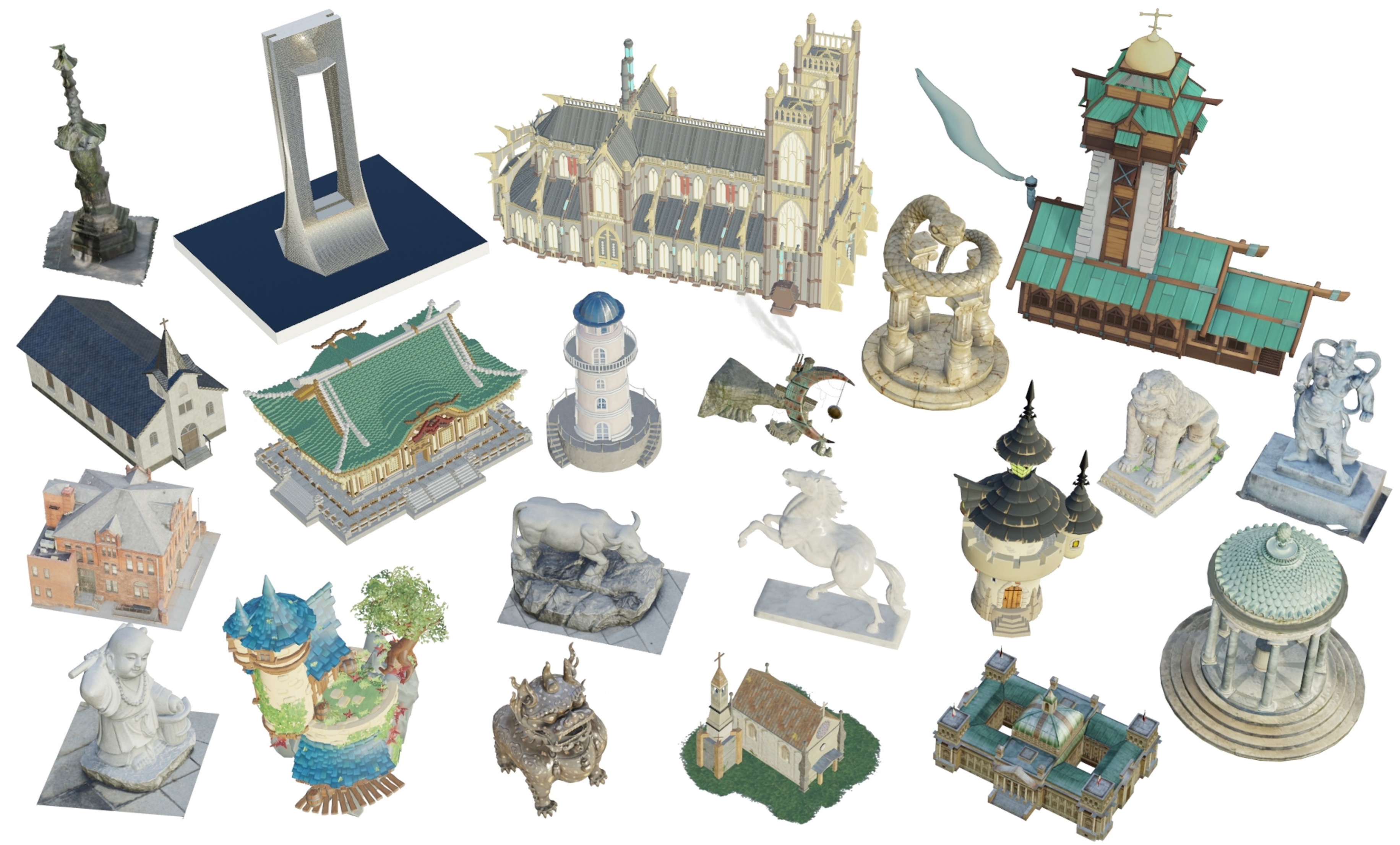}
\vspace{-0.3cm}
\caption{A representative sample of the diverse, high-fidelity 3D assets used for multi-modal surface predictor training.
}
\label{fig:Assets}
\vspace{-0.3cm}
\end{figure}

\subsubsection{Prediction} For the main predictor, the geometric input and output are fixed to
$N_i = 2048$ and $N_o = 2048$ points, respectively.
The DINO image encoder produces $K = 324$ patch tokens per view, and
feature dimensions are set to $D_{\text{t}} = 768$,
$D_{\mathcal{I}} = 768$, and $D_{\mathcal{P}} = 512$.
All modalities are projected to a shared fusion space of
dimension $D_{\text{f}} = 512$.
In the fusion stage, we stack $L = 10$ alternating-attention layers as described.
The training dataset is constructed from a diverse collection of approximately $70,000$ high-fidelity distinct 3D models, including a curated subset of Objaverse (\citealt{deitke2023objaverse}) and a small number of additional real-world scans from the Internet.
Fig. \ref{fig:Assets} showcases a representative sample of these models, illustrating the dataset's broad coverage of object categories and scales.
Each asset is processed by our automatic multi-modal pipeline to produce aligned tuples as training samples.
We implement the predictor in PyTorch and train on a server with eight NVIDIA RTX4090 GPUs, dual Intel Xeon CPUs, and $512$ GB RAM.
We use Distributed Data Parallel with a per-GPU batch size of $42$ (effective batch size $336$), automatic mixed precision, and activation checkpointing to improve throughput and reduce memory usage.
The model is trained for $50$ epochs (about $72$ hours) using AdamW with an initial learning rate of $1\times10^{-4}$ and weight decay of $1\times10^{-4}$.
A cosine-annealing-with-warm-restarts scheduler restarts every epoch and decays the learning rate to a minimum of $1\times10^{-5}$.
In inference-time geometry densification, we use Snowflake Point Deconvolution in SnowflakeNet (\citealt{xiang2021snowflakenet}) as the base upsampling architecture, with upsampling factor $k = 10$ and resolution threshold $\rho_{\text{min}} = 0.5$ m in all experiments.
These upsamplers share the same training settings as the predictor but are trained for fewer epochs.

\subsubsection{Planning} For the hierarchical planner, the global and local layers run in separate threads with $r_{\text{G}} = 1$ Hz and $r_{\text{L}} = 20$ Hz, respectively.
In global planning, the viewpoint grouping radius is $R_\text{g} = 8$ m and we match the first $m = 2$ groups in the historical sequence when performing sequence alignment.
The local planning horizon is set to $H = 10$ m along the current global path and the MINCO parameterization uses $d = 4$ (\textit{i.e.}, snap) with one polynomial segment per edge of the A*-refined $\mathbf{P}_\text{L}$.
The penalty weights in Eq. \eqref{eq:traj_opt_obj_final} are chosen as
$w_{\text{t}} = 60.0$, $w_{\text{d}} = 500.0$, $w_{\text{c}} = 500.0$, $w_{\text{sc}} = 60.0$, and $w_{\text{tc}} = 80.0$, which we empirically found to give a good trade-off among smoothness, flight time, dynamic feasibility, and safety.
Optimization is solved with a maximum of $100$ iterations and a gradient tolerance of $10^{-5}$.

\section{Experiments}
\label{sec:experiments}

In this section, we present a comprehensive evaluation of \textbf{F}ly\textbf{C}o in both real-world and simulated settings.
Our experiments are organized to answer three complementary questions:
\begin{itemize}
  \item[1)] \textit{Practicality:} Can the system reliably and precisely execute fully autonomous, prompt-grounded target structure scanning missions in diverse in-the-wild environments under real sensing, computation, and constraints?
  \item[2)] \textit{Performance:} Under fair conditions, how does \textbf{F}ly\textbf{C}o compare with representative SOTA paradigms in terms of efficiency, data integrity, safety, run-to-run stability, and required human priors?
  \item[3)] \textit{Attribution:} Which components and design choices in perception, prediction, and planning account for improvements in overall system performance?
\end{itemize}
We address the first question through extensive field deployments in Sec. \nameref{subsec:real_world_exp}.
Afterwards, Sec. \nameref{subsec:comparisons} demonstrates that \textbf{F}ly\textbf{C}o represents a more advanced system architecture, achieving a superior efficiency-completeness trade-off with high reliability while requiring substantially lower-effort human priors.
Lastly, thorough ablation experiments in Sec. \nameref{subsec:ablation_studies} confirm the individual contributions of each key module and design in our system.

\begin{figure}[t]
\centering
\includegraphics[width=0.99\linewidth]{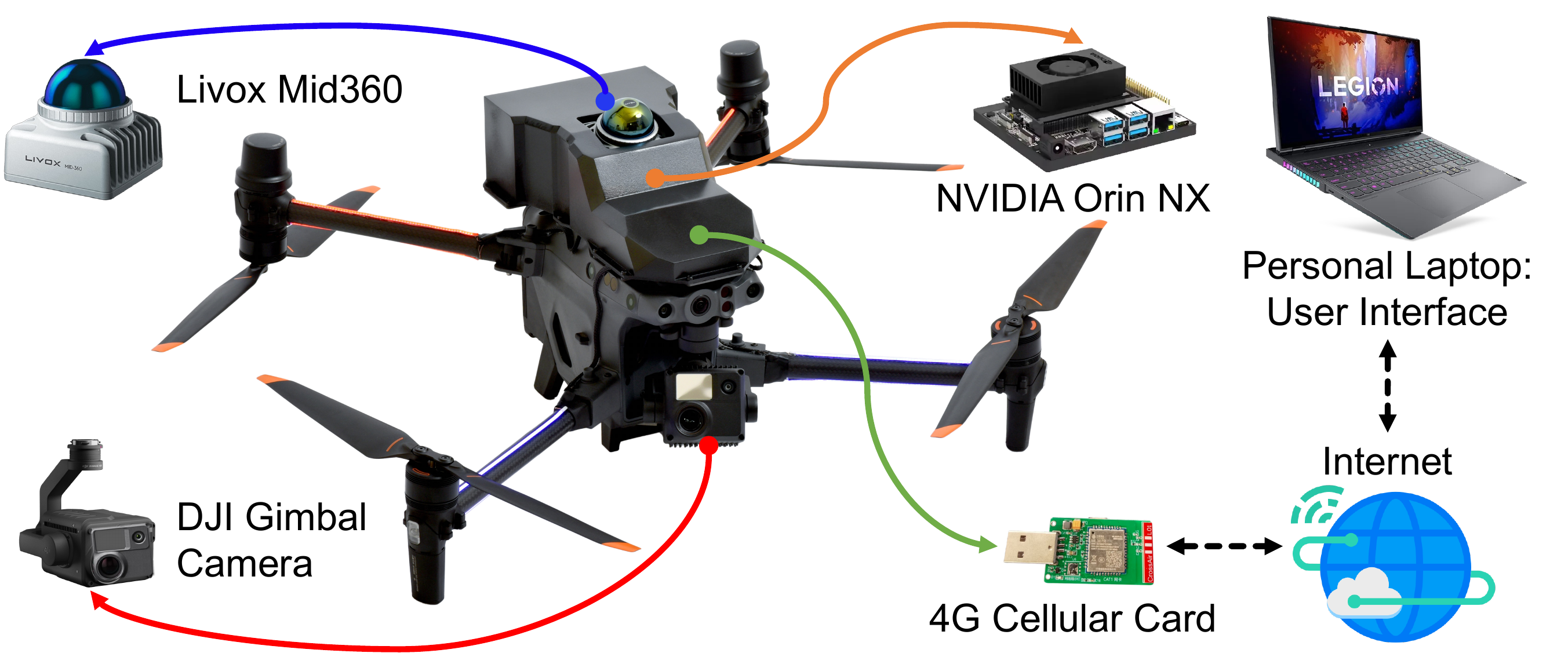}
\vspace{-0.3cm}
\caption{Hardware platform and communication setup for real-world flights.
}
\label{fig:Hardware}
\vspace{-0.3cm}
\end{figure}

\subsection{Real-world Experiments}
\label{subsec:real_world_exp}

Here we aim to substantiate that \textbf{F}ly\textbf{C}o effectively leverages FMs to enable drones autonomously and accurately scan human-specified 3D structures in diverse open-world environments, relying only on intuitive, low-effort human prompts.
We present three proof-of-concept yet challenging experiments—conducted across four diverse in-the-wild sites—that demonstrate precise scene understanding, high efficiency, real-time safety, and reliable adaptability.
In all trials, our system accomplished the missions in a single attempt without re-flights or additional operator intervention, despite receiving only high-level human inputs, underscoring its practical viability and real-world applicability.
The rest of this part first describes the hardware platform and experimental setup, then details the three representative scanning experiments, and finally profiles the real-time performance and computational load of each module.

\begin{figure*}[t]
\centering
\includegraphics[width=0.99\linewidth]{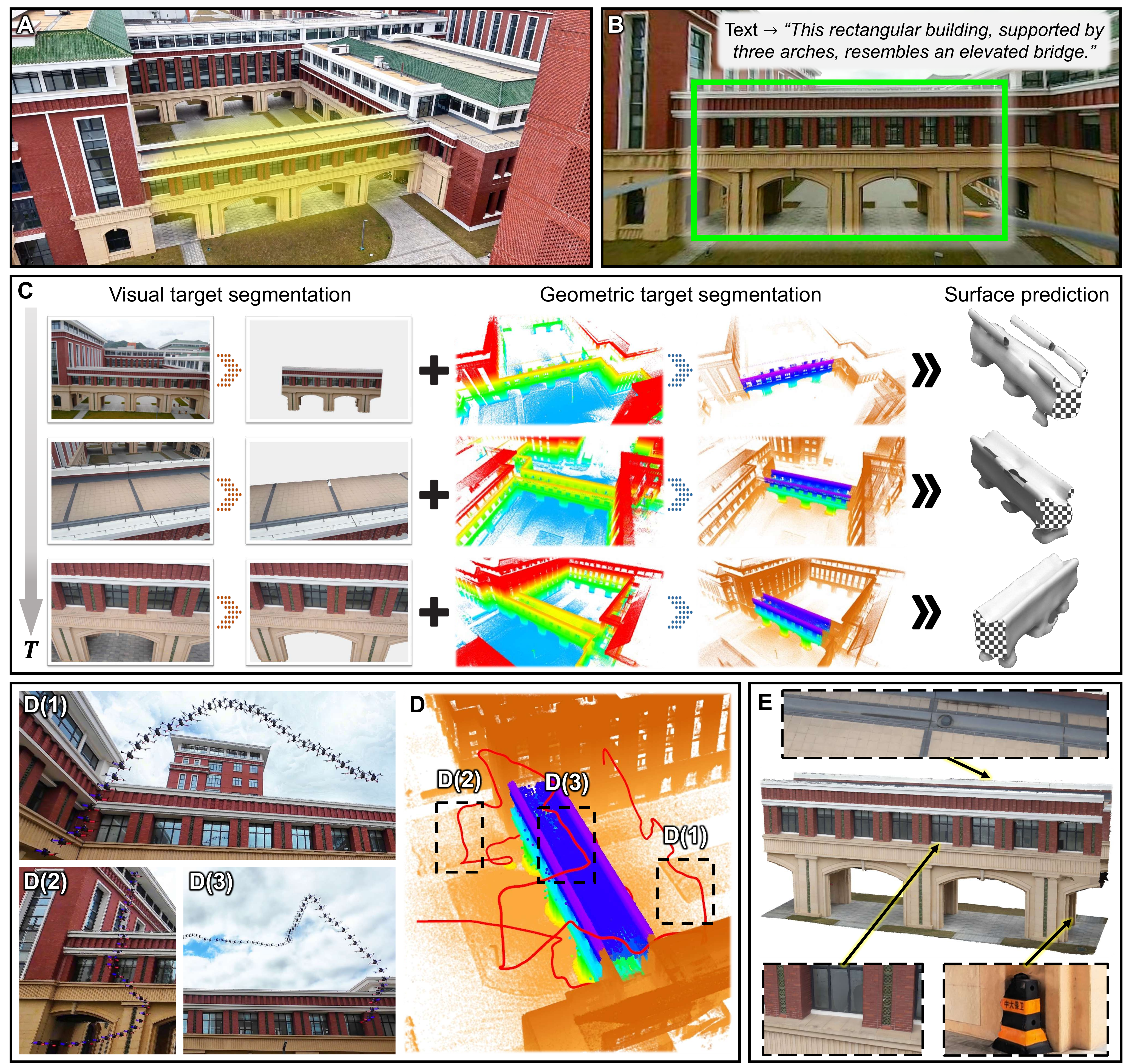}
\vspace{-0.3cm}
\caption{Precise scene understanding for adaptive scanning.
(A) The target structure of human interest: an arch bridge spanning between two large buildings.
(B) Human-inputted visual and textual prompts for target specification.
(C) Left$\rightarrow$right: precise visual/geometric target segmentation and reasonable surface prediction at three timestamps, even under the challenges of similar texture and spatial connection. 
(D) Demonstration of adaptively completing the scanning mission solely around the target and three corresponding real-world snapshots of this scanning flight shown in D(1-3).
(E) Targeted 3D reconstruction with fine-grained details focusing only on the human-specified arch bridge.
}
\label{fig:PerceptionReal}
\vspace{-0.5cm}
\end{figure*}

\subsubsection{Hardware Platform and Experimental Setup}
\label{subsubsec:hardware}

As the drone shown in Fig. \ref{fig:Hardware}, the aerial platform of \textbf{F}ly\textbf{C}o is modified from a DJI M30 quadcopter (\citealt{djim30}).
Following our customization, it possesses a takeoff weight of $4.5$ kg, a $668$ mm wheelbase, and a $0.55$ m radius.
For sensing, we equip this drone with a Livox Mid360 3D LiDAR (\citealt{livoxlidar}) and a DJI gimbal-mounted camera.
The former provides a maximum sensing range of $70$ m and a point acquisition rate of $200,000$ Hz for dense geometric scanning.
Its integrated inertial measurement unit (IMU) provides $50$ Hz updates, which are crucial for accurate localization and mapping.
Concurrently, the latter captures $1920 \times 1080$ resolution RGB images at 30 frames per second and supports a pitch rotation range from $+45\,^\circ$ to $-100\,^\circ$, crucial for flexible textural details capture.
Onboard computation is handled by an NVIDIA Jetson Orin NX (\citealt{orinnx}) that includes an Ampere GPU and an Arm Cortex-A78AE v8.2 64-bit CPU.
Remaining structural and electronic components adhere to a lightweight and compact design principle, collectively contributing to a flight endurance of approximately $24$ minutes on a single charge.
In addition, a personal laptop, featuring an Intel 24-core 5.4-GHz i9-13900HX CPU and an NVIDIA RTX4060 GPU, serves as the ground station, offering an intuitive interface for specifying user prompts and visualizing the flight in real-time.
Since the prediction module is computationally intensive for the edge device but not strictly latency-critical (empirically below $1$ s is acceptable), it is executed on this laptop, while all other modules run on the onboard computer.
Communication between onboard and ground devices is maintained through a 4G cellular network, augmented by data compression techniques to ensure real-time and stable data transmission.

The maximum speed and gimbal angular speed of the drone are limited to $1.0$ m/s and 20 $^\circ$/s, respectively, to balance perception quality and flight efficiency.
No prior site information or pre-built environmental maps about the experimental scenarios are provided to the drone.
The only human input is a one-sentence textual description together with sparse visual annotations at the start of each mission, where no manual intervention is allowed during flight.

\subsubsection{Precise Scene Understanding for Adaptive Scanning}
\label{subsubsec:scene_understanding}

We first validate our system's ability to reliably understand the physical world, a capability enabled by foundation models and fundamental to translating abstract user prompts into accurate drone scanning actions.
A key aspect of this understanding mirrors human perception: humans not only identify discrete structures but effortlessly recognize key parts (\textit{e.g.} a house's front door or a teacup's handle) (\citealt{hoffman1984parts, biederman1987recognition}), focusing on these meaningful components.
This ``part attention'' demands more nuanced scene understanding—including fine-grained part-whole relationships—than recognizing standalone structures.
Consequently, adapting to scan specific parts of a larger structure is far more challenging than scanning entire ones, particularly when the target is visually interconnected with its surroundings.
This raises a critical question: Can drones develop precise scene understanding to adaptively scan user-specified targets, even when they are integral parts of larger, interconnected entities?
To answer this, we performed a real-world experiment in which the system was instructed to scan a distinctive arch bridge connecting large buildings on both sides (target highlighted in yellow, Fig. \ref{fig:PerceptionReal}A).

With our system, the user was only required to pick the target arch bridge by two simple steps: first, drawing a rough bounding box around the desired region in the camera feed; second, providing a simple text description that explicitly mentions the arch bridge (Fig. \ref{fig:PerceptionReal}B).
Even though the inputs were not very precise, and the bridge looked visually similar to the nearby walls, our system still successfully segmented the target structure in the 2D image (Fig. \ref{fig:PerceptionReal}C, first two columns). 
This achievement derives from the fusion of visual and textual understanding of the target from FMs: these modalities together established a reliable semantic reference, effectively clearing up ambiguities caused by imprecise user inputs and visual similarity.
Next, this detailed 2D segmentation helped accurately split the 3D point cloud using projective geometry.
This step cleanly and stably separated the bridge from the connected buildings it attached to, despite their physical adjacency (Fig. \ref{fig:PerceptionReal}C, middle two columns).
Crucially, boosted by 2D-3D cross-modal refinement, our perception module effectively transfers the world understanding of FMs into 3D space, enabling the drone to continuously track the target bridge throughout the flight, even as the drone's viewpoint shifted dynamically (Fig. \ref{fig:PerceptionReal}C). 
This process of linking the user's request to the real-world target stayed stable and consistent throughout the entire mission.

Drawing on these accurate perception results, our predictor inferred the complete surface of the arch bridge, even when some of its pillars were occluded from view by others (Fig. \ref{fig:PerceptionReal}C, right column). 
An initial deviation in the height estimation of the upper walkway was progressively reduced as the drone collected more observations during flight. 
Building on this precise scene understanding and predictive foresight, the planner then generated adaptive, target-centric scanning paths that compactly encircled the bridge, precluding time waste on non-target regions (Fig. \ref{fig:PerceptionReal}D).
During flight, it adapted trajectories in real-time to safely avoid obstacles—for instance, navigating between bridge pillars, bypassing nearby walls and buildings—while simultaneously ensuring efficient, comprehensive coverage of the specified target structure.

Once completing its missions, the drone transmitted all visual and geometric data it had scanned from the arch bridge to the user.
This raw data was subsequently processed with RealityScan (\citealt{rc}) to reconstruct a high-fidelity, metrically accurate 3D model, even for such a complex structure (Fig. \ref{fig:PerceptionReal}E).
The final deliverables to the user included both the raw data and the reconstructed model.
A full recording of this experiment is provided in Extension 2.

\subsubsection{Robust Scanning for Large-scale Structures}
\label{subsubsec:robust_large_scale}

\begin{figure*}[t]
\centering
\includegraphics[width=0.99\linewidth]{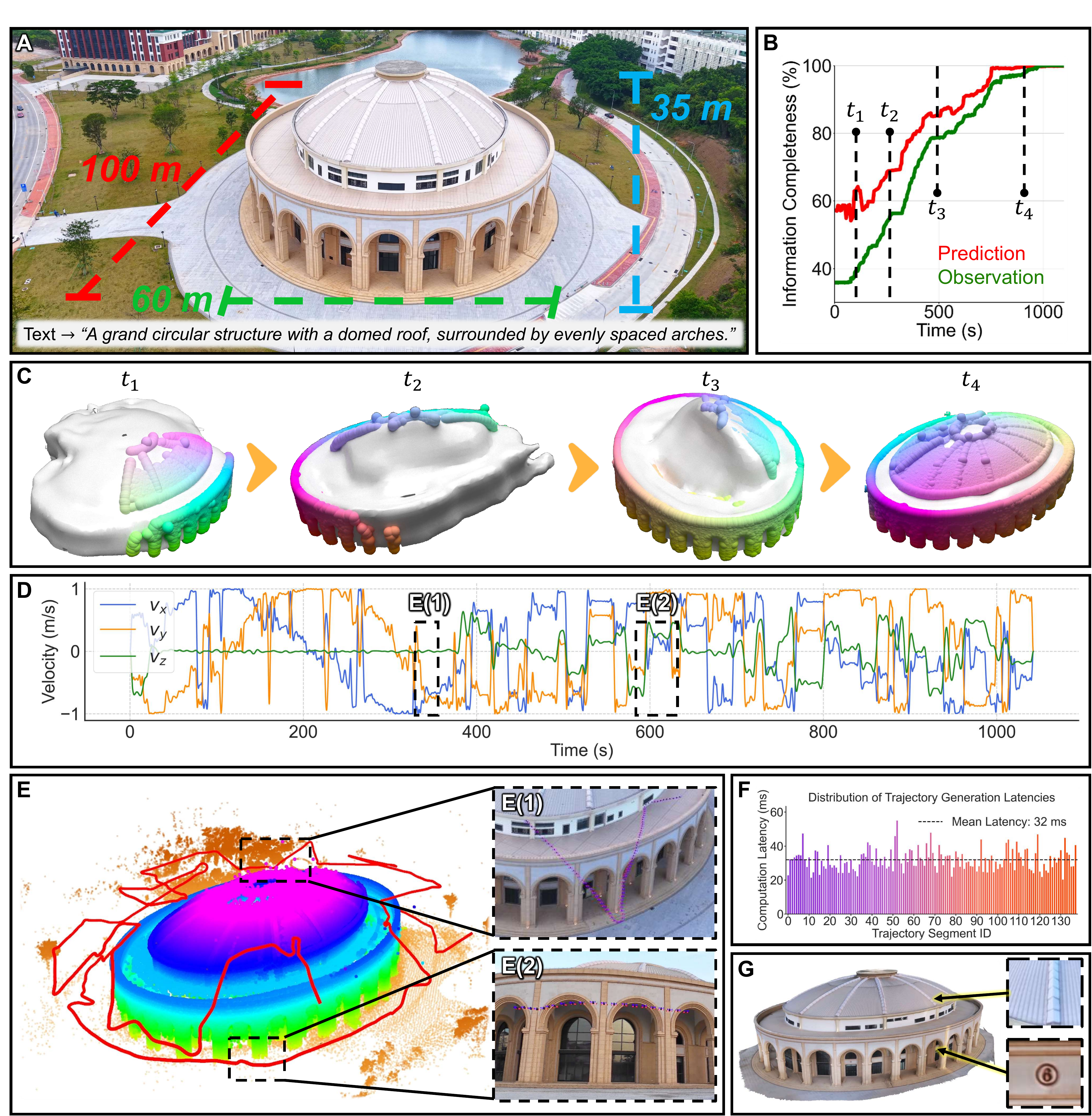}
\vspace{-0.3cm}
\caption{Robust scanning for large-scale structures.
(A) Bird's-eye view of the experiment site—large-scale concert hall (100 m $\times$ 60 m $\times$ 35 m) and the textual description entered by the user.
(B) Information completeness curve over time. The plot compares practical observations (green) against model predictions (red). The prediction model's positive information gain exhibits its effectiveness in providing foresight guidance for planning.
(C) The predicted geometry tightly aligns with partial observations and progressively converges towards the actual shape.
(D) The velocity profile of this large-scale mission.
(E) The scanning flight trajectory (red) around the concert hall. The rainbow-colored surface represents the geometric segmentation results. (E(1-2)) Two close-up views from the trajectory at key locations.
(F) Distribution of trajectory generation latencies during the mission.
(G) A resulting high-quality, photorealistic 3D model reconstructed after this autonomous scanning flight.
}
\label{fig:PredictionReal}
\vspace{-0.5cm}
\end{figure*}

In our system, knowledge embedded in FMs is leveraged for the generalizable prediction of the target surface, forming a pivotal bridge between perception and planning. 
This prediction, built on scene understanding, infers the target's complete 3D geometry to inform efficient, safe scanning paths.
However, operating in large-scale environments markedly challenges this module: due to limited sensor range, only a small portion of the target is visible at the beginning of the mission, and even later observations remain sparse relative to the overall structure.
This impairs accurate, consistent surface prediction over time.
When prediction becomes unreliable, the foresight guidance for aerial robots is compromised, leading to back-and-forth movements that waste more time or even dangerous flight paths. 
By distilling world knowledge from FMs, \textbf{F}ly\textbf{C}o strengthens spatial reasoning across diverse scenarios, thereby enabling generalizable 3D structure predictions (detailed quantitative evaluations are presented in Sec. \nameref{subsubsubsec:zero-shot}).
This capability allows the drone to maintain robust predictions, overcoming the challenges posed by limited observations in large-scale scenes.
To demonstrate this robustness in practice, we deployed our system at a $100$ m $\times$ $60$ m $\times$ $35$ m concert hall surrounded by hills and lakes (see Fig. \ref{fig:PredictionReal} and Extension 3). 
The mission covered $1028.92$ m in $1042.4$ s (Fig. \ref{fig:PredictionReal}D), which, to the best of our knowledge, is the first demonstration of a continuous autonomous flight exceeding $1000$\,m for 3D target structure scanning in the wild.

`Information completeness' in Fig. \ref{fig:PredictionReal}B measures the proximity to the complete surface of the concert hall over time.
The two curves show that throughout the flight, the prediction module consistently provided more useful information compared to the direct observations from the drone.
Notably, early on ($t_1$), even though only about $35\%$ of the target was initially observed, our predictor correctly recovered the concert hall's overall shape and key structural features. 
It maintained a decent shape completeness of around $63\%$ and significantly increased the available information by about $28\%$.
Such robustness stems from the model's ability to leverage global contextual information embedded in the user's text description (``\textit{a circular structure with a domed roof}''), which compensates for the scarcity of initial visual and geometric inputs.
Fundamentally, this reflects that the predictor inherits generalizable knowledge from FMs, enabling it to generate reasonable structural predictions even with limited other modalities data.
As the drone collected more observations, the model kept steadily improving its inference, becoming less uncertain over time and gradually converging on the correct shape (Fig. \ref{fig:PredictionReal}B, C).

Guided by these robust predictions, the scanning trajectory was notable for being smooth and coherent, with no frequent direction changes, unnecessary detours, or repeated visits to areas that had already been scanned (Fig. \ref{fig:PredictionReal}E). 
This well-organized path led to strong flight efficiency. 
The system also performed well in terms of computational efficiency in trajectory generation. 
Over $138$ planning cycles, the average time to generate each trajectory was only $32$ ms, with a standard deviation of $6.3$ ms (Fig. \ref{fig:PredictionReal}F). 
This computational performance allowed the system could respond quickly even in large-scale scenarios. 
As a result, despite mild prediction inaccuracies early in the mission (such as at $t_1$ and $t_2$), our planner maintained high adaptability to address discrepancies between predictions and reality in a timely manner.

\subsubsection{Efficient and Safety-assured Scanning within Clutters}
\label{subsubsec:safe_efficient_clutters}

\begin{figure*}[t]
\centering
\includegraphics[width=0.99\linewidth]{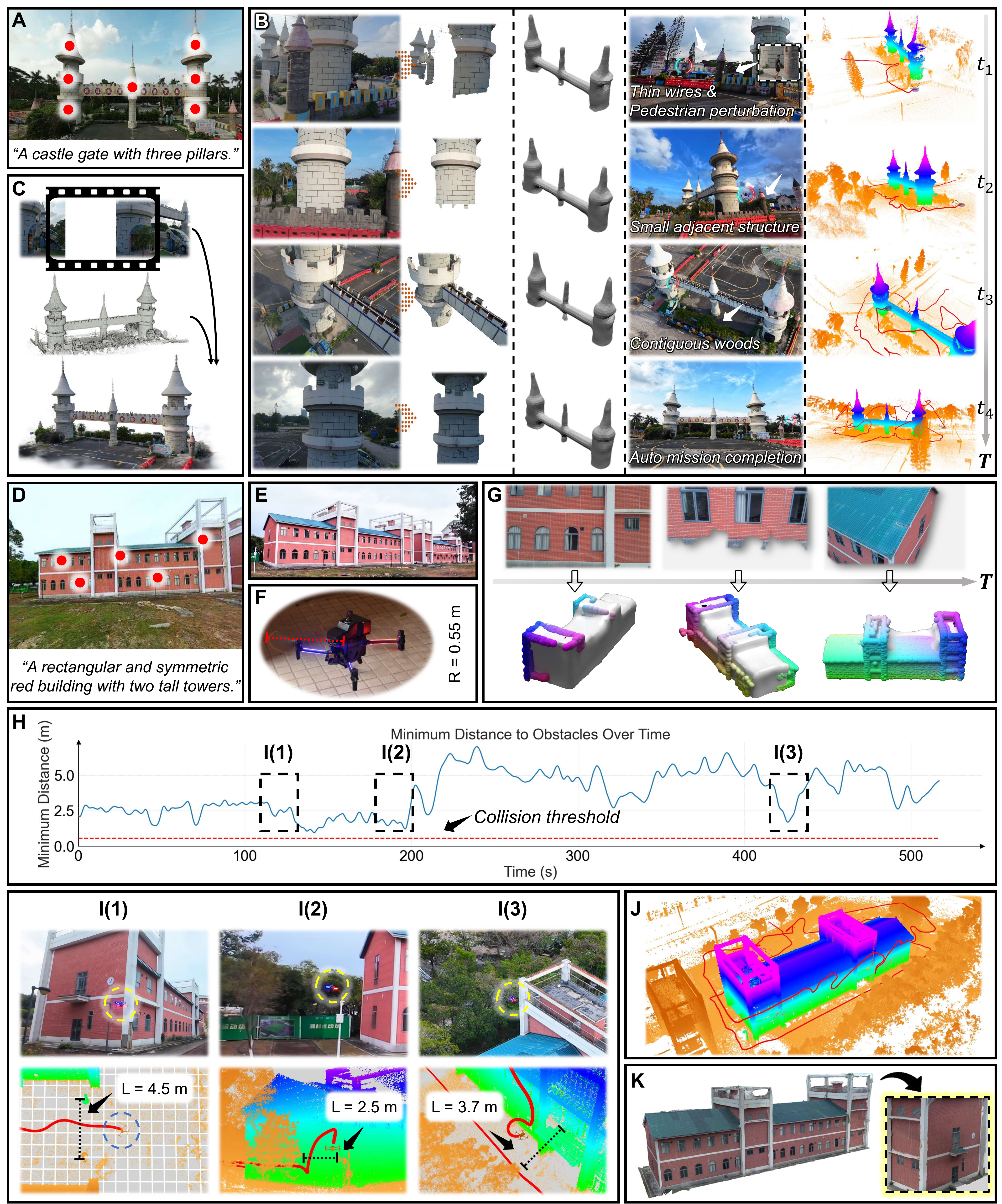}
\vspace{-0.3cm}
\caption{Efficient and Safety-assured Scanning within Clutters.
(A,D) Low-effort user prompts (text + sparse clicks) specifying the target \textit{castle gate} and \textit{red-brick building}.
(B) Castle-gate mission: (left$\rightarrow$right) visual target segmentation, predicted complete surface, representative static obstacles and dynamic elements (\textit{e.g.}, thin wires, pedestrians, adjacent structures, woods), and the corresponding 3D target segmentation (rainbow-colored) with executed trajectory.
(C,K) High-fidelity reconstructions from the scanned data for the two missions.
(E) Red-brick building site overview.
(F) Specification of the drone, detailing a radius of 0.55 m.
(G) Red-brick building: prompt-grounded segmentation and online surface predictions consistent with partial observations.
(H) Minimum obstacle distance over time; the collision threshold equals the drone radius (red dashed line), remaining satisfied throughout the flight.
(I(1-3)) Snapshots of key flight moments, demonstrating safe navigation through cluttered spaces (\textit{e.g.}, a 2.5 m gap near walls and vegetation).
(J) Complete flight result surrounding the red-brick building.
}
\label{fig:PlanningReal}
\vspace{-0.3cm}
\end{figure*}

Generating targeted drone actions to scan human-intended structures—grounded in precise scene understanding—is the final yet critical step in our system.
In unknown open-world environments, drones must not only efficiently conduct comprehensive scanning of user-specified structures but also handle unexpected dense obstacles and dynamic elements (\textit{e.g.}, pedestrians).
Therefore, how to simultaneously achieve both objectives emerges as a key challenge, which we addressed by the seamless integration of scene knowledge and flight skills—coordinating FMs with a hierarchical planner.
In this experiment, we tested our system in complex environments to validate the effectiveness of this approach (Fig. \ref{fig:PlanningReal}).
The drone was tasked with scanning two targets in detail according to textual and visual prompts from the user (Fig. \ref{fig:PlanningReal}A, D): a castle gate surrounded by park trees and a red-brick building in a dense forest.

\begin{table*}[t]
\footnotesize\sf\centering
\caption{Average per-cycle computation time across four in-the-wild missions. Target extents indicate the length, width, and height of the minimal oriented bounding box.}
\label{tab:real_world_computation}
\setlength{\tabcolsep}{5.2pt}
\renewcommand{\arraystretch}{1.1}
\begin{tabular}{l|cccc}
\toprule
\textbf{Real-world Scenarios} 
& \textbf{Target Extents (m)} 
& \textbf{Perception (ms)} 
& \textbf{Prediction (ms)}$^\dagger$ 
& \textbf{Planning (ms)}$^\ddagger$  \\
\midrule
Arch bridge (large building) 
& 8$\times$36$\times$12 
& 402.6 
& 149.3\,+\,432.5 
& 266.3\,$||$\,28.3 \\

Concert hall (campus) 
& 100$\times$60$\times$35 
& 397.7 
& 151.4\,+\,709.8 
& 457.7\,$||$\,32.0 \\

Castle gate (driving school) 
& 9$\times$35$\times$20 
& 394.3 
& 150.9\,+\,411.2 
& 234.6\,$||$\,34.5 \\

Red-brick building (forest) 
& 13$\times$45$\times$16 
& 398.2 
& 148.1\,+\,540.6 
& 309.2\,$||$\,35.7 \\
\bottomrule
\end{tabular}
\par\vspace{2pt}
\raggedright\footnotesize{$^\dagger$ Decomposed as $t_{\text{pred}} + t_{\text{dense}}$, where $t_{\text{pred}}$ is the latency of the multi-modal surface prediction network and $t_{\text{dense}}$ is the latency of inference-time geometry densification.}\\
\raggedright\footnotesize{$^\ddagger$ Decomposed as $t_{\text{global}}\,||\,t_{\text{local}}$, where $t_{\text{global}}$ is the latency of global coverage planning and $t_{\text{local}}$ is the latency of the local trajectory planning.}
\vspace{-0.4cm}
\end{table*}

After interpreting the prompts, the perception module accurately segmented the user-intended castle gate and red-brick building within cluttered surroundings (Fig. \ref{fig:PlanningReal}B and G). 
The red-brick building was challenging due to a nearby similar-looking structure (Fig. \ref{fig:PlanningReal}E), but the system resolved this ambiguity using the user's visual clicks and rectification from 3D modality, which clarified linguistic uncertainties. 
Beyond this, intricate features like undulating roofs, irregular spires, and non-convex features hindered precise surface prediction. 
Yet our multi-modal model inferred a qualitatively reasonable 3D target geometry consistent with inputs (Fig. \ref{fig:PlanningReal}B and G).

In terms of scanning flight performance (Fig. \ref{fig:PlanningReal}B, J), the drone operated efficiently with a compact trajectory around the targets, avoiding time wasted on irrelevant areas, local detours, or hasty directional shifts with unclear intent.
Meanwhile, it dynamically adjusted flight trajectories and sensor viewpoints to cover all critical regions of the targets, eliminating redundant scans.
This efficiency arises from our decoupled system architecture: (1) it concretizes abstract scene knowledge (empowered by FMs) via the form of predicted surfaces; (2) it enables independent planning that fully leverages this guidance to generate global coverage paths with long-horizon temporal consistency.
Ensuring real-time safety in these cluttered environments is equally non-trivial.
The castle gate, as the entrance to a driving school, had frequent pedestrian and vehicle traffic; surrounding woodlands, small structures, and thin wires further constrained flight (Fig. \ref{fig:PlanningReal}B).
Within the forest, navigable space was limited to a narrow $2.5$-m passage—barely enough for the drone with a $1.1$-m diameter (Fig. \ref{fig:PlanningReal}F and I).
Despite these high-risk conditions, our planner maintained consistent safe clearance from all obstacles to avoid collisions.
For the red-brick building mission specifically, Fig. \ref{fig:PlanningReal}H tracks the drone's minimum distance to obstacles over $500$ seconds: this distance remained above the $0.55$-m collision threshold (red dotted line, based on the AV's radius).
This safety assurance is attributed to the asynchronous hierarchical design of our planner, which allows the local planner can react swiftly to environmental changes while adhering to the efficient, target-centric global paths.

Taken together, in our system, FMs transform complex, unknown open-world scenes into deterministic structured knowledge, empowering the planner to easily generate precise 3D scanning trajectories of the user-intended structure, with both high efficiency and real-time safety.
Complete video demonstrations of both missions can be found in Extension 4 and 5.

\subsubsection{Real-time Performance and Computational Profiling}
\label{subsubsec:real_world_comp}

Tab. \ref{tab:real_world_computation} summarizes the average per-update runtime of each module across the four real-world missions, measured over the full flight duration and expressed as wall-clock time per cycle for perception, prediction, and planning.

On the onboard computer, the perception stack (SAM2-based segmentation with cross-modal refinement) runs at about $400$ ms per update in all scenarios, corresponding to an effective rate of roughly $2$-$2.5$ Hz. 
This shows that prompt-grounded 2D-3D target segmentation can be sustained at a stable real-time frequency even in large, cluttered outdoor environments.

The prediction pipeline, executed on the ground station, has two stages: the multi-modal surface predictor and the inference-time densification module. 
The forward pass of the predictor is nearly constant at $\sim150$ ms across all targets, as it depends only on the fixed input-output point-set sizes. 
In contrast, the densification cost naturally scales with physical target size, ranging from about $0.4$ s for the arch bridge and castle gate to about $0.7$ s for the concert hall. 
This is expected, since the cascade of upsampling networks stops adaptively once the desired spatial resolution is reached. 
Even in the largest scene, however, the overall prediction latency stays below $1$ s and runs in its own low-rate thread, providing timely foresight without becoming a bottleneck for closed-loop flight.
With the maximum speed capped at $1$ m/s, this latency means that at most newly observed surface from $1$-m movement is missing from each prediction update, so the predicted geometry remains closely aligned with the latest observations rather than drifting far from reality and misguiding the planner.

For planning, the advantages of the asynchronous hierarchical framework and computation-efficient designs are clearly reflected in the timing profile. 
Global coverage planning takes roughly $250$-$310$ ms per cycle, with runtime increasing with target scale as it reasons over the full predicted mesh and the associated viewpoint set; nevertheless, this cost is only $\sim40\%$ of the prediction latency, allowing to comfortably keep pace with prediction updates.
Conversely, the local planner maintains an almost constant runtime of about $28$-$36$ ms per call across all missions, since its complexity depends on the fixed-horizon path segment rather than the overall scene scale. 
This separation enables high-rate trajectory replanning and rapid reactions to newly revealed obstacles, regardless of how large or complex the target structure is.

Taken together, these results confirm that all components of \textbf{F}ly\textbf{C}o meet their intended real-time budgets: perception operates at a few hertz on the edge device, prediction delivers sub-second structural foresight on the ground station, and the global-local planner jointly preserves long-horizon efficiency and high-frequency safety.
This demonstrates that our FM-empowered scanning system is not only accurate and robust, but also computationally practical for real-world deployment.

\begin{table*}[h]
\footnotesize\sf\centering
\caption{Human prior inputs required by different paradigms. $E(h)$ is the human effort measure defined in Sec. \nameref{sec:problem_statement}.}
\label{tab:human_prior}
\setlength{\tabcolsep}{5.2pt}
\renewcommand{\arraystretch}{1.1}
\begin{tabular}{l|cccccc}
\toprule
\textbf{Methods} & \textbf{Text} & \textbf{Sparse 2D Anno.} & \textbf{3D BBox} & \textbf{Manual 3D Segmentation} & \textbf{Pre-defined 3D Flight Pattern} & $E(h)$ \\
\midrule
Plan3D & -- & -- & \checkmark & \checkmark & \checkmark & High \\
Star-Searcher & -- & -- & \checkmark & -- & -- & Medium \\
\rowcolor{gray!15}
Ours & \checkmark & \checkmark & -- & -- & -- & \textbf{Low} \\
\bottomrule
\end{tabular}
\par\vspace{2pt}
\raggedright\footnotesize{From left to right, the prior inputs require increasing human effort (low$\rightarrow$high).}
\vspace{-0.4cm}
\end{table*}

\subsection{Comparisons with State-of-the-Art Methods}
\label{subsec:comparisons}

To complement the field demonstrations, we conduct systematic benchmark experiments in simulation to compare \textbf{F}ly\textbf{C}o against representative SOTA approaches from existing practical paradigms: (\textit{two-stage} and \textit{exploration-based}).
From these experiments, our system achieves clearly superior performance in mission efficiency, information completeness, flight safety, and stability across runs.
We first introduce the simulated environments and experimental protocol, followed by the baselines and quantitative metrics used for evaluation.
We then report and analyze benchmark results, illustrating why \textbf{F}ly\textbf{C}o outperforms these alternative paradigms.

\vspace{-0.3cm}
\begin{figure}[h]
\centering
\includegraphics[width=0.99\linewidth]{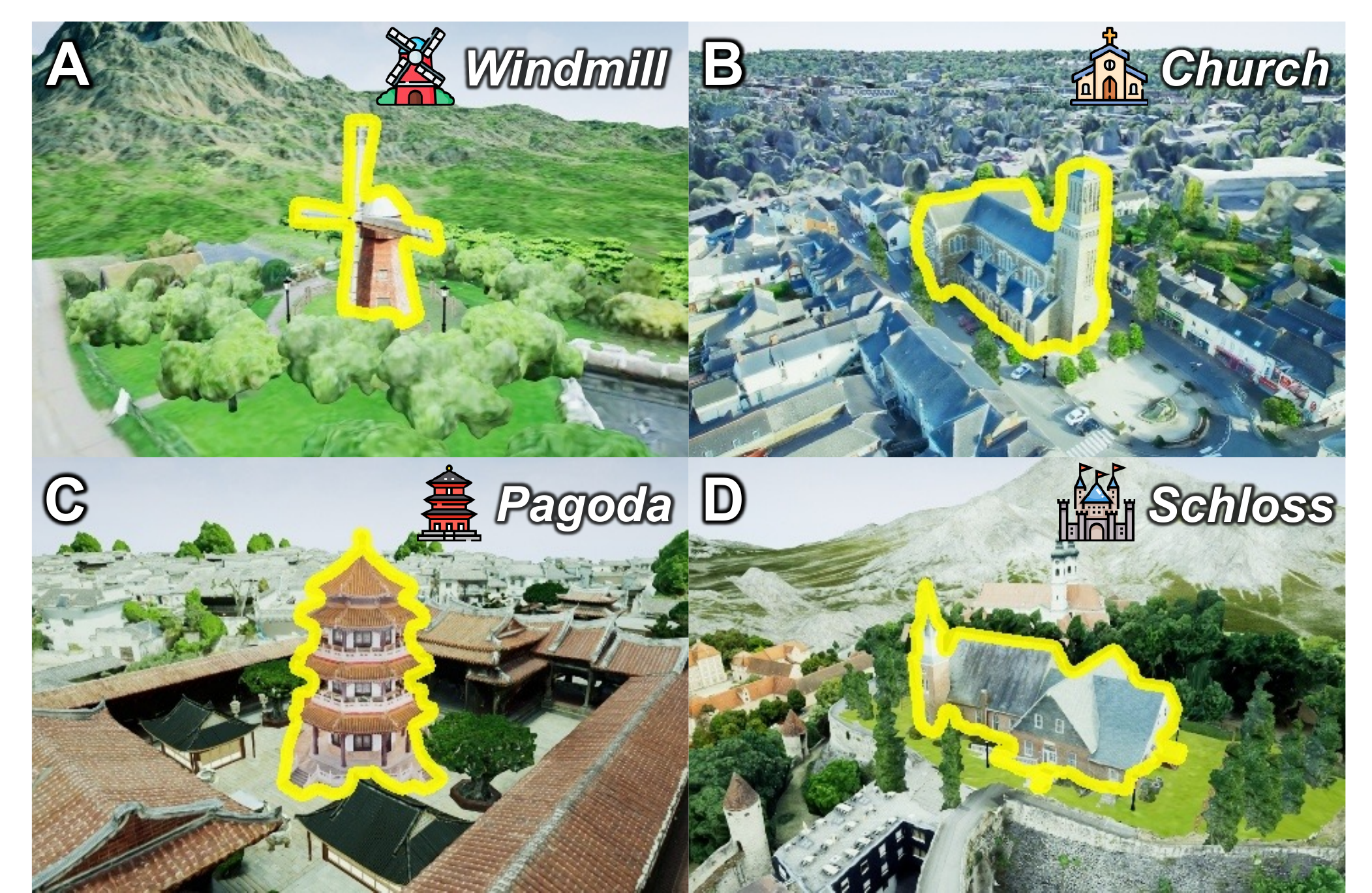}
\vspace{-0.3cm}
\caption{Overview of the simulated environments used in our experiments: (A) \textit{Windmill}, (B) \textit{Church}, (C) \textit{Pagoda}, and (D) \textit{Schloss}.
The target structures are outlined in yellow.}
\label{fig:SimEnv}
\vspace{-0.4cm}
\end{figure}

\subsubsection{Simulated Scenarios and Experimental Setup}
\label{subsubsec:sim_scenarios}

To enable rigorous, quantitative benchmark comparisons, we perform these tests in the physically realistic AirSim simulator (\citealt{airsim}) across four complex, previously unseen scenarios that mirror typical real-world inspection targets.
Specifically, we consider: (1) a \textit{Windmill} located in a rural, hilly landscape; (2) a \textit{Church} situated in a cluttered European-style town; (3) a \textit{Pagoda} standing at the center of a dense temple courtyard; and (4) a \textit{Schloss} surrounded by varied terrain and vegetation (see Fig. \ref{fig:SimEnv}). 
In all scenarios, a virtual drone is equipped with a $360\,^\circ$ LiDAR and a gimbal-mounted RGB camera, and is tasked with collecting both geometric and visual data of the designated target.
The maximum translational speed is limited to $1$\,m/s and the gimbal's maximum angular speed to $20\,^\circ$/s, which is identical for all benchmarked approaches.
For each method, we run $20$ repeatable trials per scenario on the same desktop computer (Intel Core i7-13700KF CPU, $32$ GB RAM, NVIDIA RTX4060 GPU) to ensure a fair computational comparison.

\begin{table*}[t]
\footnotesize\sf\centering
\caption{Benchmark results in simulation. For each scenario, we report mean (Avg), standard deviation (Std), maximum (Max), and minimum (Min) over 20 trials. Extent is the target's 3D size ($L_x \times L_y \times L_z$) in meters.}
\label{tab:sim_benchmark}
\setlength{\tabcolsep}{5.2pt}
\renewcommand{\arraystretch}{1.1}
\begin{tabular}{c|c|c|cccc|cccc|c}
\toprule
\multirow{2}{*}{\textbf{Scenarios}} & \multirow{2}{*}{\textbf{Extents (m)}} & \multirow{2}{*}{\textbf{Methods}}
& \multicolumn{4}{c|}{\textbf{Flight Time (s)}$\,\downarrow$}
& \multicolumn{4}{c|}{\textbf{Info. Completeness (\%)}$\,\uparrow$}
& \multirow{2}{*}{\textbf{Success Rate}$\,\uparrow$} \\
& & & \textbf{Avg} & \textbf{Std} & \textbf{Max} & \textbf{Min} & \textbf{Avg} & \textbf{Std} & \textbf{Max} & \textbf{Min} & \\
\midrule

\multirow{5}{*}{Windmill}
& \multirow{5}{*}{16$\times$30$\times$32}
& Plan3D ($Z_1$)         &676.6  &18.77  &722.4  &659.6  &71.67  &1.62  &75.54  &69.84  &15/20  \\
& & Plan3D ($Z_2$)       &798.3  &27.17  &853.9  &759.3  &89.76  &2.59  &89.76  &82.11  &13/20  \\
& & Star-Searcher ($Z_1$)&787.4  &59.56  &936.7  &676.6  &58.83  &3.38  &64.47  &52.78  &20/20  \\
& & Star-Searcher ($Z_2$)&981.9  &77.50  &1167.0  &823.5  &84.30  &3.60  &89.49  &73.67  &20/20  \\
\rowcolor{gray!15}
\cellcolor{white} & \cellcolor{white} & Ours                 &\textbf{329.4}  &\textbf{8.93}  &357.9  &321.4  &\textbf{96.04}  &\textbf{0.58}  &97.11  &94.88  &\textbf{20/20}  \\
\midrule

\multirow{5}{*}{Church}
& \multirow{5}{*}{35$\times$17$\times$18}
& Plan3D ($Z_1$)         &728.8  &27.99  &779.3  &676.6  &72.12  &1.71  &74.84  &69.58  &9/20  \\
& & Plan3D ($Z_2$)       &830.4  &30.35  &865.3  &789.3  &90.21  &1.89  &92.33  &86.53  &7/20  \\
& & Star-Searcher ($Z_1$)&543.1  &45.20  &611.2  &463.7  &56.92  &2.89  &63.34  &52.83  &19/20  \\
& & Star-Searcher ($Z_2$)&787.3  &54.27  &893.4  &706.9  &78.40  &3.77  &83.38  &68.39  &19/20  \\
\rowcolor{gray!15}
\cellcolor{white} & \cellcolor{white} & Ours                 &\textbf{336.6}  &\textbf{9.43}  &364.8  &323.6  &\textbf{94.53}  &\textbf{0.71}  &95.54  &93.23  &\textbf{20/20}  \\
\midrule

\multirow{5}{*}{Pagoda}
& \multirow{5}{*}{15$\times$15$\times$19}
& Plan3D ($Z_1$)         &432.5  &23.12  &478.8  &382.6  &51.22  &2.06  &54.70  &48.39  &10/20  \\
& & Plan3D ($Z_2$)       &571.1  &24.14  &611.9  &541.9  &73.42  &0.84  &74.88  &72.48  &9/20  \\
& & Star-Searcher ($Z_1$)&282.1  &34.42  &367.0  &231.9  &37.55  &2.71  &40.98  &30.51  &18/20  \\
& & Star-Searcher ($Z_2$)&397.9  &45.98  &504.7  &318.8  &66.79  &3.61  &73.29  &59.62  &17/20  \\
\rowcolor{gray!15}
\cellcolor{white} & \cellcolor{white} & Ours                 &\textbf{225.1}  &\textbf{6.22}  &240.3  &216.2  &\textbf{93.83}  &\textbf{0.63}  &94.62  &92.33  &\textbf{19/20}  \\
\midrule

\multirow{5}{*}{Schloss}
& \multirow{5}{*}{22$\times$33$\times$20}
& Plan3D ($Z_1$)         &756.7  &32.29  &804.3  &701.6  &73.13  &1.97  &77.58  &70.13  &18/20  \\
& & Plan3D ($Z_2$)       &981.4  &21.36  &1012.22  &961.9  &91.08  &0.69  &92.36  &90.08  &12/20  \\
& & Star-Searcher ($Z_1$)&501.9  &36.04  &593.3  &446.2  &58.55  &2.91  &63.28  &53.50  &17/20  \\
& & Star-Searcher ($Z_2$)&716.0  &55.05  &827.8  &611.9  &79.48  &3.40  &84.27  &70.43  &18/20  \\
\rowcolor{gray!15}
\cellcolor{white} & \cellcolor{white} & Ours                 &\textbf{363.9}  &\textbf{7.52}  &380.8  &349.5  &\textbf{96.50}  &\textbf{0.66}  &97.58  &95.45  &\textbf{20/20}  \\
\bottomrule
\end{tabular}
\par\vspace{2pt}
\raggedright\footnotesize
{$Z_1$ and $Z_2$, respectively, represent the operation zones of different sizes specified by the user in the form of 3D bounding boxes.}
\vspace{-0.4cm}
\end{table*}

\subsubsection{Baselines and Evaluation Metrics}
\label{subsubsec:baselines and metrics}

We benchmark \textbf{F}ly\textbf{C}o against two representative SOTA works, Plan3D (\citealt{hepp2018plan3d}) and Star-Searcher (\citealt{luo2024star}), that instantiate the dominant paradigms for 3D target structure scanning: a \textit{two-stage} pipeline and an \textit{exploration-based} system.
Plan3D follows a two-stage workflow:  the drone first performs a coarse scan within a user-specified 3D bounding box; the user then manually segments the target, after which an offline coverage planner computes a refined close-range trajectory. 
Since no official implementation is available, we realize this paradigm by combining a Boustrophedon coverage planner (\citealt{choset2000coverage}) for the coarse path and FC-Planner (\citealt{feng2024fc}) for fine 3D coverage in the second stage.
Star-Searcher represents exploration-based scanning methods for drones equipped with LiDAR and a camera.
Given a user-defined 3D bounding box, it incrementally explores all occupied space in this volume using a frontier strategy, while simultaneously planning camera motions so that all discovered surfaces are visually inspected.
We use the official implementation and align its sensor suite and motion limits with those of our system for a fair comparison.

Notably, both Plan3D and Star-Searcher depend on user-specified 3D bounding boxes to define the operational zone, and Plan3D additionally needs manual target segmentation between its coarse and fine stages.
To approximate a favourable yet realistic deployment for the baselines, we evaluate them under two bounding-box settings: a tighter box $Z_1$ and a larger box $Z_2$.
Even in the larger setting $Z_2$, we supply a near-ideal box that tightly encloses the target and take-off point while minimizing irrelevant volume, which substantially biases the comparison in favour of the baselines and thus yields conservative estimates of \textbf{F}ly\textbf{C}o's advantage.

\begin{figure}[h]
\centering
\includegraphics[width=0.99\linewidth]{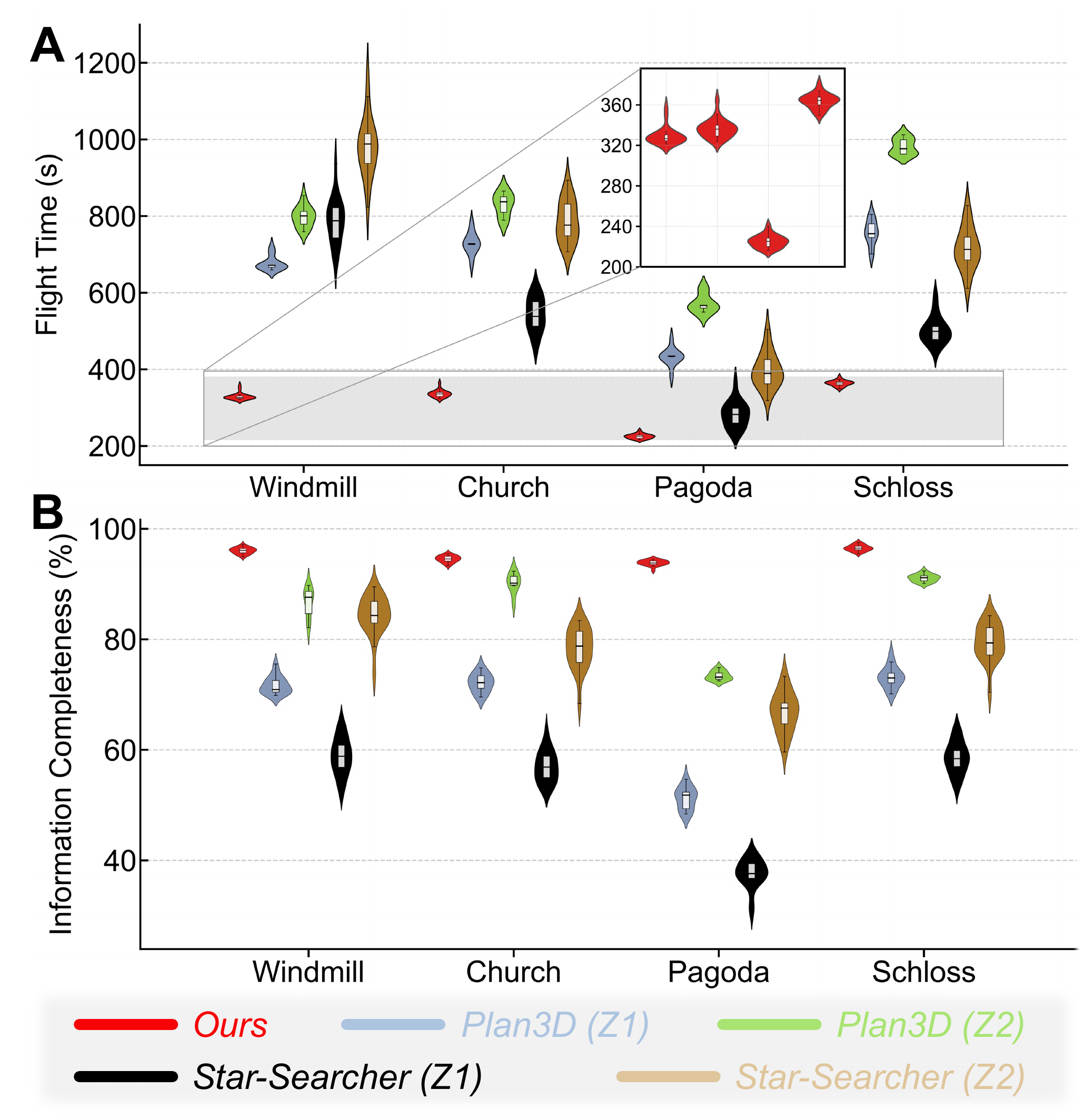}
\vspace{-0.3cm}
\caption{Violin plots of flight time (A) and information completeness (B) across four scenarios, comparing our system to SOTA baselines. Shapes show the density; embedded box plots mark the median, interquartile range, and whiskers at $\pm$1.5$\times$ IQR.}
\label{fig:BmkVioErr}
\vspace{-0.5cm}
\end{figure}

Performance is evaluated along four complementary dimensions—efficiency, scanned data integrity, safety, and stability—using the following metrics.
\textbf{Flight time.}
Mission efficiency is measured by the total airborne time (in seconds) from takeoff to landing.
We deliberately exclude any human setup time (such as bounding-box specification or manual segmentation), which effectively favors the baselines by ignoring their additional interaction overhead.
Lower values indicate higher efficiency.
\textbf{Information completeness.}
Data integrity is quantified as the percentage of the ground-truth target surface that is successfully reconstructed.
Both the reconstruction and ground-truth meshes are voxelized on a $0.05$\,m $\times$ $0.05$\,m grid, and completeness is computed as the ratio of recalled occupied grid cells to the total number of occupied cells in the ground truth.
Higher percentages indicate more complete coverage.
\textbf{Success rate.}
Safety and reliability are captured by the success rate, defined as the fraction of runs that finish without collision and within a pre-defined time budget $T_{\max} = 1500$ s.
For all methods, each trial terminates either when the algorithm declares completion (coverage saturated under its own stopping rule) or when the elapsed flight time exceeds $T_{\max}$; both collisions and timeouts are counted as failures.
\textbf{Performance consistency.}
To assess stability across runs, we report the variability of flight time and information completeness over $20$ trials per scenario, using standard deviation and the dispersion of their empirical distributions (visualized via violin plots in Fig. \ref{fig:BmkVioErr}).
Tighter distributions (smaller spread) correspond to more stable behavior across different runs.

\begin{figure*}[t]
\centering
\includegraphics[width=0.99\linewidth]{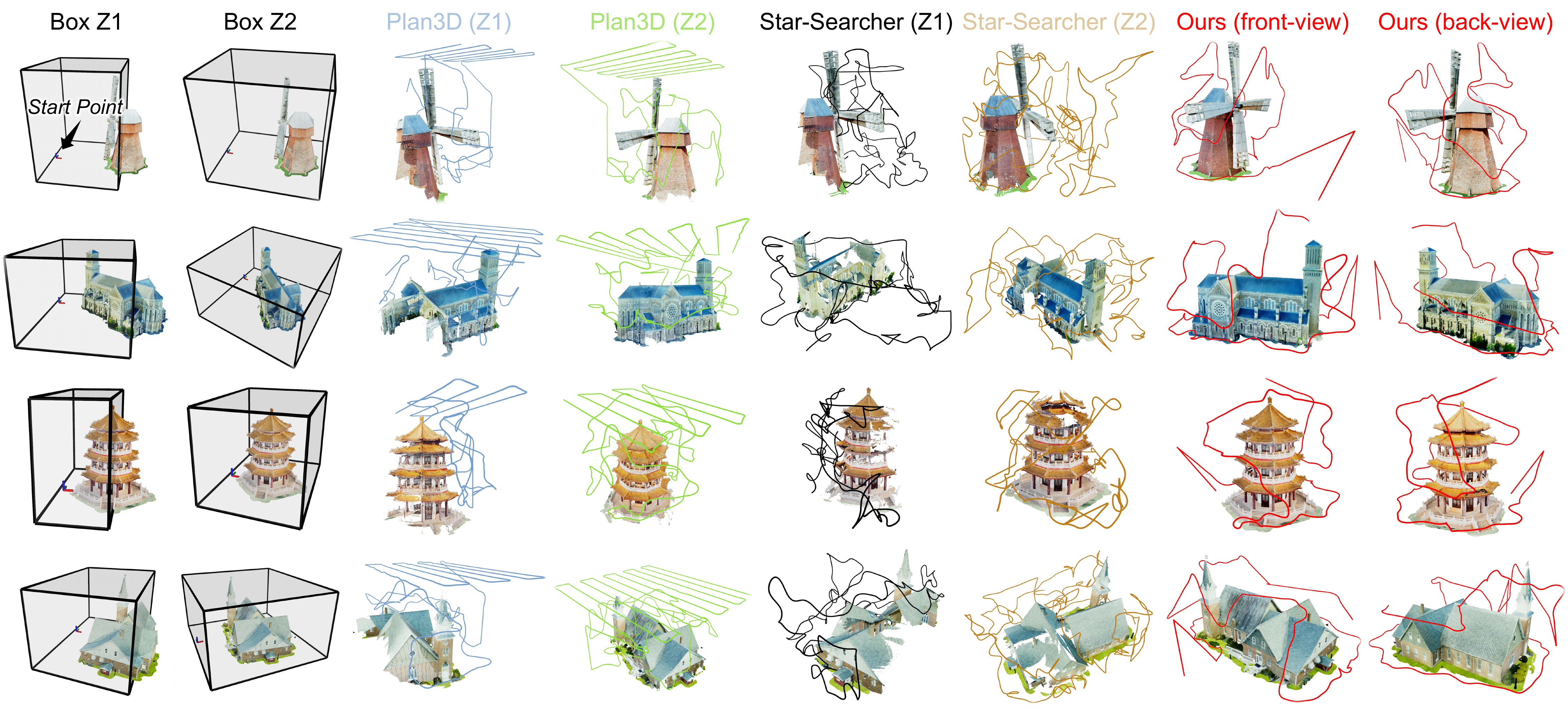}
\vspace{-0.3cm}
\caption{Benchmark visualizations across the four scenarios. Rows show flight trajectories and reconstructions from the scanned data for our method versus baselines, highlighting more complete coverage and improved efficiency achieved by \textbf{F}ly\textbf{C}o.}
\label{fig:BmkVis}
\vspace{-0.5cm}
\end{figure*}

\subsubsection{Benchmark Results and Analysis}
\label{subsubsec:benchmark_results}

We now present and analyze the benchmark results to quantify both the magnitude and the origin of \textbf{F}ly\textbf{C}o's performance gains.
Our discussion is organized around the four evaluation metrics—efficiency, data integrity, safety, and stability—and explicitly relates the observed outcomes to the objective formulated in Sec. \nameref{sec:problem_statement}.
The required human prior inputs of each paradigm are summarized in Tab. \ref{tab:human_prior}.
Quantitative results are reported in Tab. \ref{tab:sim_benchmark}, qualitative scanning trajectory and reconstruction visualizations are shown in Fig. \ref{fig:BmkVis}, and run-to-run variability is illustrated in Fig. \ref{fig:BmkVioErr}.
Complete execution traces for all benchmark comparisons are provided in Extension 6.

Tab. \ref{tab:sim_benchmark} and Fig. \ref{fig:BmkVioErr} jointly reveal that \textbf{F}ly\textbf{C}o achieves a consistent, scenario-wide advantage over both dominant paradigms, while requiring substantially lower-effort priors (Tab.~\ref{tab:human_prior}).
Across all four environments, \textbf{F}ly\textbf{C}o completes missions in $225$-$364$ s on average, whereas the baselines mostly fall in the $400$-$1000$ s range, depending on the paradigm and the bounding-box setting.
Meanwhile, \textbf{F}ly\textbf{C}o maintains high information completeness of $93.8\%$-$96.5\%$, while Plan3D and Star-Searcher often exhibit markedly lower completeness.
Importantly, these gains are accompanied by strong reliability: \textbf{F}ly\textbf{C}o succeeds in $79/80$ trials overall (with $20/20$ success in three scenarios), whereas both baselines incur non-negligible failure due to collisions or timeouts, particularly in larger or more cluttered environments.
These trends map directly to the objective in Sec. \nameref{sec:problem_statement}, which seeks to (1) minimize mission time, (2) maximize target information completeness, and (3) minimize human effort, subject to safety and feasibility constraints.
On the efficiency-integrity frontier, \textbf{F}ly\textbf{C}o strictly dominates the baselines: it simultaneously reduces flight time by roughly $1.25\times$ to $3.0\times$ (depending on scenario) while often exceeding baselines by $4.3$-$56.2$ percentage points in information completeness.
Notably, this advantage is achieved even though we exclude all human interaction time from the flight-time metric, which biases the comparison in favor of the baselines.
Beyond average performance, \textbf{F}ly\textbf{C}o also exhibits significantly greater stability.
As visualized in Fig. \ref{fig:BmkVioErr}, the distributions of both flight time and information completeness are notably tighter for \textbf{F}ly\textbf{C}o, with minimal long tails and fewer outliers across all scenarios.
Quantitatively, \textbf{F}ly\textbf{C}o achieves very small run-to-run variability in completeness (Std $\leq 0.71$ in all scenarios; Tab. \ref{tab:sim_benchmark}) and similarly low variability in flight time, indicating stable closed-loop behavior rather than occasional highly inefficient runs.
Regarding human effort, \textbf{F}ly\textbf{C}o reduces operator burden by relying only on a text prompt and sparse 2D annotations, rather than effortful low-level priors.
Taken together, the benchmark results substantiate that \textbf{F}ly\textbf{C}o empirically achieves a strictly better efficiency-completeness trade-off while simultaneously reducing human effort and maintaining high reliability, rather than improving one objective at the expense of another.

We next analyze the sources of \textbf{F}ly\textbf{C}o's gains from a system-design perspective.
Rather than attributing the improvements to individual algorithmic choices, we focus on how the FM-empowered perception-prediction-planning loop fundamentally reshapes the scanning process and resolves limitations inherent to existing paradigms.

\textbf{Perception: precise semantic grounding.}
The first source of improvement lies in perception.
With FM-based grounding, \textbf{F}ly\textbf{C}o eliminates the need for explicit, low-level geometric intervention during operation.
In two-stage systems, manual 3D segmentation is used to prevent spending flight time on irrelevant regions, but it externalizes part of the perception problem to humans, whose effort is non-negligible in real deployments.
Exploration-based methods replace this step with a user-specified 3D bounding box; however, for non-convex and irregular targets, any box inevitably over-approximates the operational volume and introduces wasted traversal, most visibly in \textit{Windmill} (Tab. \ref{tab:sim_benchmark}).
Crucially, even with near-ideal $Z_2$ boxes that tightly enclose the target, both baselines still lag behind \textbf{F}ly\textbf{C}o in efficiency.
Moreover, both baselines exhibit pronounced sensitivity to box specification: performance shifts substantially between $Z_1$ and $Z_2$ (Tab. \ref{tab:sim_benchmark}), casting doubt on their practical viability.
In contrast, FM-enabled perception anchors the scanning process directly to the semantic target throughout the mission.
This tight coupling avoids irrelevant exploration by construction, leading to immediate gains in flight efficiency without relying on either manual intervention or geometric over-approximation.

\textbf{Prediction: continuous global structural foresight.}
The second and arguably most decisive advantage arises from prediction.
Existing paradigms lack global context of the target's complete structure during execution, whereas our shape inference exactly supplies this foresight.
Two-stage pipelines attempt to recover global information via a pre-defined coarse scan, making the downstream close-range stage highly sensitive to what was observed in that first pass (\textit{e.g.}, omissions under $Z_1$).
Exploration-based methods face a more fundamental limitation: without any global structural knowledge, decisions are made purely based on local frontier information.
This tends to produce discontinuous coverage with insufficient overlap, and ultimately poor reconstruction completeness, as reflected by the lower information completeness in Tab. \ref{tab:sim_benchmark}.
By contrast, \textbf{F}ly\textbf{C}o maintains a continuously updated global prediction of the target structure, inferred from partial observations and refined online.
This predictive capability provides foresight throughout the mission, allowing the system to proactively steer the drone towards yet-unobserved but structurally relevant regions.
Thus, coverage is not only more complete but also more purposeful, explaining the consistently high information completeness achieved across all scenarios.

\begin{figure*}[t]
\centering
\includegraphics[width=0.99\linewidth]{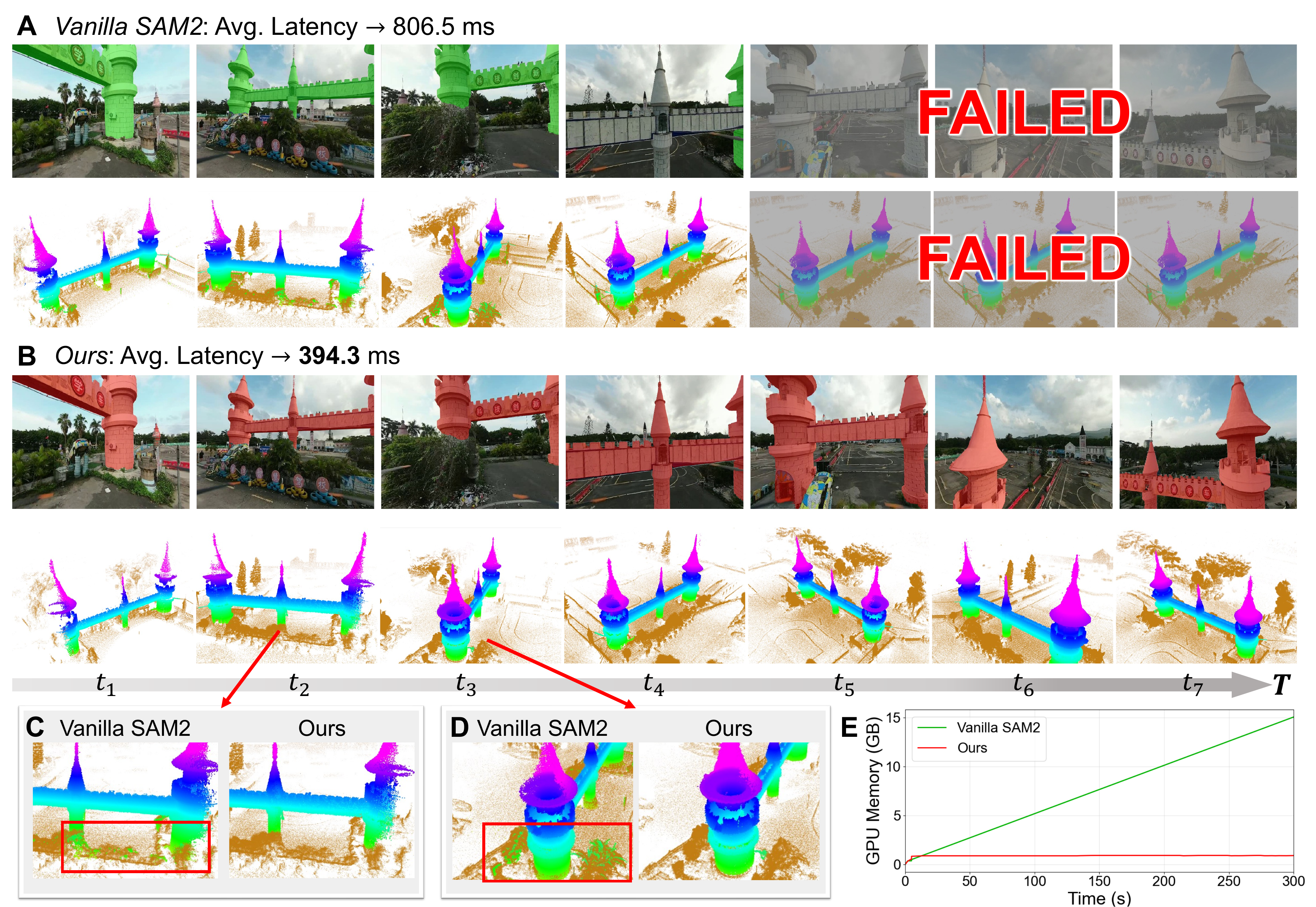}
\vspace{-0.3cm}
\caption{Perception ablation on real-world flight data.
(A) Vanilla SAM2: 2D target masks degrade and eventually collapse as viewpoint shifts.
(B) Ours: 2D/3D grounding stays temporally stable throughout the flight.
(C-D) 3D segmentation comparison: Vanilla SAM2 leaks irrelevant regions into the target structure. 
(E) GPU memory usage over time.}
\label{fig:SAMAblation}
\vspace{-0.5cm}
\end{figure*}

\textbf{Planning: efficient and safe scanning operation.}
Finally, the planning translates results of perception and prediction into concrete flight behavior.
Prediction-guided global planning enables \textbf{F}ly\textbf{C}o to reason over the entire target structure while explicitly enforcing temporal consistency.
This suppresses unnecessary revisits and redundant scans that are common in exploration-based strategies driven by purely local decisions, thereby directly improving efficiency and coherence (Fig. \ref{fig:BmkVis}).
Equally important is the asynchronous hierarchical design.
Unlike serial two-stage pipelines, where planning and execution are tightly coupled and slow updates can delay reaction to newly observed obstacles, \textbf{F}ly\textbf{C}o decouples deliberative global planning from high-rate local replanning.
This design preserves rapid responsiveness to environmental changes, which is critical in cluttered or large-scale environments.
The higher success rates observed in complex scenarios such as \textit{Church} and \textit{Pagoda} (Tab. \ref{tab:sim_benchmark}) can be directly attributed to this ability to reconcile long-horizon planning with real-time safety constraints.

Overall, these results confirm that \textbf{F}ly\textbf{C}o's advantages do not stem from isolated components, but from a coherent closed-loop system architecture empowered by FMs.
This integration enables the system to reduce flight time, increase information completeness, and maintain efficient, safe operation without relying on fragile low-level priors or manual intervention.
As such, the FM-empowered perception-prediction-planning loop represents a more appropriate and practical paradigm for open-world 3D target structure scanning than existing two-stage or exploration-based approaches.

\subsection{Ablation Studies}
\label{subsec:ablation_studies}

Comprehensive ablation studies are carried out to disentangle the contributions of individual modules in \textbf{F}ly\textbf{C}o and to validate the design choices of the proposed perception-prediction-planning loop for effectively turning FM-guidance into reliable 3D target structure scanning behavior in the open world.
We systematically remove or replace critical components to examine how each affects grounding robustness, structural foresight, flight efficiency and safety.
These experiments clearly reveal how FMs features and advanced flight skills affect the system performance.
Detailed results and discussion are presented below, with additional complete recordings and cases in Extension 6.

\subsubsection{Ablations on Perception}
\label{subsubsec:ablation_perception}

This ablation study investigates whether off-the-shelf segmentation FMs can directly satisfy the perception requirements of open-world target structure scanning, and why a tailored perception design is necessary for reliable system-level operation.
Unlike static image benchmarks, our task demands precise and temporally stable grounding of a user-specified target structure in both 2D and 3D throughout long-horizon flight under viewpoint changes, partial occlusions, and cluttered environments.

We conduct a qualitative comparison using a real-world flight bag from a castle gate scanning mission, which contains raw sensor streams recorded during autonomous flight.
We compare two approaches:
(1) \textit{Vanilla SAM2}, which applies the original SAM2 for 2D segmentation and projects the masks onto the point cloud for 3D target extraction;
and (2) \textbf{F}ly\textbf{C}o's perception module, which incorporates cross-modal refinement for temporally consistent target grounding.
Both methods receive the same one-time text prompt and sparse visual annotations at takeoff and run on the same onboard computer (NVIDIA Orin NX), with no further human input.
We evaluate the two methods along four dimensions that are critical for system integration:
(1) 2D segmentation temporal stability;
(2) geometric accuracy of the derived 3D target structure;
(3) perception latency; and
(4) GPU memory consumption over time.
Representative results are shown in Fig. \ref{fig:SAMAblation}.

\textit{Vanilla SAM2} initially produces plausible 2D segmentations, but its output becomes unstable as the viewpoint evolves (Fig. \ref{fig:SAMAblation}A).
At $t_4$, it already misses large portions of the target, and from $t_5$ onwards, the model persistently outputs empty masks, causing a complete perception failure.
This behavior highlights a fundamental limitation of directly applying frame-based segmentation models to long-horizon, streaming perception: without explicit temporal grounding and cross-modal correction, segmentation errors accumulate and eventually lead to failures.
In contrast, our method maintains stable and accurate target grounding throughout the entire flight (Fig. \ref{fig:SAMAblation}B).
By continuously refining segmentation using geometric cues, the perception module dynamically corrects 2D errors and preserves temporal coherence.

Even before outright failure, \textit{Vanilla SAM2} exhibits severe inconsistencies when lifted from 2D to 3D.
As illustrated in Fig. \ref{fig:SAMAblation}C-D, projecting 2D masks onto the point cloud often introduces surrounding trees, walls, or other obstacles into the target structure.
This occurs because 2D segmentation alone cannot reason about spatial relationships or depth ordering, resulting in erroneous inclusion of geometrically unrelated regions.
Our perception module explicitly enforces cross-modal consistency between image observations and partial geometry.
As a result, the recognized 3D target remains clean and well-aligned with the true physical structure, avoiding the accumulation of incorrect segmentations.

Beyond accuracy, perception must also be computationally sustainable for long-duration missions.
\textit{Vanilla SAM2} exhibits steadily increasing GPU memory usage over time, which arises from memory mechanisms not designed for continuous streaming inputs (Fig. \ref{fig:SAMAblation}E).
This behavior makes it unsuitable for extended flights on resource-constrained onboard platforms.
Conversely, our perception incorporates a memory bank optimized for streaming sensor processing, allowing stable and bounded GPU memory consumption throughout the mission.
Moreover, the average perception latency is reduced by more than half, from approximately $800$ ms with \textit{Vanilla SAM2} to around $400$ ms with our method.
This reduction enables real-time onboard operation without blocking the closed-loop flight.

These results demonstrate that perception for open-world target structure scanning cannot be treated as a plug-and-play application of existing segmentation FMs.
Accurate and reliable system behavior requires perception modules that explicitly account for temporal continuity, 2D-3D consistency, and computational efficiency.
The proposed perception design fulfills these requirements, providing stable, accurate, and efficient target grounding that is essential for the downstream prediction and planning stages.

\subsubsection{Ablations on Prediction}
\label{subsubsec:ablation_prediction}
This part investigates the role and effectiveness of the proposed multi-modal surface predictor.
We proceed from system-level impact to predictor-level design validation.
First, we conduct an upper-bound analysis study, quantifying the contribution of predictive foresight to flight efficiency and coverage.
We then assess the intrinsic quality of our predictor under zero-shot surface prediction benchmarks against SOTA models.
Finally, we ablate key design choices (multi-modal conditioning, the alternating-attention fusion strategy, and partial-surface regularization) to understand their effects on prediction performance.
Predictor-level evaluations are performed on the zero-shot Objaverse (\citealt{deitke2023objaverse}) and OmniObject3D (\citealt{wu2023omniobject3d}) datasets, comprising $10,000$ objects and scenes entirely unseen during training.
For a fair comparison, all methods are trained using the same protocol as our predictor with $2,048$ output points, and both predicted and ground-truth shapes are normalized to the unit sphere during inference.
We report L1 Chamfer Distance (L1-CD) for geometric fidelity and F-Score at a threshold of $0.001$ for correspondence, and include inference latency where relevant.
All experiments use the same computational platform as Sec. \nameref{subsec:comparisons} to ensure fair and consistent comparisons across variants.

\begin{table}[h]
\footnotesize\sf\centering
\caption{Ablation experiments for upper-bound analysis. For each scenario, we report average statistics over 20 trials.}
\label{tab:GTComp}
\setlength{\tabcolsep}{3.0pt}
\renewcommand{\arraystretch}{1.1}
\begin{tabular}{c|cccc|cccc}
\toprule
\multirow{2}{*}{\textbf{Variants}}
& \multicolumn{4}{c|}{\textbf{Avg Flight Time (s)}$\,\downarrow$}
& \multicolumn{4}{c}{\textbf{Avg Info. Comp. (\%)}$\,\uparrow$} \\
& \textbf{W}$^1$ & \textbf{C}$^2$ & \textbf{P}$^3$ & \textbf{S}$^4$ & \textbf{W}$^1$ & \textbf{C}$^2$ & \textbf{P}$^3$ & \textbf{S}$^4$\\
\midrule
GT &\textbf{310.1} &\textbf{301.9} &\textbf{203.2} &\textbf{338.8} &\textbf{98.42} &\textbf{97.45} &\textbf{97.86} &\textbf{98.83} \\
\rowcolor{gray!15}
Ours &329.4 &336.6 &225.1 &363.9 &96.04 &94.53 &93.83 &96.50 \\
\bottomrule
\end{tabular}
\par\vspace{2pt}
\raggedright\footnotesize
{$^1$\textbf{Windmill}, $^2$\textbf{Church}, $^3$\textbf{Pagoda}, $^4$\textbf{Schloss}.}
\vspace{-0.3cm}
\end{table}

\begin{figure}[h]
\centering
\includegraphics[width=0.99\linewidth]{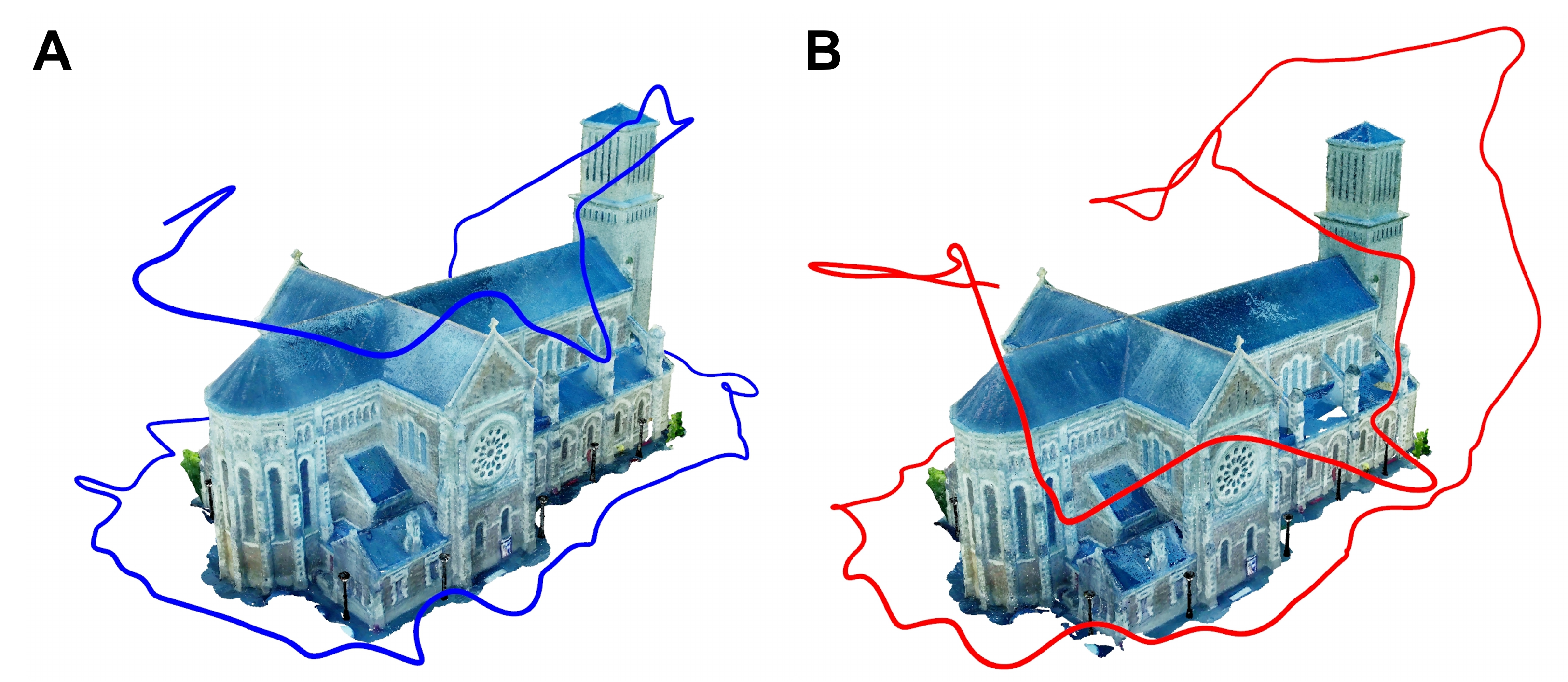}
\vspace{-0.3cm}
\caption{Upper-bound analysis with privileged geometry. (A) GT-guided flight trajectory and resulting 3D reconstruction. (B) Our trajectory and reconstruction from the same mission.}
\label{fig:GTVis}
\vspace{-0.2cm}
\end{figure}

\begin{figure*}[t]
\centering
\includegraphics[width=0.9\linewidth]{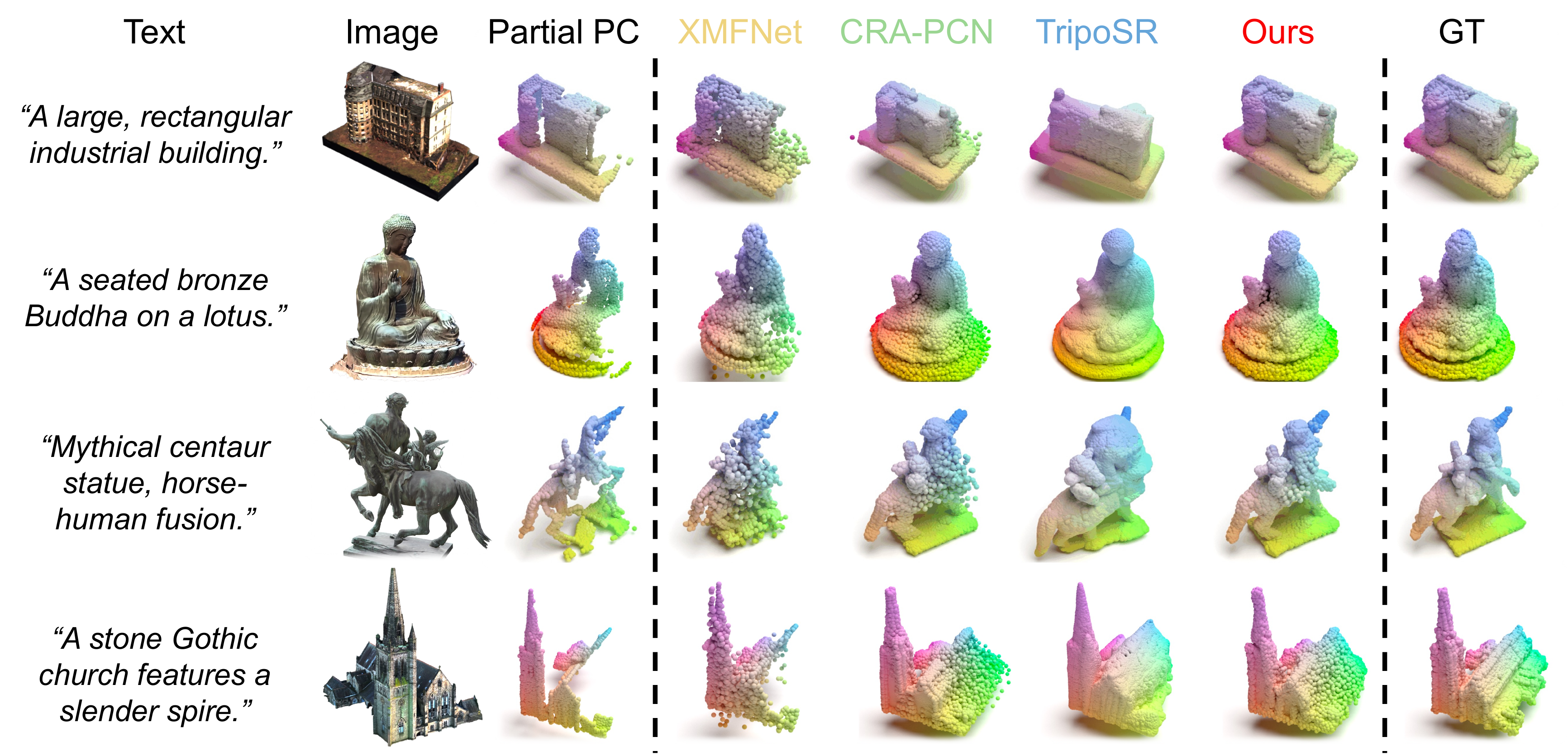}
\vspace{-0.3cm}
\caption{Qualitative comparison of surface prediction on four representative zero-shot test cases. For each row, the columns show (left$\rightarrow$right): (1) inputs, (2) predicted outputs from XMFNet, CRA-PCN, TripoSR, and our predictor, and (3) ground truth.}
\label{fig:ZeroShot}
\vspace{-0.5cm}
\end{figure*}

\paragraph{Upper-bound analysis with privileged geometry.}
We first probe a system-level question: \emph{Is our FM-empowered predictor informative enough to drive efficient and high-quality scanning behavior?}
To answer this, we replace the predicted surface used by the planner with the ground-truth target geometry, yielding a privileged-geometry variant (termed as \textbf{GT}) that approximates the practical performance ceiling where complete structure is available throughout the mission.
Following the experimental setup in Sec. \nameref{subsec:comparisons}, results in Tab. \ref{tab:GTComp} show that our prediction helps close most of the gap to this ceiling.
Compared with \textbf{GT}, our method proves remarkably competitive: only $20$-$35$ s slower in all four sites, corresponding to a modest $6.2\%$-$11.5\%$ increase in flight time.
Fig. \ref{fig:GTVis} further visualizes the scanning trajectory comparison in \textit{Church} scenario. 
Meanwhile, the data integrity drop is small: \textbf{F}ly\textbf{C}o maintains $93.83\%$-$96.50\%$ information completeness, staying within $2.4$-$4.0$ percentage points of \textbf{GT}'s near-perfect coverage.
Although \textbf{F}ly\textbf{C}o never observes the full structure and receives no prior map, its multi-modal predictor provides sufficiently reliable global structural foresight to guide planning almost as if the full structure were available.
The remaining gap to \textbf{GT} is expected, as it reflects the intrinsic advantage of having privileged geometry at all times, while the FM-empowered predictor already matches the majority of \textbf{GT}'s system-level benefit.
In other words, this upper-bound analysis experiment directly validates the powerful effectiveness of predictive capabilities in promoting open-world aerial scanning performance.

\begin{table}[h]
\footnotesize\sf\centering
\caption{Zero-shot surface prediction comparisons. We report averages on the zero-shot test set.}
\label{tab:PredComp}
\setlength{\tabcolsep}{5.2pt}
\renewcommand{\arraystretch}{1.1}
\begin{tabular}{l|ccc>{\columncolor{gray!15}}c}
\toprule
\textbf{Metrics} & \textbf{XMFNet} & \textbf{CRA-PCN} & \textbf{TripoSR} & \textbf{Ours} \\
\midrule
Inf. Lat. (ms)$\,\downarrow^{\dagger}$ &156.9 &158.2 &645.8 &\textbf{150.3} \\
\midrule
L1-CD$\,\downarrow$ &37.393 &31.053 &44.831 &\textbf{28.037} \\
\midrule
F-Score$\,\uparrow$ &0.601 &0.639 &0.569 &\textbf{0.705} \\
\bottomrule
\end{tabular}
\par\vspace{2pt}
\raggedright\footnotesize
{$^\dagger$Inference latency measured with a batch size of 1.}
\vspace{-0.35cm}
\end{table}

\paragraph{Zero-shot surface prediction comparisons.}
\phantomsection
\makeatletter
\def\@currentlabelname{Zero-shot surface prediction comparisons.}
\makeatother
\label{subsubsubsec:zero-shot}
The generalization of the predictive module is critical for bridging perception and planning in open-world applications.
To evaluate our predictor's zero-shot performance, we benchmark it against three representative SOTA models from 3D vision fields: CRA-PCN (\citealt{rong2024cra}) for point cloud completion, XMFNet (\citealt{aiello2022cross}) for cross-modal prediction, and TripoSR (\citealt{tochilkin2024triposr}) for 3D generation.
Since TripoSR cannot provide the metric scale, we align its output to the ground truth using ICP (rigid transform + scale) before evaluation.

The quantitative results in Tab. \ref{tab:PredComp} confirm that our predictor performs the best performance on both geometric fidelity (L1-CD) and correspondence (F-Score), while also being the fastest at inference.
Compared to TripoSR, we achieve a remarkable $16.794$-point improvement in L1-CD and a $23.90\%$ increase in F-score.
This significant margin highlights our method's ability to address the limitations of off-the-shelf generative models when applied to physical-scale tasks.
Against CRA-PCN, a point-cloud-only method, our predictor also surpasses the L1-CD by $3.018$ and the F-Score by $10.32\%$.
This advantage stems from our effective utilization of multi-modal visual and textual information beyond geometry-only inference, which helps mitigate ill-posedness in geometry-unobserved or weakly constrained regions where outliers or noisy predictions are prone to emerge (Fig. \ref{fig:ZeroShot}).
Even when compared to XMFNet, which also fuses cross-modal data, our approach maintains a clear lead, we outperform it by $9.356$ in L1-CD and $17.31\%$ in F-score.
The visualizations further reveal XMFNet's unsatisfactory generalization on out-of-distribution cases, where it fails to infer complete geometries (Fig. \ref{fig:ZeroShot}).
Moreover, our model also maintains the lowest single-sample latency ($150.3$ ms), supporting practical deployment in the closed-loop system.
In summary, these findings demonstrate our predictor's superior zero-shot generalization, which originates from its integration of world knowledge distilled from language and vision FMs.
This ability empowers the proposed system to robustly handle diverse and novel open-world scenarios with high adaptability and low cost.

\begin{table*}[t]
\footnotesize\sf\centering
\caption{Ablation experiments on key design choices of the predictor, reporting averages on the zero-shot test set.}
\label{tab:PredAblation}
\setlength{\tabcolsep}{5.2pt}
\renewcommand{\arraystretch}{1.1}
\begin{tabular}{l|cccc|c|cc|cc}
\toprule
\multirow{2}{*}{\textbf{Variants}}
& \multicolumn{4}{c|}{\textbf{Input Modalities}}
& \multirow{2}{*}{\textbf{Fusion Strategy}}
& \multicolumn{2}{c|}{\textbf{Supervision}}
& \multirow{2}{*}{\textbf{L1-CD}$\,\downarrow$} & \multirow{2}{*}{\textbf{F-Score}$\,\uparrow$} \\
& \textbf{Image} & \textbf{Cam. Param.} & \textbf{Text} & \textbf{Point Cloud}
&  
& $\mathcal{L}_\text{comp}$ & $\mathcal{L}_\text{par}$
& & \\
\midrule
w/o Image & \xmark & \cmark & \cmark & \cmark & Alternating Att.$^1$ & \cmark & \cmark & 31.565 & 0.620 \\
w/o Cam. Param. & \cmark & \xmark & \cmark & \cmark & Alternating Att.$^1$ & \cmark & \cmark & 30.747 & 0.636 \\
w/o Text & \cmark & \cmark & \xmark & \cmark & Alternating Att.$^1$ & \cmark & \cmark & 29.553 & 0.673 \\
\midrule
Fusion-CA & \cmark & \cmark & \cmark & \cmark & Cross-att. Only$^2$  & \cmark & \cmark & 33.479 & 0.614 \\
Fusion-GSA & \cmark & \cmark & \cmark & \cmark & Gloal Self-att.$^3$  & \cmark & \cmark & 32.628 & 0.621 \\
\midrule
w/o Partial Loss & \cmark & \cmark & \cmark & \cmark & Alternating Att.$^1$ & \cmark & \xmark & 30.578 & 0.637 \\
\midrule
\rowcolor{gray!15}
Ours (Full Model) & \cmark & \cmark & \cmark & \cmark & Alternating Att.$^1$ & \cmark & \cmark & \textbf{28.037} & \textbf{0.705} \\
\bottomrule
\end{tabular}
\par\vspace{2pt}
\raggedright\footnotesize
{$^1$Alternating attention, $^2$Cross-attention only, $^3$Global self-attention.}
\vspace{-0.3cm}
\end{table*}

\begin{figure}[h]
\centering
\includegraphics[width=0.99\linewidth]{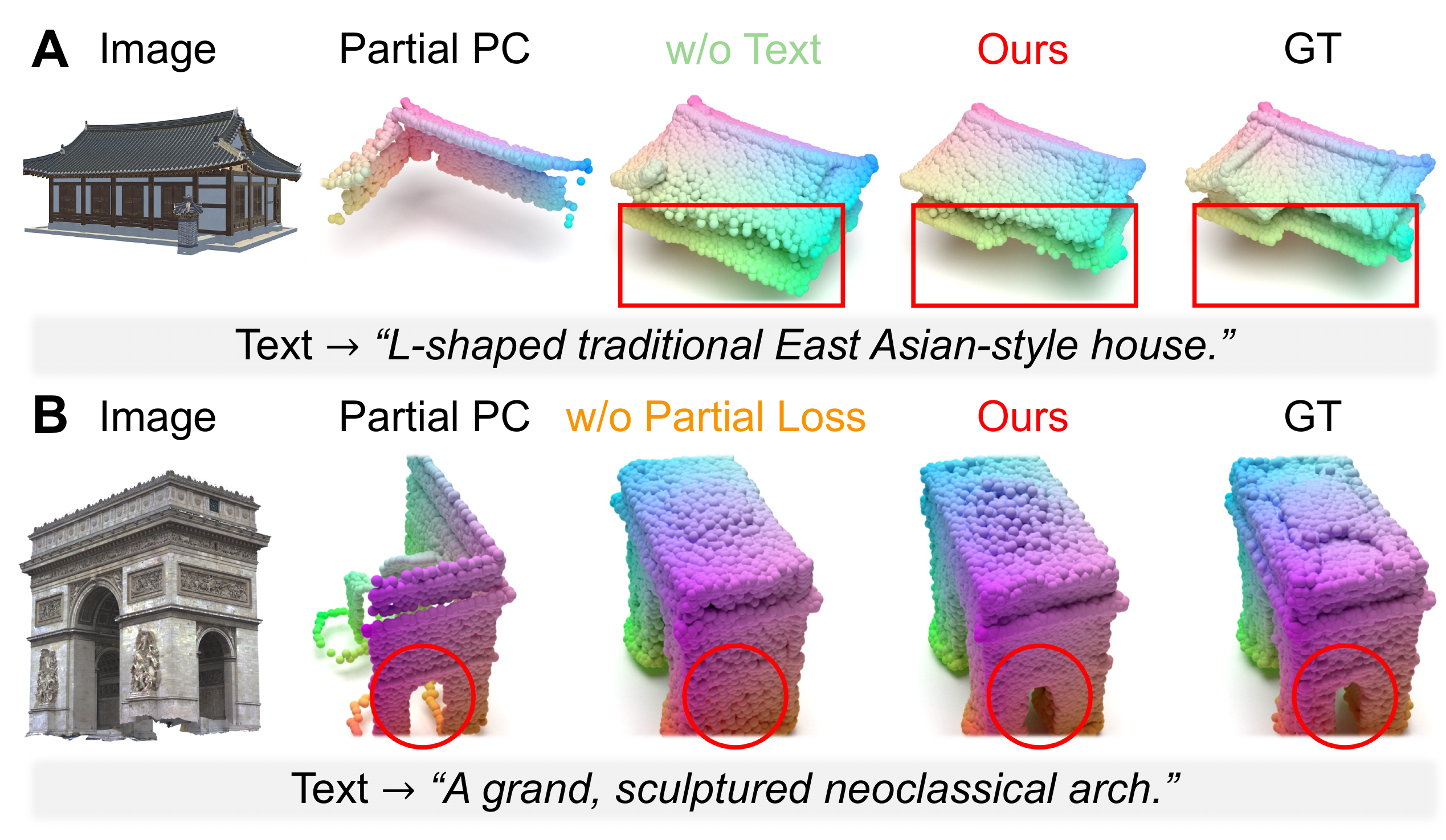}
\vspace{-0.3cm}
\caption{Qualitative prediction ablations. (A) Removing text inputs prevents leveraging global semantics. (B) Disabling partial-surface loss harms geometry consistency.}
\label{fig:PredExample}
\vspace{-0.45cm}
\end{figure}

\paragraph{Multi-modal conditioning.}
\phantomsection
\makeatletter
\def\@currentlabelname{Multi-modal conditioning.}
\makeatother
\label{subsubsubsec:modality}
In this experiment, we study whether prediction accuracy benefits from jointly conditioning on multi-modal inputs, beyond the basic partial geometric observations alone.
To this end, we train three ablated alternatives, each omitting one key input: (1) image, (2) camera parameters, or (3) text.
Our findings substantiate that the network effectively leverages each additional modality for a non-redundant positive gain (Tab. \ref{tab:PredAblation}).
Removing the image input leads to the largest drop, with the L1-CD increasing to $31.565$ and the F-Score decreasing to $0.620$.
In the absence of visual context, the model cannot refer to FM-derived visual priors and thus struggles to infer the correct surface geometry.
Omitting camera parameters also hurts performance (L1-CD rising to $30.747$ and the F-Score falling to $0.636$), as it prevents the model from reliably aligning image evidence with 3D space, causing noisy or malformed predictions.
The removal of text input degrades accuracy as well, with the L1-CD increasing to $29.553$ and the F-Score decreasing to $0.673$.
The qualitative results highlight the importance of this semantic guidance for high-level geometric control (Fig. \ref{fig:PredExample}A).
The textual prompt contains the crucial global structural descriptor ``\textit{L-shaped}''.
Without this text, the model defaults to reasoning a simple linear geometry from the ambiguous partial point cloud.
In contrast, thanks to the language FMs' precise understanding of the text input, our predictor correctly infers the overall layout, matching the ground truth.
Collectively, these ablations reveal that each modality provides unique and complementary information, critical for enabling generalizable surface predictions.

\paragraph{Alternating attention for knowledge fusion.}
\phantomsection
\makeatletter
\def\@currentlabelname{Alternating attention for knowledge fusion.}
\makeatother
\label{subsubsubsec:fusion}
We next examine how the fusion design affects the predictor's ability to incorporate foundation-model knowledge.
To isolate the contribution of our alternating-attention fusion, we construct two variants with identical trainable parameters: (1) \textit{Fusion-CA}, which removes all self-attention and performs fusion purely via cross-attention, and (2) \textit{Fusion-GSA}, which directly concatenates the encoded modality features and applies only global self-attention for mixing.
Tab. \ref{tab:PredAblation} exhibits that both alternatives markedly underperform our full model, demonstrating that alternating attention is more advantageous.
Concretely, \textit{Fusion-CA} yields a large degradation (L1-CD $33.479$, F-Score $0.614$), as removing self-attention limits cross-modal exchange to a single query-key-value interaction and weakens semantic alignment between vision and language priors before they are injected into the geometric stream.
Thus, the cross-attention stage receives less mutually consistent FM priors, and the predicted surface becomes less accurate.
\textit{Fusion-GSA} performs even worse (L1-CD $32.628$, F-Score $0.621$): while global self-attention permits unrestricted cross-modal mixing, it also blends modality information indiscriminately, injecting redundant or noisy context that is not necessarily relevant to completing the observed partial geometry.
Without an explicit step that allows the geometric stream selectively attend to the most useful FM cues, the fused representation becomes less task-aligned for surface prediction.
In contrast, alternating attention preserves rich cross-modal interaction while keeping fusion geometry-driven and selective, leading to more effective use of FM knowledge for precise surface completion.

\begin{table}[h]
\footnotesize\sf\centering
\caption{Effect of partial-surface regularization on flight efficiency. Results are averaged over 20 trials per scenario.}
\label{tab:PartialLossComp}
\setlength{\tabcolsep}{4.0pt}
\renewcommand{\arraystretch}{1.1}
\begin{tabular}{c|cccc}
\toprule
\multirow{2}{*}{\textbf{Variants}}
& \multicolumn{4}{c}{\textbf{Avg Flight Time (s)}$\,\downarrow$} \\
& \textbf{Windmill} & \textbf{Church} & \textbf{Pagoda} & \textbf{Schloss}\\
\midrule
w/o Partial Loss &418.8 &395.2 &261.7 &440.5 \\
\rowcolor{gray!15}
Ours &\textbf{329.4} &\textbf{336.6} &\textbf{225.1} &\textbf{363.9} \\
\bottomrule
\end{tabular}
\vspace{-0.3cm}
\end{table}

\begin{figure}[h]
\centering
\includegraphics[width=0.99\linewidth]{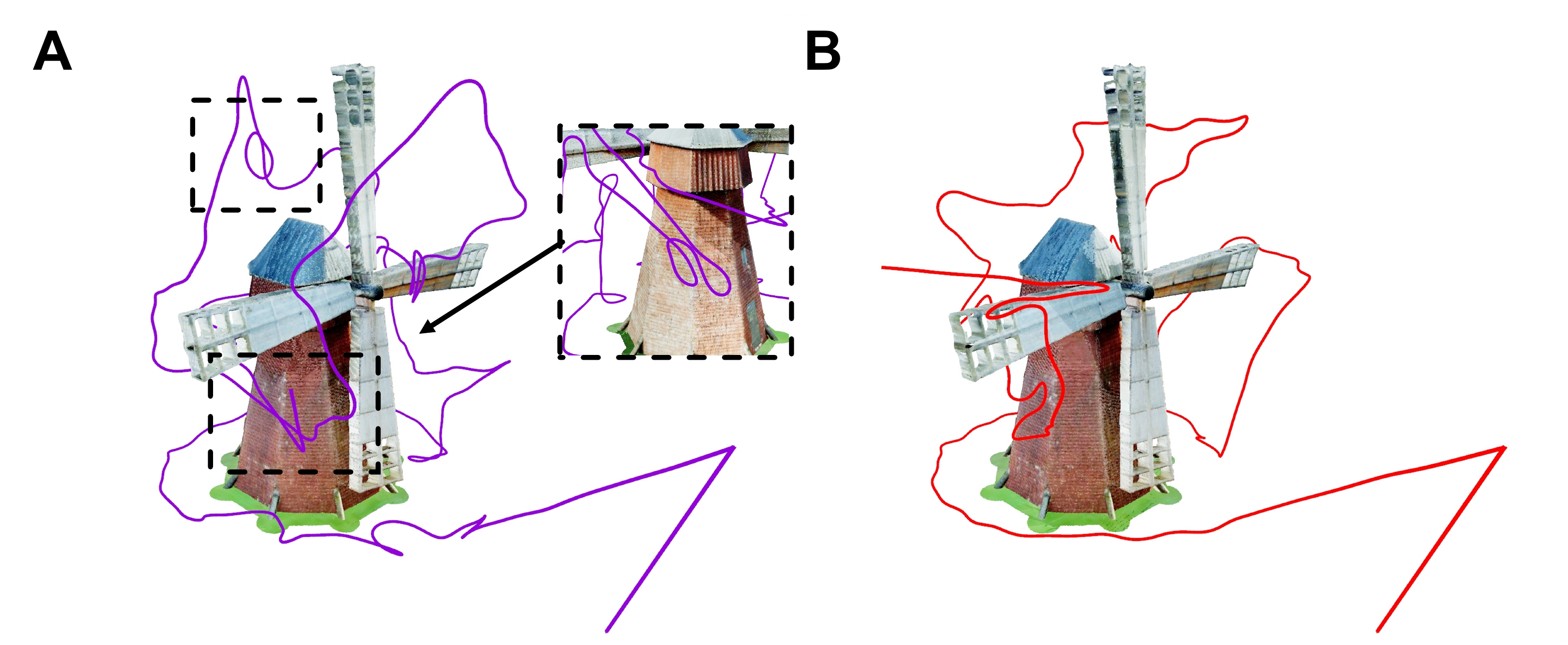}
\vspace{-0.3cm}
\caption{Qualitative impact of partial-surface regularization. (A) flight trajectory and resulting 3D reconstruction of \textit{w/o Partial Loss} variant. (B) Our trajectory and reconstruction from the same mission.}
\label{fig:PartialLossVis}
\vspace{-0.3cm}
\end{figure}

\paragraph{Partial-surface regularization.}
\phantomsection
\makeatletter
\def\@currentlabelname{Partial-surface regularization.}
\makeatother
\label{subsubsubsec:partial_loss}
Lastly, we examine the role of the partial-surface regularization that encourages consistency between the predicted surface and the input partial point cloud.
We train a variant by removing the partial-surface term from Eq. \eqref{eq:loss}.
As reported in Tab. \ref{tab:PartialLossComp}, this ablation noticeably degrades flight efficiency ($\sim16$-$27\%$), because temporally unstable and geometry-inconsistent predictions (in scale or shape) trigger redundant revisits and unnecessary detours during scanning; Fig. \ref{fig:PartialLossVis} visualizes a representative flight in the worst-affected scenario (\textit{Windmill}).
To understand the source of this behavior, we further evaluate prediction accuracy.
Removing the partial loss increases L1-CD by $2.541$ and reduces the F-Score from $0.705$ to $0.637$ (Tab. \ref{tab:PredAblation}).
The quantitative drop suggests that, without an explicit constraint from the observed geometry, the network is more prone to hallucinate local structures or ignore fine details present in the input.
For instance, in Fig. \ref{fig:PredExample}B, although the partial point cloud already reveals the two supporting pillars and the hollow region under the arch, the ablated model fails to preserve these concavities and instead predicts a spurious surface.
Such geometric artifacts directly mislead the planner to allocate viewpoints to non-existent regions, wasting flight time without improving coverage.
This experiment highlights that the partial loss regularization is essential, as it anchors the prediction to the latest measurements and promotes the prediction accuracy, thereby preventing prediction-induced inefficiency in downstream scanning.

\subsubsection{Ablations on Planning}
\label{subsubsec:ablation_planning}
Here we take a closer look at how each component of our prediction-aware hierarchical planner effectively integrates flight skills into our system, yielding superior performance in both efficiency and safety.
We first isolate the configuration-space construction step and validate its utility across the four simulated scenarios.
Afterwards, we conduct ablations on the remaining planner design choice, including (1) the temporal consistency awareness in the global planner, (2) the asynchronous computation framework, and (3) the viewpoint-constrained trajectory optimization.
Their experimental protocol and metrics are consistent with those in Sec. \nameref{subsec:comparisons}, with $20$ trials executed for each variant in every simulated scenario.

\begin{table}[h]
\footnotesize\sf\centering
\caption{Effect of the proposed configuration-space generation. Values denote per-scenario averages across 20 trials.}
\label{tab:FastGen}
\setlength{\tabcolsep}{4.0pt}
\renewcommand{\arraystretch}{1.1}
\begin{tabular}{l|l|cc>{\columncolor{gray!15}}c}
\toprule
\textbf{Metrics} & \textbf{Scenarios} & \textbf{SDF} & \textbf{Ray Casting} & \textbf{Ours} \\
\midrule

\multirow{4}{*}{Runtime (ms)$\,\downarrow$} 
& Windmill & 1214.8 & 28.5 & \textbf{11.3} \\
& Church   & 1514.4 & 36.6 & \textbf{15.2} \\
& Pagoda   & 923.2 & 17.9 & \textbf{14.7} \\
& Schloss  & 1390.0 & 44.1 & \textbf{14.5} \\
\midrule

\multirow{4}{*}{\makecell{Viewpoint\\Feasibility (\%)}$\,\uparrow$} 
& Windmill & 100.0 & 95.6 & \textbf{100.0} \\
& Church   & 100.0 & 93.7 & \textbf{100.0} \\
& Pagoda   & 100.0 & 98.1 & \textbf{100.0} \\
& Schloss  & 100.0 & 94.2 & \textbf{100.0} \\
\bottomrule
\end{tabular}
\vspace{-0.35cm}
\end{table}

\paragraph{Fast generation of safe configuration space.}
\phantomsection
\makeatletter
\def\@currentlabelname{Fast generation of safe configuration space.}
\makeatother
\label{subsubsubsec:cfg_gen}
We evaluate the proposed fast configuration-space determination (Sec. \nameref{subsubsubsec:fast_gen}) by replacing it with two common alternatives: SDF-based queries and ray casting against the predicted mesh.
All three variants use identical viewpoint sampling settings.
We measure (1) computational cost, reported as wall-clock runtime per update including both configuration-space construction and viewpoint checking, and (2) viewpoint feasibility, defined as the fraction of generated viewpoints that lie in free space outside the target geometry (\textit{i.e.}, no viewpoints are allowed inside the target volume). 
Tab. \ref{tab:FastGen} summarizes the results.
Our solution achieves the lowest runtime in all four scenarios, taking only $11.3$-$15.2$ ms per update, compared with $17.9$-$44.1$ ms for ray casting and $923.2$-$1514.4$ ms for SDF queries.
This translates to a $1.2$-$3.0\times$ speedup over ray casting and a two-order-of-magnitude improvement over SDF-based fashion, while under the same structural complexity.
Meanwhile, our method maintains $100\%$ viewpoint feasibility across all cases.
In contrast, ray casting occasionally produces infeasible viewpoints (feasibility $93.7\%$-$98.1\%$), which can lead to physically unreachable viewpoint requests and compromise downstream safety guarantees.
Overall, this study confirms that the proposed method is both computationally lightweight and strictly feasible, making it well-suited for viewpoint generation in complex 3D structures.

\begin{table*}[t]
\footnotesize\sf\centering
\caption{Ablation experiments on key design choices of the planner. For each scenario, we report mean (Avg), standard deviation (Std), maximum (Max), and minimum (Min) over 20 trials.}
\label{tab:PlanComp}
\setlength{\tabcolsep}{5.2pt}
\renewcommand{\arraystretch}{1.1}
\begin{tabular}{c|c|cccc|cccc|c|c}
\toprule
\multirow{2}{*}{\textbf{Scenarios}} & \multirow{2}{*}{\textbf{Variants}$^{\dagger}$}
& \multicolumn{4}{c|}{\textbf{Flight Time (s)}$\,\downarrow$}
& \multicolumn{4}{c|}{\textbf{Info. Completeness (\%)}$\,\uparrow$}
& \multirow{2}{*}{\makecell{\textbf{Avg Response}\\\textbf{Latency (ms)}}$\,\downarrow$}
& \multirow{2}{*}{\textbf{Success Rate}$\,\uparrow$} \\
& & \textbf{Avg} & \textbf{Std} & \textbf{Max} & \textbf{Min} & \textbf{Avg} & \textbf{Std} & \textbf{Max} & \textbf{Min} & & \\
\midrule

\multirow{4}{*}{Windmill}
& w/o CAGP   &543.4  &79.95  &713.9  &437.8  &93.57  &2.69  &96.82  &88.16  &26.54 &20/20  \\
& w/o Async. &363.8  &12.68  &388.1  &348.5  &95.50  &\textbf{0.45}  &96.43  &94.57  &270.91 &15/20  \\
& w/o VCTO   &432.7  &23.64  &504.5  &399.6  &95.18  &1.38  &97.39  &92.02 &\textbf{23.12} &20/20  \\
\rowcolor{gray!15}
\cellcolor{white} & Ours                 &\textbf{329.4}  &\textbf{8.93}  &357.9  &321.4  &\textbf{96.04}  &0.58  &97.11  &94.88  &25.04 &\textbf{20/20}  \\
\midrule

\multirow{4}{*}{Church}
& w/o CAGP   &598.3  &114.38  &846.7  &477.8  &91.48  &1.87  &95.74  &89.05  &27.93 &19/20  \\
& w/o Async. &350.8  &11.71  &372.1  &330.7  &\textbf{94.79}  &0.74  &95.78  &92.81  &290.37 &9/20  \\
& w/o VCTO   &421.6  &13.06  &441.6  &401.1  &92.92  &2.01  &95.23  &90.05  &\textbf{24.07} &19/20  \\
\rowcolor{gray!15}
\cellcolor{white} & Ours                 &\textbf{336.6}  &\textbf{9.43}  &364.8  &323.6  &94.53  &\textbf{0.71}  &95.54  &93.23  &25.18 &\textbf{20/20}  \\
\midrule

\multirow{4}{*}{Pagoda}
& w/o CAGP   &372.5  &51.57  &436.1  &295.4  &91.84  &1.52  &94.56  &88.47  &27.85 &19/20  \\
& w/o Async. &243.4  &15.43  &269.2  &218.6  &93.65  &0.97  &95.13  &92.60  &202.65 &10/20  \\
& w/o VCTO   &240.0  &10.87  &266.7  &231.8  &93.08  &0.69  &94.86  &92.15  &25.52 &\textbf{20/20}  \\
\rowcolor{gray!15}
\cellcolor{white} & Ours                 &\textbf{225.1}  &\textbf{6.22}  &240.3  &216.2  &\textbf{93.83}  &\textbf{0.63}  &94.62  &92.33  &\textbf{24.39} &19/20  \\
\midrule

\multirow{4}{*}{Schloss}
& w/o CAGP   &671.9  &117.33  &941.2  &510.3  &95.48  &2.27  &98.21  &93.76 &29.26 &20/20  \\
& w/o Async. &377.1  &9.47  &391.6  &357.2  &\textbf{96.87}  &0.84  &97.79  &95.44  &266.39 &14/20  \\
& w/o VCTO   &415.3  &16.42  &441.5  &374.2  &95.27  &1.09  &97.16  &93.40  &\textbf{23.01} &20/20  \\
\rowcolor{gray!15}
\cellcolor{white} & Ours                 &\textbf{363.9}  &\textbf{7.52}  &380.8  &349.5  &96.50  &\textbf{0.66}  &97.58  &95.45  &24.27 &\textbf{20/20}  \\
\bottomrule
\end{tabular}
\par\vspace{2pt}
\raggedright\footnotesize
{$^{\dagger}$\textbf{CAGP}: consistency awareness in global planning; \textbf{Async.}: asynchronous computation framework; \textbf{VCTO}: viewpoint-constrained traj. opt.}
\vspace{-0.4cm}
\end{table*}

\begin{figure*}[h]
\centering
\includegraphics[width=0.99\linewidth]{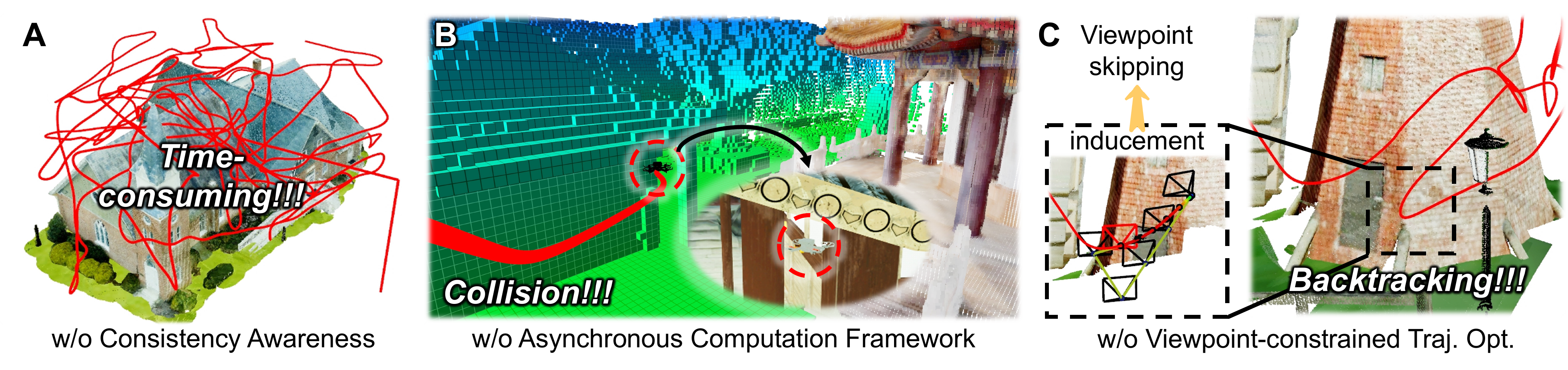}
\vspace{-0.4cm}
\caption{Qualitative effects of planner ablations. (A) A tangled, time-consuming trajectory resulting from the removal of temporal consistency awareness. (B) Synchronous execution inflates response latency and causes a collision. (C) Deactivating viewpoint constraints introduces backtracking detours to recover missed coverage.}
\label{fig:PlanVis}
\vspace{-0.5cm}
\end{figure*}

\paragraph{Consistency-aware global planning.}
\phantomsection
\makeatletter
\def\@currentlabelname{Consistency-aware global planning.}
\makeatother
\label{subsubsubsec:consistency}
In this study, we disable all consistency-promoting designs in the global planner, forcing it to re-generate the viewpoint set and re-compute the visiting sequence from scratch at every planning cycle.
In other words, this setting reverts our global planning to off-the-shelf memoryless 3D coverage planners that optimize each cycle in isolation, without accounting for the flight history.
The results show that removing these operations severely degrades mission efficiency, leading to less predictable and significantly longer flight times.
As quantified in Tab. \ref{tab:PlanComp}, the average flight time increases substantially across all scenarios: by $214.0$ seconds in \textit{Windmill} ($+64.9\%$), $261.7$ seconds in \textit{Church} ($+77.7\%$), $147.4$ seconds in \textit{Pagoda} ($+65.5\%$), and $308.0$ seconds in the \textit{Schloss} ($+84.6\%$).
Furthermore, the flight-time standard deviations reveal a drastic increase in the variance of flight times, indicating that the resulting trajectories are not only longer on average but also highly erratic and unpredictable.
This inefficiency is caused by frequent and unnecessary reshuffling of viewpoint distributions and their visiting order, which overlooks global temporal consistency with historical paths, resulting in excessive back-and-forth maneuvers, as qualitatively illustrated by the tangled trajectory in Fig. \ref{fig:PlanVis}A.

\paragraph{Asynchronous computation framework.}
\phantomsection
\makeatletter
\def\@currentlabelname{Asynchronous computation framework.}
\makeatother
\label{subsubsubsec:async}
Here, we assess the benefit of our decoupled, asynchronous architecture by replacing it with a conventional synchronous, serial execution scheme.
Under this configuration, the high-frequency local planner must wait for the computationally intensive global planner to finish each cycle before producing a new trajectory.
This enforced blocking severely compromises system responsiveness and proves critically detrimental to flight safety.
As a result, the mission success rate drops sharply from $98.8\%$ to $60.0\%$ (Tab. \ref{tab:PlanComp}).
The surge in failures directly stems from the system's reduced ability to respond to unexpected obstacles.
Specifically, as reported in Tab. \ref{tab:PlanComp}, the average response latency increases by nearly an order of magnitude—from approximately $30$ ms to $200$-$300$ ms.
Such excessive latency prevents the drone from reacting in time to obstacles that emerge between planning cycles, leading to frequent collisions, as qualitatively illustrated in Fig. \ref{fig:PlanVis}B.
These results confirm that asynchronous design is essential for maintaining real-time responsiveness and ensuring flight safety.

\paragraph{Viewpoint-constrained trajectory optimization.}
\phantomsection
\makeatletter
\def\@currentlabelname{Viewpoint-constrained trajectory optimization.}
\makeatother
\label{subsubsubsec:vcto}
This analysis validates our local trajectory generation strategy, which enforces that the trajectory preserves the structural coverage prescribed by the global path.
To this end, we evaluate a variant where this hard constraint is removed, allowing the optimizer to freely adjust intermediate points as movable waypoints rather than requiring the trajectory to pass through all designated viewpoints.
As listed in Tab. \ref{tab:PlanComp}, this relaxed alternative eventually reaches comparable information completeness, but at a clear cost in mission efficiency.
Without the constraint, the average flight time expands across all scenarios without this constraint, respectively by $31.4\%$, $25.3\%$, $6.6\%$, and $14.1\%$ (Tab. \ref{tab:PlanComp}).
This inefficiency arises from suboptimal behaviors like backtracking; the planner might initially generate a shorter, smoother trajectory by skipping some crucial viewpoints, only to discover later that a target surface region was missed to scan. 
This forces the planner to produce a long, inefficient detour to return to the vicinity of the skipped viewpoints to complete coverage (see Fig. \ref{fig:PlanVis}C for an example).
This evaluation demonstrates that our viewpoint-constrained optimization is a more efficient strategy, balancing the requirements of data integrity and minimizing flight time.

\section{Conclusion}
\label{sec:conclusion}

In this work, we study the aerial 3D target structure scanning problem in open-world environments and answer a central system-level question: \textit{What architecture is currently well-suited for this problem?}

We find that dominant paradigms remain constrained by either effortful human priors or restrictive assumptions, lack scene understanding that is necessary for high efficiency, and afford limited safety assurance in cluttered settings.
These limitations hinder practical viability across diverse scanning missions.
To this end, we propose \textbf{F}ly\textbf{C}o, a tailored perception-prediction-planning architecture that efficiently leverages foundation models to enable fully autonomous, prompt-driven target scanning in various, unknown open-world scenes, requiring only low-effort texts and visual annotations from human inputs.
Comprehensive evaluations, encompassing both in-the-wild demonstrations and quantitative simulations, substantiate that the proposed system demonstrates significant advantages over existing paradigms and represents a more suitable recipe for this open-world aerial scanning task.

This superior performance is rooted in the synergistic and efficient integration of three FM-empowered components that jointly close the loop from prompt to scanning flight.
First, perception utilizes the world understanding of FMs to robustly ground abstract prompts onto physical targets without requiring manually defined operational zones, overcoming a key usability barrier of existing works when deployed in the open world.
Second, by distilling knowledge in FMs, the partial-to-complete predictor analyzes multi-modal cues and provides global foresight guidance to suppress redundant scanning, while its strong zero-shot generalization ensures system reliability in novel environments.
Third, the prediction-aware hierarchical planner converts this foresight into efficient and safe trajectories, achieving $1.25$-$3.0\times$ efficiency gains with higher information completeness over competing baselines, while maintaining real-time responsiveness for safety-critical obstacle avoidance. 
In summary, \textbf{F}ly\textbf{C}o offers a practical and extensible system blueprint: it seamlessly incorporates the powerful understanding and inherent knowledge of FMs with advanced flight skills into aerial robots without expensive data requirements while keeping online components within real-time budgets, and it supports flexible modular replacement to naturally accommodate future progress in FMs and robotics planning.
The full system implementation and development configurations will be released to enable reproducibility and future research.

Beyond this, \textbf{F}ly\textbf{C}o enables a general capability for prompt-grounded, targeted 3D data acquisition in complex real environments, where manual specification of operational volumes or exhaustive exploration is undesirable.
This can support a range of field workflows that require repeatable, structure-centric capture, \textit{e.g.}, infrastructure inspection, cultural-heritage documentation, and ecological habitat monitoring.
By turning lightweight human intent into safe and efficient scanning trajectories, our system may reduce operator burden and improve the scalability of collecting task-relevant 3D observations.
At the same time, realizing this potential broadly calls for further advances on several fronts, which we discuss next as limitations and future directions.

\subsection{Limitations and Future Work}

\subsubsection{Limited utilization of textual information.}
Although text prompts play a crucial role in grounding user intent, their expressive power is not yet fully exploited in the current system.
First, we intentionally restrict the textual input to short, structured phrases describing salient target attributes.
This design choice improves robustness and reduces ambiguity in open-world deployments, but it also limits the expressiveness of free-form natural language and constrains how flexibly users can specify complex intent.
Second, the quantitative gains brought by textual conditioning in surface prediction remain marginal compared with geometric and visual cues.
A key reason lies in data imbalance during training: a single 3D model can generate many partial point clouds and image observations, all paired with the same textual description.
This many-to-one mapping biases learning towards geometry and vision while weakening the relative influence of language.
Similar dilemma has been observed in recent vision-language-action models (VLA), where language often serves as a weak signal rather than a dominant driver (\citealt{zitkovich2023rt}).
Future work will support richer language forms, explore improved data balancing strategies, and tighter cross-modal alignment objectives to better elevate the impact of text.

\subsubsection{Dependence on offboard computation for prediction.}
At present, the surface prediction module runs on a ground device rather than fully onboard the aerial platform.
This choice is motivated by two considerations: prediction is not strictly time-critical for closed-loop flight, and incorporating FMs incurs substantial computational overhead that makes sub-second onboard inference impractical on current edge hardware.
While reliable communication is available in most outdoor scanning scenarios, this dependency may become a limiting factor in communication-constrained environments such as underground mines, enclosed industrial facilities, or electromagnetically disturbed sites.
Therefore, fully onboard deployment is a necessary step towards broader applicability.
An important future direction is to compress and distill foundation-model knowledge into lighter-weight predictors, enabling accurate yet efficient inference on edge devices and eliminating reliance on external communication.

\subsubsection{Scalability beyond single-drone, single-target missions.}
The current system is constrained by the endurance of a single aerial robot and is thus designed to scan one target structure per mission.
Scaling this capability to city-scale or large natural environments calls for coordinated multi-drone operation.
This introduces new challenges, including dynamic high-level task allocation, inter-robot coordination, and real-time reassignment of scanning objectives based on partial observations.
Future work will investigate multi-agent extensions of the proposed architecture, where multiple drones collaboratively divide target structures, adaptively replan assignments, and jointly minimize overall mission time while maintaining safety and coverage guarantees.
Such extensions would upgrade the system from single-structure autonomy to large-scale, cooperative open-world scanning.

\begin{acks}
We thank Fei Gao for valuable suggestions to the manuscript.
We also thank Weiye Zhang and Mingjie Zhang for their early investigation on hardware platform selection.
\end{acks}

\bibliographystyle{SageH}
\bibliography{references}

@misc{infrasturcture,
  author       = {BeyondVision},
  title			   = {Transforming Construction and Infrastructure: Discover Amazing Construction Site Drones Surveillance}, 
  year			   = {2023},
  note			   = {https://www.youtube.com/watch?v=Np1CJ1uWEGU}
}

@misc{disaster_management,
  author       = {Esri},
  title			   = {Mapping Disaster Response}, 
  year			   = {2023},
  url			     = {https://www.youtube.com/watch?v=IqHSdvTukfc}
}

@misc{cultural_preservation,
  author       = {Mapalytics},
  title			   = {Heritage Preservation - Digital Double - UNESCO - Mapalytics - 3D Mapping}, 
  year		     = {2024},
  url			     = {https://www.youtube.com/watch?v=Svzd291ESSY}
}

@misc{djiflightplanner,
  title        = {DJIFlightPlanner},
  author       = {DJI},
  note         = {https://www.djiflightplanner.com/},
  year         = {2024}
}

@misc{djiterra,
  title        = {DJI Terra},
  author       = {DJI},
  note         = {https://enterprise.dji.com/dji-terra},
  year         = {2025}
}

@misc{pix4dcapture,
  title        = {PIX4Dcapture Pro},
  author       = {PIX4D},
  note         = {https://www.pix4d.com/product/pix4dcapture/},
  year         = {2024}
}

@misc{flylitchi,
  title        = {Litchi},
  author       = {Litchi},
  note         = {https://flylitchi.com/},
  year         = {2025}
}

@misc{skydio,
  title        = {Skydio Software},
  author       = {Skydio},
  note         = {https://www.skydio.com/software},
  year         = {2025}
}

@inproceedings{roberts2017submodular,
  title={Submodular trajectory optimization for aerial 3d scanning},
  author={Roberts, Mike and Dey, Debadeepta and Truong, Anh and Sinha, Sudipta and Shah, Shital and Kapoor, Ashish and Hanrahan, Pat and Joshi, Neel},
  booktitle={Proceedings of the IEEE International Conference on Computer Vision},
  pages={5324--5333},
  year={2017}
}

@article{hepp2018plan3d,
  title={Plan3d: Viewpoint and trajectory optimization for aerial multi-view stereo reconstruction},
  author={Hepp, Benjamin and Nie{\ss}ner, Matthias and Hilliges, Otmar},
  journal={ACM Transactions on Graphics (TOG)},
  volume={38},
  number={1},
  pages={1--17},
  year={2018},
  publisher={ACM New York, NY, USA}
}

@article{smith2018aerial,
  title={Aerial path planning for urban scene reconstruction: a continuous optimization method and benchmark},
  author={Smith, Neil and Moehrle, Nils and Goesele, Michael and Heidrich, Wolfgang},
  journal={ACM Transactions on Graphics (TOG)},
  volume={37},
  number={6},
  pages={1--15},
  year={2018},
  publisher={ACM New York, NY, USA}
}

@inproceedings{peng2019adaptive,
  title={Adaptive view planning for aerial 3D reconstruction},
  author={Peng, Cheng and Isler, Volkan},
  booktitle={2019 International Conference on Robotics and Automation (ICRA)},
  pages={2981--2987},
  year={2019},
  organization={IEEE}
}

@inproceedings{feng2024fc,
  title={Fc-planner: A skeleton-guided planning framework for fast aerial coverage of complex 3d scenes},
  author={Feng, Chen and Li, Haojia and Zhang, Mingjie and Chen, Xinyi and Zhou, Boyu and Shen, Shaojie},
  booktitle={2024 IEEE International Conference on Robotics and Automation (ICRA)},
  pages={8686--8692},
  year={2024},
  organization={IEEE}
}

@inproceedings{jing2016view,
  title={View planning for 3d shape reconstruction of buildings with unmanned aerial vehicles},
  author={Jing, Wei and Polden, Joseph and Tao, Pey Yuen and Lin, Wei and Shimada, Kenji},
  booktitle={2016 14th International Conference on Control, Automation, Robotics and Vision (ICARCV)},
  pages={1--6},
  year={2016},
  organization={IEEE}
}

@article{zhou2020offsite,
  title={Offsite aerial path planning for efficient urban scene reconstruction},
  author={Zhou, Xiaohui and Xie, Ke and Huang, Kai and Liu, Yilin and Zhou, Yang and Gong, Minglun and Huang, Hui},
  journal={ACM Transactions on Graphics (TOG)},
  volume={39},
  number={6},
  pages={1--16},
  year={2020},
  publisher={ACM New York, NY, USA}
}

@article{zhang2021continuous,
  title={Continuous aerial path planning for 3D urban scene reconstruction.},
  author={Zhang, Han and Yao, Yucong and Xie, Ke and Fu, Chi-Wing and Zhang, Hao and Huang, Hui},
  journal={ACM Trans. Graph.},
  volume={40},
  number={6},
  pages={225--1},
  year={2021}
}

@inproceedings{song2017online,
  title={Online inspection path planning for autonomous 3D modeling using a micro-aerial vehicle},
  author={Song, Soohwan and Jo, Sungho},
  booktitle={2017 IEEE International Conference on Robotics and Automation (ICRA)},
  pages={6217--6224},
  year={2017},
  organization={IEEE}
}

@article{bircher2018receding,
  title={Receding horizon path planning for 3D exploration and surface inspection},
  author={Bircher, Andreas and Kamel, Mina and Alexis, Kostas and Oleynikova, Helen and Siegwart, Roland},
  journal={Autonomous Robots},
  volume={42},
  pages={291--306},
  year={2018},
  publisher={Springer}
}

@article{song2021view,
  title={View path planning via online multiview stereo for 3-d modeling of large-scale structures},
  author={Song, Soohwan and Kim, Daekyum and Choi, Sunghee},
  journal={IEEE Transactions on Robotics},
  volume={38},
  number={1},
  pages={372--390},
  year={2021},
  publisher={IEEE}
}

@article{luo2024star,
  title={Star-searcher: A complete and efficient aerial system for autonomous target search in complex unknown environments},
  author={Luo, Yiming and Zhuang, Zixuan and Pan, Neng and Feng, Chen and Shen, Shaojie and Gao, Fei and Cheng, Hui and Zhou, Boyu},
  journal={IEEE Robotics and Automation Letters},
  year={2024},
  publisher={IEEE}
}

@inproceedings{yamauchi1997frontier,
  title={A frontier-based approach for autonomous exploration},
  author={Yamauchi, Brian},
  booktitle={Proceedings 1997 IEEE International Symposium on Computational Intelligence in Robotics and Automation CIRA'97.'Towards New Computational Principles for Robotics and Automation'},
  pages={146--151},
  year={1997},
  organization={IEEE}
}

@inproceedings{connolly1985determination,
  title={The determination of next best views},
  author={Connolly, Cl},
  booktitle={Proceedings. 1985 IEEE international conference on robotics and automation},
  volume={2},
  pages={432--435},
  year={1985},
  organization={IEEE}
}

@inproceedings{feng2023predrecon,
  title={Predrecon: A prediction-boosted planning framework for fast and high-quality autonomous aerial reconstruction},
  author={Feng, Chen and Li, Haojia and Gao, Fei and Zhou, Boyu and Shen, Shaojie},
  booktitle={2023 IEEE International Conference on Robotics and Automation (ICRA)},
  pages={1207--1213},
  year={2023},
  organization={IEEE}
}

@inproceedings{guedon2023macarons,
  title={MACARONS: mapping and coverage anticipation with RGB online self-supervision},
  author={Gu{\'e}don, Antoine and Monnier, Tom and Monasse, Pascal and Lepetit, Vincent},
  booktitle={Proceedings of the IEEE/CVF Conference on Computer Vision and Pattern Recognition},
  pages={940--951},
  year={2023}
}

@inproceedings{chen2024gennbv,
  title={Gennbv: Generalizable next-best-view policy for active 3d reconstruction},
  author={Chen, Xiao and Li, Quanyi and Wang, Tai and Xue, Tianfan and Pang, Jiangmiao},
  booktitle={Proceedings of the IEEE/CVF Conference on Computer Vision and Pattern Recognition},
  pages={16436--16445},
  year={2024}
}

@article{liu2024dg,
  title={DG-NBV: A Cognitive Framework for Direct Generation of Next Best View in Continuous View Space},
  author={Liu, Zhicheng and Cao, Zhiqiang and Li, Jianjie and Guan, Peiyu and Yu, Junzhi},
  journal={IEEE Transactions on Cognitive and Developmental Systems},
  year={2024},
  publisher={IEEE}
}

@inproceedings{yuan2018pcn,
  title={Pcn: Point completion network},
  author={Yuan, Wentao and Khot, Tejas and Held, David and Mertz, Christoph and Hebert, Martial},
  booktitle={2018 international conference on 3D vision (3DV)},
  pages={728--737},
  year={2018},
  organization={IEEE}
}

@inproceedings{rong2024cra,
  title={Cra-pcn: Point cloud completion with intra-and inter-level cross-resolution transformers},
  author={Rong, Yi and Zhou, Haoran and Yuan, Lixin and Mei, Cheng and Wang, Jiahao and Lu, Tong},
  booktitle={Proceedings of the AAAI Conference on Artificial Intelligence},
  volume={38},
  number={5},
  pages={4676--4685},
  year={2024}
}

@article{aiello2022cross,
  title={Cross-modal learning for image-guided point cloud shape completion},
  author={Aiello, Emanuele and Valsesia, Diego and Magli, Enrico},
  journal={Advances in Neural Information Processing Systems},
  volume={35},
  pages={37349--37362},
  year={2022}
}

@article{tochilkin2024triposr,
  title={Triposr: Fast 3d object reconstruction from a single image},
  author={Tochilkin, Dmitry and Pankratz, David and Liu, Zexiang and Huang, Zixuan and Letts, Adam and Li, Yangguang and Liang, Ding and Laforte, Christian and Jampani, Varun and Cao, Yan-Pei},
  journal={arXiv preprint arXiv:2403.02151},
  year={2024}
}

@article{li2025step1x,
  title={Step1x-3d: Towards high-fidelity and controllable generation of textured 3d assets},
  author={Li, Weiyu and Zhang, Xuanyang and Sun, Zheng and Qi, Di and Li, Hao and Cheng, Wei and Cai, Weiwei and Wu, Shihao and Liu, Jiarui and Wang, Zihao and others},
  journal={arXiv preprint arXiv:2505.07747},
  year={2025}
}

@article{ahn2022can,
  title={Do as i can, not as i say: Grounding language in robotic affordances},
  author={Ahn, Michael and Brohan, Anthony and Brown, Noah and Chebotar, Yevgen and Cortes, Omar and David, Byron and Finn, Chelsea and Fu, Chuyuan and Gopalakrishnan, Keerthana and Hausman, Karol and others},
  journal={arXiv preprint arXiv:2204.01691},
  year={2022}
}

@article{driess2023palm,
  title={Palm-e: An embodied multimodal language model},
  author={Driess, Danny and Xia, Fei and Sajjadi, Mehdi SM and Lynch, Corey and Chowdhery, Aakanksha and Wahid, Ayzaan and Tompson, Jonathan and Vuong, Quan and Yu, Tianhe and Huang, Wenlong and others},
  year={2023}
}

@article{zhang2024uni,
  title={Uni-navid: A video-based vision-language-action model for unifying embodied navigation tasks},
  author={Zhang, Jiazhao and Wang, Kunyu and Wang, Shaoan and Li, Minghan and Liu, Haoran and Wei, Songlin and Wang, Zhongyuan and Zhang, Zhizheng and Wang, He},
  journal={arXiv preprint arXiv:2412.06224},
  year={2024}
}

@inproceedings{zitkovich2023rt,
  title={Rt-2: Vision-language-action models transfer web knowledge to robotic control},
  author={Zitkovich, Brianna and Yu, Tianhe and Xu, Sichun and Xu, Peng and Xiao, Ted and Xia, Fei and Wu, Jialin and Wohlhart, Paul and Welker, Stefan and Wahid, Ayzaan and others},
  booktitle={Conference on Robot Learning},
  pages={2165--2183},
  year={2023},
  organization={PMLR}
}

@article{kim2024openvla,
  title={Openvla: An open-source vision-language-action model},
  author={Kim, Moo Jin and Pertsch, Karl and Karamcheti, Siddharth and Xiao, Ted and Balakrishna, Ashwin and Nair, Suraj and Rafailov, Rafael and Foster, Ethan and Lam, Grace and Sanketi, Pannag and others},
  journal={arXiv preprint arXiv:2406.09246},
  year={2024}
}

@article{black2410pi0,
  title={$\pi$0: A vision-language-action flow model for general robot control. CoRR, abs/2410.24164, 2024. doi: 10.48550},
  author={Black, Kevin and Brown, Noah and Driess, Danny and Esmail, Adnan and Equi, Michael and Finn, Chelsea and Fusai, Niccolo and Groom, Lachy and Hausman, Karol and Ichter, Brian and others},
  journal={arXiv preprint ARXIV.2410.24164},
  year={2024}
}

@article{zhou2025vision,
  title={Vision-Language-Action Model with Open-World Embodied Reasoning from Pretrained Knowledge},
  author={Zhou, Zhongyi and Zhu, Yichen and Wen, Junjie and Shen, Chaomin and Xu, Yi},
  journal={arXiv preprint arXiv:2505.21906},
  year={2025}
}

@article{ravi2024sam,
  title={Sam 2: Segment anything in images and videos},
  author={Ravi, Nikhila and Gabeur, Valentin and Hu, Yuan-Ting and Hu, Ronghang and Ryali, Chaitanya and Ma, Tengyu and Khedr, Haitham and R{\"a}dle, Roman and Rolland, Chloe and Gustafson, Laura and others},
  journal={arXiv preprint arXiv:2408.00714},
  year={2024}
}

@inproceedings{wang2023image,
  title={Image as a foreign language: Beit pretraining for vision and vision-language tasks},
  author={Wang, Wenhui and Bao, Hangbo and Dong, Li and Bjorck, Johan and Peng, Zhiliang and Liu, Qiang and Aggarwal, Kriti and Mohammed, Owais Khan and Singhal, Saksham and Som, Subhojit and others},
  booktitle={Proceedings of the IEEE/CVF Conference on Computer Vision and Pattern Recognition},
  pages={19175--19186},
  year={2023}
}

@article{zhang2024evf,
  title={Evf-sam: Early vision-language fusion for text-prompted segment anything model},
  author={Zhang, Yuxuan and Cheng, Tianheng and Zhu, Lianghui and Hu, Rui and Liu, Lei and Liu, Heng and Ran, Longjin and Chen, Xiaoxin and Liu, Wenyu and Wang, Xinggang},
  journal={arXiv preprint arXiv:2406.20076},
  year={2024}
}

@inproceedings{ester1996density,
  title={A density-based algorithm for discovering clusters in large spatial databases with noise},
  author={Ester, Martin and Kriegel, Hans-Peter and Sander, J{\"o}rg and Xu, Xiaowei and others},
  booktitle={kdd},
  volume={96},
  number={34},
  pages={226--231},
  year={1996}
}

@article{xu2022fast,
  title={Fast-lio2: Fast direct lidar-inertial odometry},
  author={Xu, Wei and Cai, Yixi and He, Dongjiao and Lin, Jiarong and Zhang, Fu},
  journal={IEEE Transactions on Robotics},
  volume={38},
  number={4},
  pages={2053--2073},
  year={2022},
  publisher={IEEE}
}

@article{oquab2023dinov2,
  title={Dinov2: Learning robust visual features without supervision},
  author={Oquab, Maxime and Darcet, Timoth{\'e}e and Moutakanni, Th{\'e}o and Vo, Huy and Szafraniec, Marc and Khalidov, Vasil and Fernandez, Pierre and Haziza, Daniel and Massa, Francisco and El-Nouby, Alaaeldin and others},
  journal={arXiv preprint arXiv:2304.07193},
  year={2023}
}

@inproceedings{zhai2023sigmoid,
  title={Sigmoid loss for language image pre-training},
  author={Zhai, Xiaohua and Mustafa, Basil and Kolesnikov, Alexander and Beyer, Lucas},
  booktitle={Proceedings of the IEEE/CVF international conference on computer vision},
  pages={11975--11986},
  year={2023}
}

@article{qi2017pointnet++,
  title={Pointnet++: Deep hierarchical feature learning on point sets in a metric space},
  author={Qi, Charles Ruizhongtai and Yi, Li and Su, Hao and Guibas, Leonidas J},
  journal={Advances in neural information processing systems},
  volume={30},
  year={2017}
}

@inproceedings{yu2021pointr,
  title={Pointr: Diverse point cloud completion with geometry-aware transformers},
  author={Yu, Xumin and Rao, Yongming and Wang, Ziyi and Liu, Zuyan and Lu, Jiwen and Zhou, Jie},
  booktitle={Proceedings of the IEEE/CVF international conference on computer vision},
  pages={12498--12507},
  year={2021}
}

@article{kasten2023point,
  title={Point cloud completion with pretrained text-to-image diffusion models},
  author={Kasten, Yoni and Rahamim, Ohad and Chechik, Gal},
  journal={Advances in Neural Information Processing Systems},
  volume={36},
  pages={12171--12191},
  year={2023}
}

@inproceedings{yan2025symmcompletion,
  title={SymmCompletion: High-Fidelity and High-Consistency Point Cloud Completion with Symmetry Guidance},
  author={Yan, Hongyu and Li, Zijun and Luo, Kunming and Lu, Li and Tan, Ping},
  booktitle={Proceedings of the AAAI Conference on Artificial Intelligence},
  volume={39},
  number={9},
  pages={9094--9102},
  year={2025}
}

@inproceedings{li2023blip,
  title={Blip-2: Bootstrapping language-image pre-training with frozen image encoders and large language models},
  author={Li, Junnan and Li, Dongxu and Savarese, Silvio and Hoi, Steven},
  booktitle={International conference on machine learning},
  pages={19730--19742},
  year={2023},
  organization={PMLR}
}

@inproceedings{radford2021learning,
  title={Learning transferable visual models from natural language supervision},
  author={Radford, Alec and Kim, Jong Wook and Hallacy, Chris and Ramesh, Aditya and Goh, Gabriel and Agarwal, Sandhini and Sastry, Girish and Askell, Amanda and Mishkin, Pamela and Clark, Jack and others},
  booktitle={International conference on machine learning},
  pages={8748--8763},
  year={2021},
  organization={PmLR}
}

@article{achiam2023gpt,
  title={Gpt-4 technical report},
  author={Achiam, Josh and Adler, Steven and Agarwal, Sandhini and Ahmad, Lama and Akkaya, Ilge and Aleman, Florencia Leoni and Almeida, Diogo and Altenschmidt, Janko and Altman, Sam and Anadkat, Shyamal and others},
  journal={arXiv preprint arXiv:2303.08774},
  year={2023}
}

@inproceedings{huang2023neural,
  title={Neural kernel surface reconstruction},
  author={Huang, Jiahui and Gojcic, Zan and Atzmon, Matan and Litany, Or and Fidler, Sanja and Williams, Francis},
  booktitle={Proceedings of the IEEE/CVF Conference on Computer Vision and Pattern Recognition},
  pages={4369--4379},
  year={2023}
}

@inproceedings{katz2015visibility,
  title={On the visibility of point clouds},
  author={Katz, Sagi and Tal, Ayellet},
  booktitle={Proceedings of the IEEE international conference on computer vision},
  pages={1350--1358},
  year={2015}
}

@inproceedings{cao2020hierarchical,
  title={Hierarchical coverage path planning in complex 3d environments},
  author={Cao, Chao and Zhang, Ji and Travers, Matt and Choset, Howie},
  booktitle={2020 IEEE International Conference on Robotics and Automation (ICRA)},
  pages={3206--3212},
  year={2020},
  organization={IEEE}
}

@article{wu2024uav,
  title={A UAV-based sparse viewpoint planning framework for detailed 3D modelling of cultural heritage monuments},
  author={Wu, Zebiao and Marais, Patrick and R{\"u}ther, Heinz},
  journal={ISPRS Journal of Photogrammetry and Remote Sensing},
  volume={218},
  pages={555--571},
  year={2024},
  publisher={Elsevier}
}

@article{du2025efficient,
  title={An Efficient UAV Coverage Path Planning Method for 3-D Structures},
  author={Du, Jiaxin and Huang, Baoqi and Jia, Bing},
  journal={IEEE Internet of Things Journal},
  year={2025},
  publisher={IEEE}
}

@article{hart1968formal,
  title={A formal basis for the heuristic determination of minimum cost paths},
  author={Hart, Peter E and Nilsson, Nils J and Raphael, Bertram},
  journal={IEEE transactions on Systems Science and Cybernetics},
  volume={4},
  number={2},
  pages={100--107},
  year={1968},
  publisher={IEEE}
}

@article{han2009pid,
  title={From PID to active disturbance rejection control},
  author={Han, Jingqing},
  journal={IEEE transactions on Industrial Electronics},
  volume={56},
  number={3},
  pages={900--906},
  year={2009},
  publisher={IEEE}
}

@article{helsgaun2000effective,
  title={An effective implementation of the Lin--Kernighan traveling salesman heuristic},
  author={Helsgaun, Keld},
  journal={European journal of operational research},
  volume={126},
  number={1},
  pages={106--130},
  year={2000},
  publisher={Elsevier}
}

@article{wang2022geometrically,
  title={Geometrically constrained trajectory optimization for multicopters},
  author={Wang, Zhepei and Zhou, Xin and Xu, Chao and Gao, Fei},
  journal={IEEE Transactions on Robotics},
  volume={38},
  number={5},
  pages={3259--3278},
  year={2022},
  publisher={IEEE}
}

@article{liu1989limited,
  title={On the limited memory BFGS method for large scale optimization},
  author={Liu, Dong C and Nocedal, Jorge},
  journal={Mathematical programming},
  volume={45},
  number={1},
  pages={503--528},
  year={1989},
  publisher={Springer}
}

@inproceedings{han2019fiesta,
  title={Fiesta: Fast incremental euclidean distance fields for online motion planning of aerial robots},
  author={Han, Luxin and Gao, Fei and Zhou, Boyu and Shen, Shaojie},
  booktitle={2019 IEEE/RSJ International Conference on Intelligent Robots and Systems (IROS)},
  pages={4423--4430},
  year={2019},
  organization={IEEE}
}

@inproceedings{deitke2023objaverse,
  title={Objaverse: A universe of annotated 3d objects},
  author={Deitke, Matt and Schwenk, Dustin and Salvador, Jordi and Weihs, Luca and Michel, Oscar and VanderBilt, Eli and Schmidt, Ludwig and Ehsani, Kiana and Kembhavi, Aniruddha and Farhadi, Ali},
  booktitle={Proceedings of the IEEE/CVF conference on computer vision and pattern recognition},
  pages={13142--13153},
  year={2023}
}

@inproceedings{wu2023omniobject3d,
  title={Omniobject3d: Large-vocabulary 3d object dataset for realistic perception, reconstruction and generation},
  author={Wu, Tong and Zhang, Jiarui and Fu, Xiao and Wang, Yuxin and Ren, Jiawei and Pan, Liang and Wu, Wayne and Yang, Lei and Wang, Jiaqi and Qian, Chen and others},
  booktitle={Proceedings of the IEEE/CVF Conference on Computer Vision and Pattern Recognition},
  pages={803--814},
  year={2023}
}

@inproceedings{xiang2021snowflakenet,
  title={Snowflakenet: Point cloud completion by snowflake point deconvolution with skip-transformer},
  author={Xiang, Peng and Wen, Xin and Liu, Yu-Shen and Cao, Yan-Pei and Wan, Pengfei and Zheng, Wen and Han, Zhizhong},
  booktitle={Proceedings of the IEEE/CVF international conference on computer vision},
  pages={5499--5509},
  year={2021}
}

@misc{djim30,
  title        = {DJI Matrice 30},
  author       = {DJI},
  url          = {https://enterprise.dji.com/matrice-30},
  year         = {2025}
}

@misc{livoxlidar,
  title        = {Livox Mid-360 User Manual},
  author       = {Livox},
  url          = {https://livoxtech.com/mid-360},
  year         = {2025}
}

@misc{orinnx,
  title        = {NVIDIA Jetson Orin NX},
  author       = {NVIDIA},
  url          = {https://www.nvidia.com/en-us/autonomous-machines/embedded-systems/jetson-orin/},
  year         = {2025}
}

@article{hoffman1984parts,
  title={Parts of recognition},
  author={Hoffman, Donald D and Richards, Whitman A},
  journal={Cognition},
  volume={18},
  number={1-3},
  pages={65--96},
  year={1984},
  publisher={Elsevier}
}

@article{biederman1987recognition,
  title={Recognition-by-components: a theory of human image understanding.},
  author={Biederman, Irving},
  journal={Psychological review},
  volume={94},
  number={2},
  pages={115},
  year={1987},
  publisher={American Psychological Association}
}

@misc{rc,
  title        = {RealityScan},
  author       = {Epic},
  url          = {https://www.realityscan.com/en-US},
  year         = {2025}
}

@misc{airsim,
  title        = {AirSim},
  author       = {MicrosoftResearch},
  url          = {https://microsoft.github.io/AirSim/},
  year         = {2025}
}

@article{choset2000coverage,
  title={Coverage of known spaces: The boustrophedon cellular decomposition},
  author={Choset, Howie},
  journal={Autonomous Robots},
  volume={9},
  number={3},
  pages={247--253},
  year={2000},
  publisher={Springer}
}

\appendix
\section{Appendix}

This appendix provides supplementary video materials that qualitatively illustrate and further validate the proposed \textbf{F}ly\textbf{C}o system.

\textbf{Extension 1: Overall System Overview and Motivation.} 
This video provides an overview of the \textbf{F}ly\textbf{C}o system and its research context.
It first introduces the motivation behind autonomous 3D target structure scanning in open-world environments and highlights the limitations of existing paradigms.
The video then introduces the motivations behind our system design and summarizes the advancement against existing paradigms, highlighting the novelty of integrating foundation models to empower drones.
Finally, representative results are presented, including in-the-wild flight demonstrations and simulation visualizations, to illustrate the system's effectiveness and its ability to address the identified research challenges.

\textbf{Extension 2: Precise Scene Understanding for Adaptive Scanning—Arch Bridge Linking Large Buildings.}
This video corresponds to Sec. \nameref{subsubsec:scene_understanding}.
It demonstrates \textbf{F}ly\textbf{C}o's ability to accurately ground user-specified targets in cluttered real-world scenes and to adaptively scan a distinctive arch bridge linking two large buildings, based solely on low-effort human prompts.

\textbf{Extension 3: Robust Scanning for Large-scale Structures—Concert Hall within Campus.}
This video corresponds to Sec. \nameref{subsubsec:robust_large_scale}.
It showcases \textbf{F}ly\textbf{C}o's performance in a large-scale scenario: scanning a 100 m $\times$ 60 m $\times$ 35 m concert hall nestled among hills and lakes. 
This experiment specifically validates the robustness and stability during large-scale missions—addressing performance degradation caused by sparse initial observations, all enabled by the system's generalizable surface prediction capability.

\textbf{Extension 4: Efficient and Safety-assured Scanning within Clutters—Castle Gate amidst Park Trees.}
This video corresponds to Sec. \nameref{subsubsec:safe_efficient_clutters}.
It demonstrates the full deployment of the \textbf{F}ly\textbf{C}o system for scanning a castle gate. 
This test showcases the system's ability to seamlessly integrate flight skills with guidance from foundation models, achieving efficient and safe scanning flights while ensuring complete target coverage.

\textbf{Extension 5: Efficient and Safety-assured Scanning within Clutters—Red-brick Building within a Dense Forest.}
This video corresponds to Sec. \nameref{subsubsec:safe_efficient_clutters}.
It records the complete procedure of \textbf{F}ly\textbf{C}o autonomously scanning a red-brick building surrounded by trees. 
The results demonstrate \textbf{F}ly\textbf{C}o’s ability to simultaneously maintain scanning efficiency, comprehensive coverage, and real-time obstacle avoidance in highly cluttered environments.

\textbf{Extension 6: Benchmark Comparisons and Ablation Studies.}
This video corresponds to Sec. \nameref{subsec:comparisons} and Sec. \nameref{subsec:ablation_studies}.
It presents qualitative and quantitative comparisons between \textbf{F}ly\textbf{C}o and representative state-of-the-art approaches, highlighting advancements in mission efficiency, coverage completeness, and flight safety.
The video further includes visualizations of ablation studies, validating the contribution of individual system components and supporting the design choices of the proposed system architecture.

\textbf{Access to High-resolution Videos.}
High-resolution versions of all supplementary videos are publicly available via the YouTube playlist: \url{https://www.youtube.com/playlist?list=PLqjZjnqsCyl40rw3y15Yzc7Mdo-z1y2j8}.

\end{document}